# Deep Apprenticeship Learning for Playing Games

Dejan Markovikj

Linacre College

University of Oxford

Supervised by Prof. Nando de Freitas

Department of Computer Science, University of Oxford

A dissertation submitted in partial fulfilment

of the requirements for the degree of

*Master of Science in Computer Science*

Trinity 2014

*Due to the nature and complexity of the project, this research work was performed in full collaboration with Miroslav Bogdanovic. Such arrangement was fully acknowledged and approved by my project supervisor, Prof. Nando de Freitas.*

***Man's mind, once stretched by a new idea, never regains its original dimensions.***

- Oliver Wendell Holmes, Sr.

# Abstract

Reinforcement learning is becoming an increasingly popular research area, largely based on the fact that it combines the methodology of artificial intelligence and machine learning with the theory of psychology in order to solve learning problems based on interaction with the environment, which represents one of the core learning principles for humans and animals. Nevertheless, the fundamental assumption of this learning paradigm is knowledge of the reward function that formally explains the problem under investigation, which can be overoptimistic for certain tasks. On the other hand, apprenticeship learning is a paradigm where such a function is not assumed to be know. Instead, an expert behaviour for such a task is given.

In the last decade, deep learning has achieved great success in machine learning tasks where the input data can be represented on different levels of abstractions. Driven by the recent research in reinforcement learning using deep learning models, we explore the feasibility of using the apprenticeship learning approach for complex, multidimensional input data. We propose a novel method for apprenticeship learning based on the previous research on supervised learning techniques in reinforcement learning. Our method is applied to video frames from Atari games in order to teach an artificial agent to play those games. Even though the reported results are not comparable with the state-of-the-art results in reinforcement learning, we demonstrate that such an approach has the potential to achieve strong performance in the future and is worthwhile for further research.

# Acknowledgements

This dissertation is both the biggest academic challenge I have ever faced and my greatest achievement so far. It is a result of the guidance, commitment and support of many people throughout my entire life. Here, I would like to take the time and acknowledge them.

First of all, I would like to express the deepest appreciation to my supervisor, Prof. Nando de Freitas, for his guidance and suggestions on this project and for always finding time to meet me and discuss interesting ideas. During the last year, he has been the greatest source of inspiration for me. All the reading group sessions he organized, the talks and lectures which he suggested I go to, and the meetings during which we initially discussed possible projects, provided a lifetime inspiration for me and my research, which has reached far beyond the boundary of the subject of this project. I am thankful to him for introducing me to deep learning, computer vision, apprenticeship learning and giving me a new perspective of reinforcement learning and machine learning in general.

This project represents a joint project with Miroslav Bogdanovic, and it would not have been close to the level of quality that it is now, without his ideas, insight and work. I am thankful to Miroslav for being a wonderful colleague and it was a great pleasure for me to work with him. I also want to thank him for all the discussions on topics including machine learning, but also generally academia, research, and many more.

In addition, I would like to thank Misha Denil for his ideas and technical help during this project. I can easily say that he saved us months of work because of his experience with similar problems as the ones encountered in this project.

I express gratitude to my course supervisor, Prof. Edith Elkind, for following my progress throughout the entire course and giving me invaluable advice that led me to this point. I am grateful to all the lecturers and academic staff at the Department of Computer Science, which directly or indirectly contributed to the academic knowledge I gained during this course. In particular, I would like to thank Prof. Phil Blunsom, who acquainted me with a new approach for machine learning that changed my perspective for the entire field. I would also like to thank Wendy Adams, for her help during this course, and the IT office at my department, especially Ash, for doubling the computing resources I used many times.

I would like to express appreciation to Linacre College for being my home far away from home, and to all its personnel and students for being my family for the duration of this course, and generally, for making this journey much more joyful and amazing.

I would like to express special appreciation and thanks to my supervisor and professor during my undergraduate studies, Prof. Sonja Gievska, for helping me to discover my love for artificial intelligence and the passion for research. I learned so much from her, and no words can describe my gratitude for all she has done for me.

Furthermore, I wish to thank Marija Kuzmanovic, for giving me insight about studying abroad and convincing me that I am a good fit for institutions such as the University of Oxford. A thank you also goes out to Petar Joksimovic for our long Skype conversations, to Milan Petrov for our even longer philosophical discussions at five in the morning, and to Aneta Naumoska for proofreading this dissertation and giving valuable remarks.

I would like to express my indebtedness to the Ministry of Education and Science of the Republic of Macedonia for fully funding my studies. Additionally, I wish to thank and acknowledge Vaska Radulovska Kiprijanovska, Trajche Stoev, Dimitar Panovski, Talija Penshovska, and Snezhana Stojcheska for being there for me when I needed it the most.

Finally, I would like to express appreciation to my family and friends for all the support, encouragement and unconditional love they have given me. Special thanks to my parents for always believing in me. I hope I will continue to make you proud in the future.

I dedicate this dissertation

to my parents Ivica and Sofche Markovikj, aunt Katica, Milan,

and in the memory of aunt Sonja.

# Contents

# List of Figures

# List of Tables

# List of Algorithms

CHAPTER 1

# Introduction

*"First, inevitably, the idea, the fantasy, the fairy tale. Then, scientific calculation. Ultimately, fulfillment crowns the dream."*

– Konstantin E. Tsiolkovsky, *1926*

## 1.1 Motivation

The technological and scientific advancement in the last few decades culminated with breakthroughs of scientific discoveries that completely changed our perspective of the world. In that process, particularly the contribution of one scientific field, stands out of all the science disciplines. That field, *computer science*, albeit with very short history, proved its potential to impact the society. One of the most popular areas of this discipline is inevitably *artificial intelligence* (AI), which has a goal to create beings (be it software agents or mechanical robots) that will reach and even surpass the human intelligence. The result would be an ideal society where humans do not need to work in order to have time to be mentally challenged as well as to progress intellectually. All of the work will be done by these intellectual beings, which will eventually find solutions to all the problems that humanity has encountered, or can encounter in the future. It was soon after the initial hype that people realized that such a utopistic scenario is far, far away from our capabilities. Many scientists were sceptical with the whole idea and wanted to abandon it. Many others, however, thought that such a complex idea needs to be approached by dividing it into simple sub-problems that are manageable. Indeed, as Caesar said a long time ago "Divide at impera", being the same approach that led to some success in the case of the AI problem. As a consequence, numerous computational models, methods, and techniques arose, each attacking a specific sub-problem. This process, nevertheless, was far from constantly successful, since there were many setbacks and challenges, and while some of them were solved years or decades later, some of them are not solved even to this day. Although there were periods of massive cuts in funding in the AI research, known historically as 'AI winters', after many years of research, some of the methods and approaches produced satisfactory results. Driven by the relative success compared to the initial disappointment, researchers realized that systematic and careful methodology used to solve sub-problems, as part of a bigger, more complex problem is a beneficial approach. Eventually, such philosophy resulted with many algorithms and exact methods. We can combine many of those methods in a field called *machine learning*. Machine learning can be seen as the applicable side of AI, consisting of algorithms that have the goal to learn some concepts based on given input data. Many successful methods and applications resulted in a huge interest for machine learning. Even though, we are far from the final goal of the AI, machine learning contributed to significant scientific discoveries.

Machine learning methods need data in order to learn concepts. The problem is that such data is not often available. Even when it is, the process of data collection and labelling is very expensive, since it require human supervision. However, with the advancement of technology and its incorporation in our everyday life, this problem is being alleviated as time passes. Another consequence of the advancement of technology is that more complex data is available as well. Consequently, other problems arose pertaining to using standard machine learning methods on such complex input datasets. Namely, traditional models are not powerful enough to fit complex data. Extensive scientific research combined with the huge technological progress contributed to successful implementations of learning algorithms on complex data. Such models are called *deep models*, and the paradigm itself is called *deep learning*.

Although the concept was introduced more than 30 years ago, deep learning had its first working implementations in the last decade. It is especially successful in computer vision problems, as well as speech recognition and since lately in natural language processing, which will be discussed in more detail in Section 2.5. As an example, one very interesting such project is Google Brain, where deep neural networks were trained to recognize cats in YouTube videos (Markoff, 2012). Although widely used in the last few years, deep models were successfully employed only for supervised and unsupervised learning tasks. As some might know, traditionally, there is another learning approach called *reinforcement learning*, where an agent has the task to learn how to behave in a given environment by perceiving the impact of that environment on the agent in different situations. Moreover, such an approach is very natural in terms of how human beings and animals learn in many situations. Other such examples will be explained further in this dissertation. Furthermore, one might ask why there is no research in terms of using deep models for reinforcement learning problems. Albeit such question was valid until recently, a few months ago, an artificial intelligence start-up called DeepMind proposed a method for training deep models for reinforcement learning problems, where the input is of many dimensions (Mnih *et al.* , 2013). Immediately upon the presentation of their method, it was revealed that the tech-giant Google acquired this start-up for an astonishing price (Gibbs, 2014). This attracted even more attention to deep learning, not only in academic circles but in the public eye as well.

In terms of how humans learn, one might notice that some learning tasks cannot be represented as reinforcement learning problems. Indeed, consider a case where an infant tries to behave as her/his parents (for example, imitating how they walk or how they eat). Other similar examples include a teenager who learns how to drive a car, a child who learns to play a new sport, and people imitating their idols. Imitation as a behaviour is deeply ingrained in our genetic code, but we will not delve into detail on this topic because such discussion will be more appropriate for the topic of psychology or sociology research. Therefore, one might ask how we can present such problems in machine learning. A few years ago, a paradigm called *apprenticeship learning* was introduced, according to which, instead of feedback from the environment, the agent is presented with expert behaviour which is used to learn that specific behaviour. It should be noted however that there are only few papers that propose such methods, which means that much research can and need to be done in that direction. Since imitation is a very important aspect of human psychology and how we learn, it seems natural to try to investigate such an approach by using models that have been especially successful in the last few years. Driven by the success of DeepMind in deep reinforcement learning, the idea of this research is to employ a *deep learning model for apprenticeship learning tasks*. In our knowledge, there is no other apprenticeship learning approach using deep models on complex input data. If successful, such an approach can have great scientific impact and innovative applications. Although, as it will be seen, apprenticeship learning is a complex paradigm with many problems, if at some point we overcome those problems, such an approach can be used for very interesting applications; an example being a general purpose robot that will learn how to perform simple and even more complex tasks solely by observing its owner as to how such tasks need to be performed. As one might notice, this example presents a huge step toward the AI goal. Therefore, inspired by such scenarios which are unrealistic at the current time, we intend to research the possibilities of successful application of an apprenticeship learning method.

## 1.2 Problem Statement

While reinforcement learning offers a framework for solving problems which are not suitable for supervised or unsupervised learning, one of the fundamental concepts which this approach relies on is an initially known a reward function. As will be later discussed, that assumption may not be appropriate for numerous learning tasks due to the fact that such functions can be hard to define or construct. For such tasks, we can use apprenticeship learning, which presents a methodology for solving problems, where instead of reward function, expert behaviour is given to reflect the goal of the problem. In this project, we present an approach for using apprenticeship learning, specifically when the input data is complex and consists of features having more levels of abstraction. We apply such an approach in games from one of the most popular consoles in the 80-ies - Atari. The task is to *teach an artificial agent how to play Atari games only by having video sequences of game states and actions, trajectories, of a human player that demonstrates expert behaviour for the specific games.*

## 1.3 Objectives

For the purposes of this research project, we will define the objectives and hypotheses that will help us to evaluate the performance of our method when used in the application area relevant to our research.

### 1.3.1 Comparison with Human Agent

The method used can successfully retrieve a policy, which, if followed by an artificial agent for a task defined as learning to play a game, will result in a score comparable to human's score on that game. Such an agent will attain at least $30\%$ of the cumulative reward obtained by the human player, where the cumulative reward is defined as the number of points, or total score, for the given game.

### 1.3.2 Comparison with Artificial Agent

The method used can successfully retrieve a policy, which, if followed by an artificial agent for a task defined as learning to play a game, will result in a score comparable to a SARSA agent. A SARSA agent is an artificial agent, which uses one of the standard reinforcement learning algorithms - SARSA, in order to obtain a policy. In such a case our agent will attain at least $30\%$ of the cumulative reward obtained by the SARSA agent, where a cumulative reward is defined as the number of points, or total score, for the given game.

### 1.3.3 Comparison with Random Agent

The method used can successfully retrieve a policy, which, if followed by an artificial agent for a task defined as learning to play a game, will result in a higher score compared to a random player. A random player executes each of the actions uniformly with constant probability, which means that each of the actions can be executed with equal probability regardless of the state. Such an agent will attain at least $100\%$ of the cumulative reward obtained by the random player, where the cumulative reward is defined as the number of points, or total score, for the given game.

### 1.3.4 Detecting Features

The network structure used for learning mapping from states to actions, can successfully identify groups of pixels of the input images that are most informative for the outcome of the learning process. Those groups of pixels represent the features important for the task. A group of pixels is considered to represent the recognized feature when the gradient of those pixels with respect to the parameters of the model, change significantly during the learning process.

### 1.3.5 Effect of Used Non-Linearity to the Learning Process

The method used produces better results when the neural network employed has rectifier non-linearity compared to other non-linearities as an activation function. Better results are defined by a lower test error (error on the test set).

### 1.3.6 Pre-Training for More Efficient Learning Process

Pre-training techniques such as Independent Component Analysis (ICA) and Reconstructed ICA can produce better results compared to training where the weights initially have random values. Better results are defined by a lower test error.

## 1.4 Dissertation Structure

This dissertation can be divided into three logical sections. The first section introduces the reader to the background theory needed to fully comprehend the research work. The next section presents the apprenticeship learning approach we propose. In a detailed way we discuss topics, such as the structure and the training of our networks, and the data collection process, among others. The final logical section presents an overview of future work, including both short-term plans as well as future directions, and the conclusions drawn from our experience gained while working on this project.

**Chapter 2** covers four main areas that our work is based on: neural networks, reinforcement learning, apprenticeship learning, and deep learning. Substantial time will be spent on explaining the basics of neurons and neural networks, and how to train them, followed by discussion about one specific type of neural networks, convolutional networks, as a very important concept for our work. The part regarding reinforcement learning covers the theoretical basics, such as Markov Decision Processes, policies, value functions, and Q-learning. The material about apprenticeship learning is divided into two portions, based on approaches for solving the apprenticeship learning problem. We emphasize the inverse reinforcement learning approach as well as imitation learning as key concepts for our work. We end this chapter with introductory discussion about deep learning, covering one important method for reinforcement learning in deep settings with a similar application as in our work.

**Chapter 3** includes details about the approach that we propose to implement for apprenticeship learning tasks. First, the whole process of data creation and pre-processing is explained, including a short discussion about possible problems that can arise in this process in case someone attempts to replicate our work. A detailed discussion then follows about the networks that we use for learning the state to action mapping and estimating the reward function, as well as discussion on the training process. This chapter concludes with a summary of our approach in an algorithm called Deep Apprenticeship Learning (DAL).

**Chapter 4** gives an interpretation of the results of our work. First, analysis of the networks structure is given accompanied by a discussion as to which structure produces the best results and why. Moreover, images of the learned filters and feature maps are presented. Next, the reward function obtained is analysed, including a discussion about the cumulative reward for different cases. Finally, we present influence maps, which for each pixel of the input image show how important that pixel is in the learning process, effectively identifying the fragments of the image where the features are recognized.

**Chapter 5** presents future work pertaining to this project as well as the conclusions drawn from our work. The section on future work is divided into two portions: one part on our short term goals, and the other on possible future directions.

Finally, two appendices are given. The first one (**Appendix A**) includes the code needed to implement our method and to collect the data, while the second one (**Appendix B**) contains a video demonstration of an agent who follows a policy as described in Chapter 3.

CHAPTER 2

# Theoretical Background

*"Research is what I'm doing when I don't know what I'm doing."*

– Wernher von Braun, *1957*

In this chapter we cover the underlying theory on which our work is based on. For readers not familiar with machine learning, or more specifically, reinforcement learning, apprenticeship learning, neural networks, and deep learning, we aim to provide an introduction to the theory needed for comprehension of the subject covered in our research. For researchers who are already introduced to the subject matter, however, the goal is to provide concise review of the underlying theory, for easier replication of our research, and potential creation of new research methodologies and directions on this subject, inspired by our approach.

We start by defining neuron, the basic building block for neural networks. Then we introduce neural networks, followed by an overview of the training algorithm for such models, known as backpropagation. After we introduce neural networks, we focus on one specific subtype called convolutional neural networks. They are very important for our work, so we will spend substantial time in explaining how these networks work. What follows is an overview of the basic reinforcement learning theory which includes formal definition through Markov decision processes, value functions, methods of obtaining policies, and one very successful such method, called Q-Learning. We continue with a discussion on apprenticeship learning, the paradigm which our method belongs to. After a general introduction, two possible ways of solving apprenticeship learning problems are discussed. More time will be spent on inverse reinforcement learning and possible methods for solving such tasks. After the introduction of both learning paradigms, we discuss their connection and present a high-level overview of all methods. This is subsequently followed by deep learning, as a paradigm which has had huge success in the last decade in solving problems based on complex input representations. We discuss deep reinforcement learning and one specific approach which in terms of application is the only project that has similarities with ours.

For this chapter, we assume that the reader has basic knowledge in the fundamental concepts of the probability theory. We also assume that she has an understanding of the basic concepts in machine learning, such as: models, overfitting, training, etc. When we discuss neural networks we assume knowledge of the logistic regression model. For introduction in all this matter we recommend Murphy (2012) or Bishop (2006).

## 2.1 Neurons and Neural Networks

### 2.1.1 Neurons

First, we should note that in this dissertation by neuron we refer to an artificial neuron which represents a model of a biological neuron. A *neuron*, also called *perceptron*, can be defined as a unit which takes given inputs, applies a certain function to those inputs and produces output. The inputs are added up after each of them is weighed with a corresponding weight between 0 and 1. The resulting value after these operations is also called *pre-synaptic*[1] value for that neuron. A function called *activation function* is then applied to the pre-synaptic value. The result of applying this function is called *post-synaptic* value and is the output of the neuron. Such model of a neuron is shown in Figure 2.1, where $x_0, x_1, ..., x_K$ are the inputs ($x_0 = 1$ is called bias determined by $w_0$), $w_0, ..., w_K$ are the weights, $a$ is the pre-synaptic value, $f$ is the activation function, $z$ is the post-synaptic value, and $K$ is the number of inputs to the neuron.

[1] Derived from the biological definition of neuron, where a synapse is basically a neural function which passes an electrical signal to other neurons.

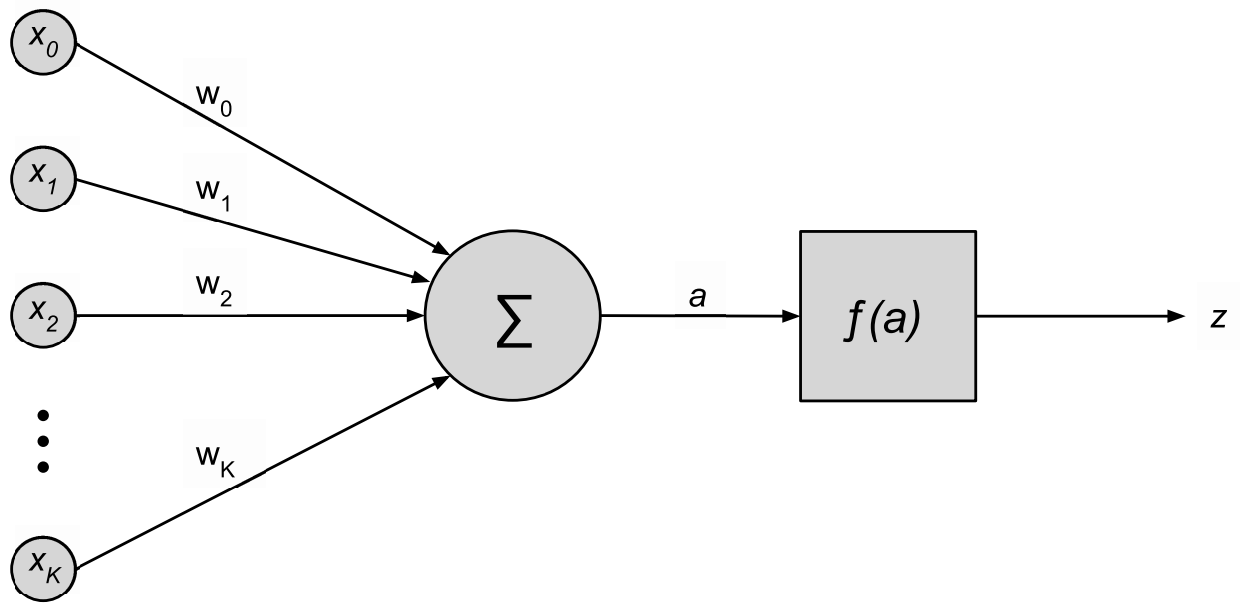

**Figure 2.1:** Model of Artificial Neuron

The computational model of a neuron is determined by its activation function. One of the most used such functions is the logistic function defined as

$$\sigma(x) = \frac{1}{1 + \exp^{-x}}. \tag{2.1}$$

The graph of this function is given in Figure 2.2. A neuron with logistic function as an activation function behaves as a logistic regression model. For more information on logistic regression we recommend Murphy (2012). In the last decade another function, called rectifier, is very popular as a choice for activation function, especially in the deep learning research community. This function is defined as

$$f(x) = \max(0, x), \tag{2.2}$$

but most often a smooth approximation is used, given by

$$f(x) = \log(1 + \exp^{x}). \tag{2.3}$$

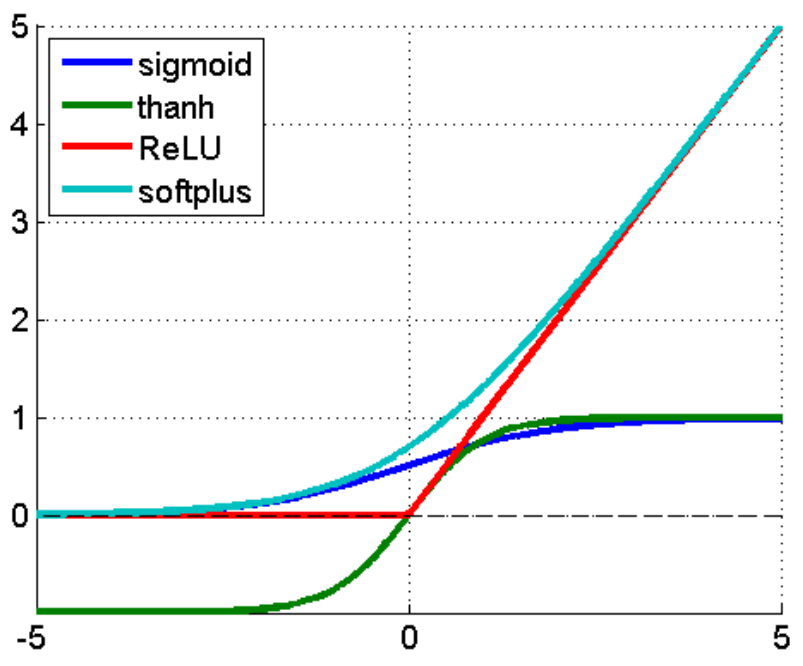

**Figure 2.2:** Commonly used activation functions

Neurons employing rectifier as an activation function are called rectified linear units (ReLU). It is claimed that this function is biologically plausible and more practical then other functions including logistic and hyperbolic tangent, among others (Glorot *et al.* , 2011). This function is given in 2.2[2]. The biggest advantage of rectifier is that it solves the 'gradient vanishing' problem (Hochreiter *et al.* , 2001), which will be discussed in Section 2.5. One of the goals in our project is to test if this statement applies in our case.

### 2.1.2 Neural Networks

*Artificial neural networks* are computational models inspired by the structure of the brain. The idea is to use the architecture of known, relatively simple, computational models as building blocks to create a structure that approximates the human brain. They consist of set of neurons, which are arranged in a structure where the neurons are interconnected in a specific way. A neural network is organized in hierarchical layers, such that each neuron must uniquely belong to only a single layer. A neuron of one layer can be connected only to neurons in the preceding or following layer. The inputs of a neural network are propagated as input to neurons, as explained in Section 2.1.1. These neurons form a layer, called hidden layer, hence the term hidden neurons. This name comes from the fact that the values of such neurons are not observed, contrary to the input or the output values of the network. The weights that determine the impact of the input to the pre-synaptic value of the hidden neurons in one layer form a matrix, called weight matrix. This matrix has one value for each pair of input and a hidden neuron, which controls how that input will affect the pre-synaptic value of the neuron[3]. One network can have more hidden layers, where the output (post-synaptic) values of neurons in one layer are taken as input to the neurons in the next (upper) layer. The last layer is called output layer and the post-synaptic values of this layer represent the output of the whole neural network model. Here, we conclude our short descriptive introduction about neural networks. Next, we introduce formalism for these models.

[2] Retrieved on 03.08.2014 from
`http://imiloainf.wordpress.com/2013/11/06/rectifier-nonlinearities/`

We shall denote the layers as $l_1, ..., l_n$, where $n$ is the number of layers. In terms of this notation, layers $l_1, ..., l_{n-1}$ represent hidden layers, while $l_n$ is the output layer. In the following discussion, we assume feed forward neural network, where the input is propagated unidirectionally from the first hidden layer, through all the other hidden layers, toward the output layer, without having loops or cycles in the network. For now, we shall assume a fully connected network, which implies that each of the neurons in layer $l_k$ is connected to each of the neurons in layer $l_{k+1}$, where $k = 1, ..., n-1$. More specifically, the post-synaptic values of neurons in layer $l_k$ represent input to neurons in layer $l_{k+1}$. All the connections between the two layers construct the weight matrix denoted as $W^{ij}$ between layers $l_i$ and $l_j$. This matrix has a number of rows equivalent to the number of neurons in the lower layer (in our case, $l_i$) and a number of columns equivalent to the number of neurons in the higher layer (in our case, $l_j$), subject to the constraint $i < j$. As it was already said, the weight between two neurons implies the extend to which the post-synaptic value of the neuron in the lower layer affects the pre-synaptic value of the neuron in the higher layer. The value of each weight can be between 0 and 1, such that 1 means a strong connection between two neurons, and 0 vice versa. In terms of the weight matrix, the weight between neuron $a$ from the layer $l_i$ and neuron $b$ from the layer $l_j$ (subject to $i < j$) is equivalent to the value $W^{ij}_{a,b}$, which is the cell from the $a$-th row and $b$-th column from the matrix $W^{ij}$. The pre-synaptic

[3] We can see the inputs as distinct features, and the task of the weights is to regulate how each of the features is important for the decision that needs to be made by the neuron

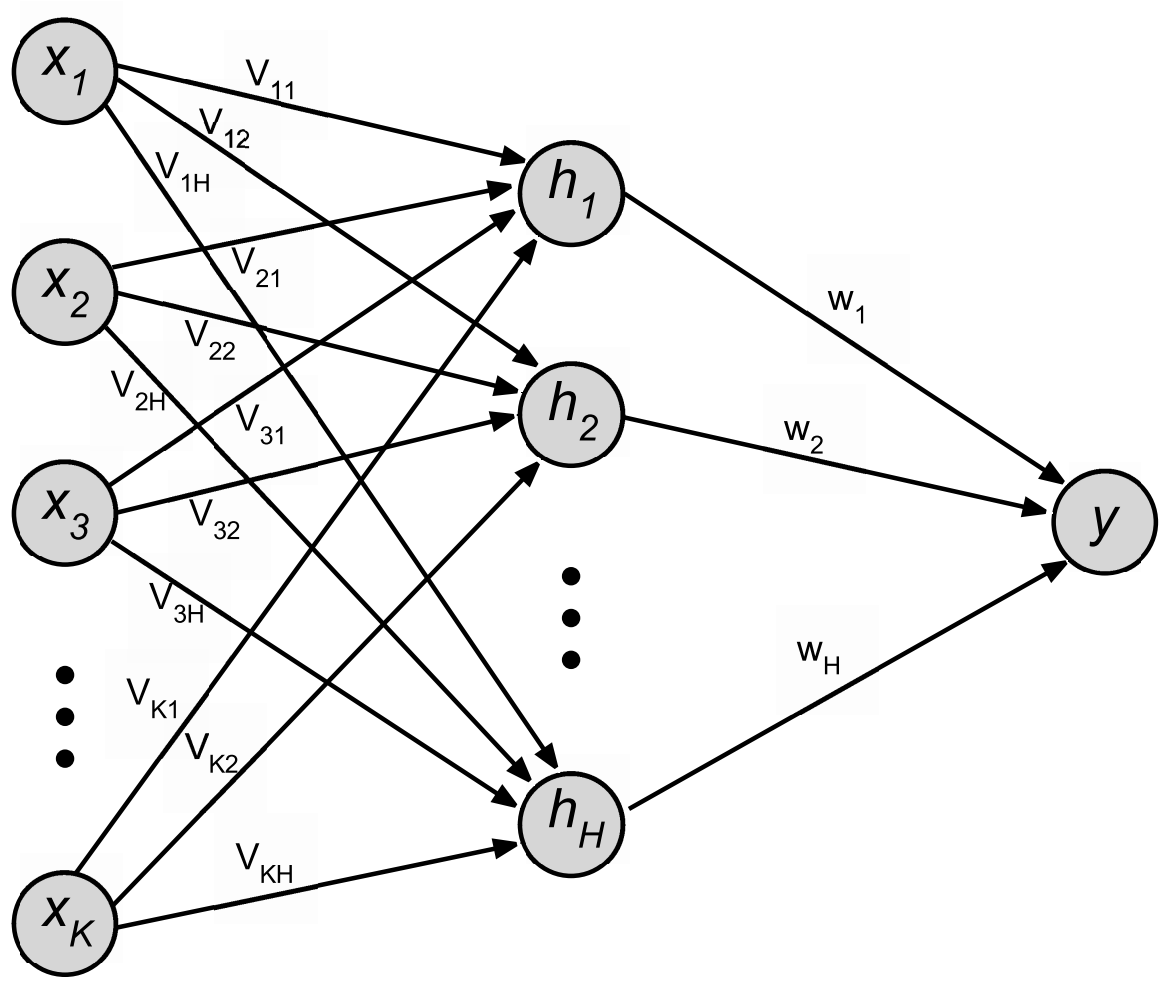

**Figure 2.3:** Structure of Simple Neural Network

value for the $i$-th hidden neuron can be computed as

$$a_i = \sum_j W_{ji} x_j, \tag{2.4}$$

while the post-synaptic value for the same neuron is given by

$$z_i = f(a_i) = f(\sum_j W_{ji} x_j). \tag{2.5}$$

#### 2.1.2.1 Neural Networks Training

After defining the structure of neural networks, we will now discuss how to train such networks. For the purposes of this section, we shall assume a neural network with $K$ inputs, one neuron in the output layer, and one hidden layer with $H$ neurons, as given in Figure 2.3. Let $\boldsymbol{\theta} = (\mathbf{V}, \mathbf{w})$ represent the parameters of the network, where $\mathbf{V}$ and $\mathbf{w}$ are weight matrices for the hidden and the output layer, respectively. For simplicity, we omit bias neurons, but in case we need biases, we should add one more input and one more hidden neuron, both with a value set to one. Now we define $(\mathbf{x_n}, \mathbf{y_n}) \in D$ as the $n$-th data point of a dataset $D$, such that $\mathbf{x_n} \in \mathbb{R}^K$ and $\mathbf{y_n} \in \mathbb{R}$. The pre-synaptic vector of the hidden layer for the $n$-th input instance $\mathbf{a_n}$, is given by $\mathbf{a_n} = [a_{n1}, a_{n2}, ..., a_{nH}]$ where

$$a_{nh} = \mathbf{V_h^T x_n} = \sum_{k=1}^{K} V_{kh} x_{nk} \quad s.t. \quad \forall h \in \{1, .., H\}, \forall k \in \{1, .., K\}, \tag{2.6}$$

while the post-synaptic vector of this layer is $\mathbf{z_n} = f(\mathbf{a_n})$, where $f$ is the activation function of the hidden neurons, a non-linearity which is applied to the pre-synaptic vector of the hidden layer. We define the pre-synaptic value of the output layer (a single neuron in this case) as

$$b_n = \mathbf{w^T z_n} = \sum_{h=1}^{H} w_h z_{nh} \quad s.t. \quad \forall h \in \{1, .., H\}, \tag{2.7}$$

and the post-synaptic value of the same layer as $\hat{y}_n = o(b_n)$. Here, $o(\cdot)$ represents a function applied to the pre-synaptic values of the output neuron(s) in order to obtain the post-synaptic output. It can be a smoothing function, for example soft-max, but here we will assume $o(b_n) = b_n$ for simplicity.

Based on the theory for logistic regression (see more in Murphy, 2012, Chapter 8), for a network where the neurons behave as logistic regression models, the cost function is given by

$$J(\boldsymbol{\theta}) = \sum_{n=1}^{N} (\hat{y}_n(\boldsymbol{\theta}) - y_n)^2. \tag{2.8}$$

In order to minimize this cost, which we will call *loss function*, we need to find estimates of the parameters $\boldsymbol{\theta}$ that minimize Equation 2.8. The problem here is that the loss function given in Equation 2.8 is not necessary convex in respect to the parameters $\boldsymbol{\theta}$. Therefore, we can not use direct methods, such as Lagrange multipliers, in order to find optimal estimates of the parameters, denoted as $\hat{\boldsymbol{\theta}}$. Instead, we use the *gradient descent* approach, where we iteratively minimize the loss function by tuning the parameters toward the direction of the respective gradients. The result is finding a local minimum after a number of iterations, when the parameters converge toward stationary values. For each data point, the gradient of the cost function in respect to the weights between the hidden and the output layer is given by

$$\nabla_{\mathbf{w}} J_n = \frac{\partial J_n}{\partial b_n}\frac{\partial b_n}{\partial \mathbf{w}} = \frac{\partial J_n}{\partial b_n}\mathbf{z_n}, \tag{2.9}$$

which follows from the chain rule in calculus and the definition of $b_n$. From Equation 2.8 we have

$$\frac{\partial J_n}{\partial b_n} \triangleq (\hat{y}_n - y_n) = \delta_n^w, \tag{2.10}$$

which is the difference between the output from the network (estimated output) and the output that is observed. Therefore, Equation 2.10 represents the error of the output layer, which will be denoted as $\delta_n^w$. When we use Equation 2.10 in Equation 2.9, we get

$$\nabla_{\mathbf{w}} J_n = \delta_n^w \mathbf{z_n}. \tag{2.11}$$

For the gradient of the cost function in terms of the weights between the input and the hidden layer we have

$$\nabla_{\mathbf{V_h}} J_n = \frac{\partial J_n}{\partial a_{nh}}\frac{a_{nh}}{\mathbf{V_h}} = \frac{\partial J_n}{\partial a_{nh}}\mathbf{x_n} = \delta_{nh}^V \mathbf{x_n}, \tag{2.12}$$

where $\delta_{nh}^V$ denotes the error of the hidden layer. For this error, we have

$$\delta_{nh}^V = \frac{\partial J_n}{\partial a_{nh}} = \frac{\partial J_n}{\partial b_n}\frac{\partial b_n}{\partial z_{nh}}\frac{\partial z_{nh}}{\partial a_{nh}} = \delta_n^w w_h f'(a_{nh}), \tag{2.13}$$

where $f'(a_{nh})$ is computed from the definition of $f$, but we will not give the derivation for specific example functions here.

From Equation 2.13, we can conclude that in order to compute the error signal of the hidden layer we first need to compute the error of the output layer. The same applies when a neural network consists of many hidden layers, such that the errors are back propagated from the output layer, through the higher hidden layers toward the lower hidden layers. Because the error of higher layers needs to be back-propagated in order to compute errors for lower layers, this algorithm is called *backpropagation*, given in Algorithm 1 (Werbos, 1974; Rumelhart *et al.* , 1988).

**Algorithm 1** Backpropagation Algorithm

***Given***

Dataset $D = \{(\mathbf{x_n}, y_n)\}_{n=0}^{N}$

Learning rate $\eta \in (0, 1)$

Parameters $\theta = (\mathbf{V}, \mathbf{w})$, initialized with random values

**for** $n = 1, ..., N$ **do**

***Forward pass*** data instance $\mathbf{x_n}$ through the network

- compute the pre-synaptic vector of the hidden layer $\mathbf{a_n}$ as defined in Eq. 2.6
- compute the post-synaptic vector of the hidden layer $\mathbf{z_n}$, where $\mathbf{z_n} = f(\mathbf{a_n})$
- compute the pre-synaptic value of the output layer $b_n$ as defined in Eq. 2.7
- compute the output $\hat{y}_n$, where $\hat{y}_n = o(b_n)$

***Back-propagate*** the error of Eq. 2.8 in order to update the parameters

- compute the error of the output layer $\delta_n^w$, defined as in Eq. 2.10
- update the weights between the hidden and the output layer $\mathbf{w}$

  update rule: $\mathbf{w} \leftarrow \mathbf{w} - \Delta\mathbf{w}$

  where $\Delta\mathbf{w} = \eta\frac{\partial J_n(\boldsymbol{\theta})}{\partial \mathbf{w}}$, such that $\frac{\partial J_n(\boldsymbol{\theta})}{\partial \mathbf{w}}$ is computed as in Eq. 2.11
- using $\delta_n^w$ compute the error of the hidden layer $\delta_n^V$ according to Eq. 2.13
- update the weights between the input and the hidden layer $\mathbf{V}$

  update rule: $\mathbf{V_h} \leftarrow \mathbf{V_h} - \Delta\mathbf{V_h} \quad s.t. \quad \forall h \in H$

  where $\Delta\mathbf{V_h} = \eta\frac{\partial J_n(\boldsymbol{\theta})}{\partial \mathbf{V_h}}$, such that $\frac{\partial J_n(\boldsymbol{\theta})}{\partial \mathbf{V_h}}$ is computed as in Eq. 2.12

**if** $\Delta\mathrm{W} \approx 0$ **and** $\Delta\mathrm{V} \approx 0$ **then**

**BREAK**

**end if**

**end for**

***return*** $\hat{\boldsymbol{\theta}}^* = (\mathbf{V}, \mathbf{w})$

### 2.1.3 Convolutional Neural Networks

From the introductory discussion about neural networks, we can conclude that during the training phase, hidden layers extract feature representation of the input. This can be seen as a process of learning features that best describe the underlying problem. If there are more hidden layers, upper hidden layers learn more abstract representations since they are fed by lower level layer outputs, which represent more low-level, simple features. Therefore, neural networks with many layers are more powerful models. However, the main problem is that such networks are hard to train.

Since the hidden neurons learn features, it seems natural to use neural networks in vision problems, where the input is represented as pixels of image. Indeed, the image can be seen as a composition of simple features (such as lines and angles) that form more complex features (such as basic forms), which in turn constitute even more abstract features (such as objects) and so on. Therefore, if we manage to create a neural network model, which will learn rudimentary features in the lower layers and more abstract representations in the higher layers, we can successfully use such a model for vision problems. However, there are few problems in such a scenario. First, in the case of image, the input is relatively complex, which implies that there will be an enormous number of parameters. The consequences are that such networks will need a huge amount of data in order to train a good model and the training process will be painfully slow. Moreover, those networks would not be invariant to distortions and translations, which means that if some pattern is found in different locations of an image it cannot be recognized as a single feature. Even in the case of successful training, many of the weights will be the same since they detect identical patterns in different locations of the image, which is not optimal. Another significant property that holds when working with picture inputs, is that the spatial structure of the image is very important. In other words, close pixels are much more strongly correlated compared to distant ones. This implies that it could be a good practice, in vision problems, to take only a small portion of locally close pixels as input to each unit in the hidden layer, which is not the case with the standard neural networks that were covered earlier. All these potential problems are solved in a category of neural networks called *convolutional neural networks*. As it will be seen, these networks are especially suitable for vision problems since they exploit the properties of image inputs as discussed above. Convolutional networks were introduced by Fukushima (1980); however, without successful training. The first working implementation is by LeCun *et al.* (1998), also known as 'LeNet5' which will be covered here. There are three main changes in convolutional compared to standard[4] neural networks: shared weight matrices, local receptive field, and subsampling. Each of these important concepts will be separately discussed.

Convolutional networks are inspired by the human visual cortex and are very broadly used in computer vision problems. The input of such a network is an image, such that each of the pixels is taken as an input unit. The neurons in each hidden layer are organized in planes. For each of the neurons in a hidden layer, the pre-synaptic input takes only a local sub-area of the image - a local patch. Therefore, the dimensions of a weight matrix for one hidden unit correspond to the dimensions of that patch. All hidden units in one plane take different local patches (with same dimensions) from the image. This can be seen as having a 'sliding window' with dimensions corresponding to the patch, such that we shift that 'window' for each hidden neuron in a single plane. A very important fact in such a network is that the weights are shared[5] for neurons in one plane, which implies that each plane in the hidden layer will be capable of recognizing a single pattern in different parts of the image (i.e the network will be invariant to distortions and translations to some extent). As a consequence of the weight sharing, convolutional networks have considerably less parameters, hence they are easier and faster to train. Note however, the weight matrices are not shared for neurons in different planes. Because of that, each plane can recognize only one pattern/feature, so these planes are more frequently referred to as *feature maps*[6]. Since the weight matrices for hidden neurons in one feature map are the same, they can be seen as a *filter* which 'checks' if a certain patch contains the pattern that the feature map recognizes. Mathematically, the passing of an input patch through a filter is equivalent to convolution operation. The post-synaptic value of the neuron at position $(i, j)$ of the $k$-feature map denoted as $h^k_{ij}$, is

[4] By standard neural networks we refer to feed-forward, fully-connected neural networks as discussed in Section 2.1.2

[5] In other words, each neuron in one plane has the same set of weights as other neurons in that plane

[6] More specifically, the matrix consisting of the post-synaptic values (outputs) of the hidden neurons in one plane is called feature map.

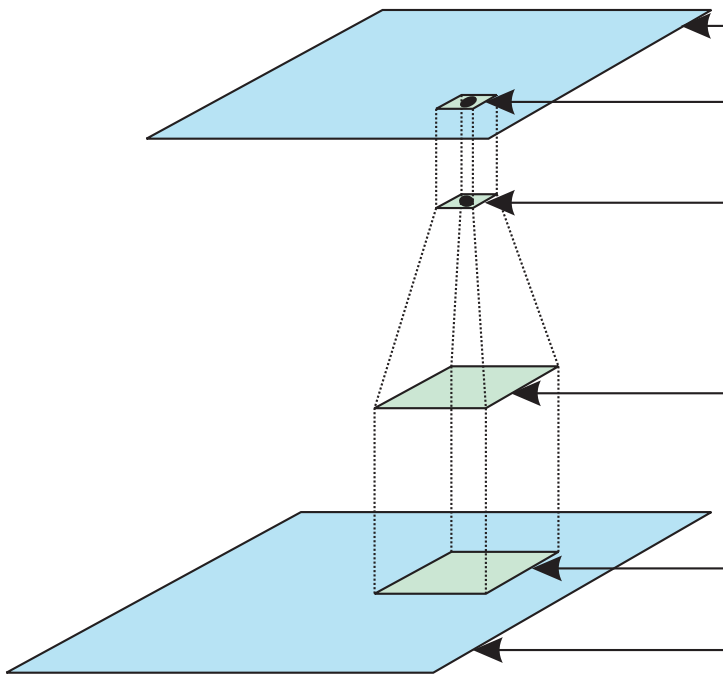

**Figure 2.4:** Convolution operation in convolutional networks

given by

$$h_{ij}^{k} = f\left(\mathbf{w^k} * \mathbf{x}\right), \tag{2.14}$$

where $f$ is the non-linearity applied , $b_k$ is the bias term of the $k$-th feature map, $\mathbf{w^k}$ is the weight matrix (filter) of the $k$-th feature map, $\mathbf{x}$ is the input image represented as a matrix of pixels, and $*$ represents two dimensional convolution operation[7], defined as

$$Conv^{2D}(a,b) = f(a,b) * g(a,b) = \sum_{u=0}^{a}\sum_{v=0}^{b}\left[f(u,v)g(u-a,v-b)\right]. \tag{2.15}$$

The convolution operation of input with filter for convolutional network can be seen visually in Figure 2.4.

In order to make the model even more resistant to distortions and translations we rely on another method called *subsampling*. Subsampling takes a small patch of a feature map and reduces it to one output by finding maximum or average of that patch (i.e. the outputs of neurons in that patch are processed and a single output is created). Through this operation, we basically reduce the resolution of the image, which consequently contributes to an even more invariant model in terms of distortions and translations. It is evident that with subsampling we lower the number of parameters of the network even more. Such operation is presented in Figure 2.5.

[7] Hence, these networks are called convolutional neural networks.

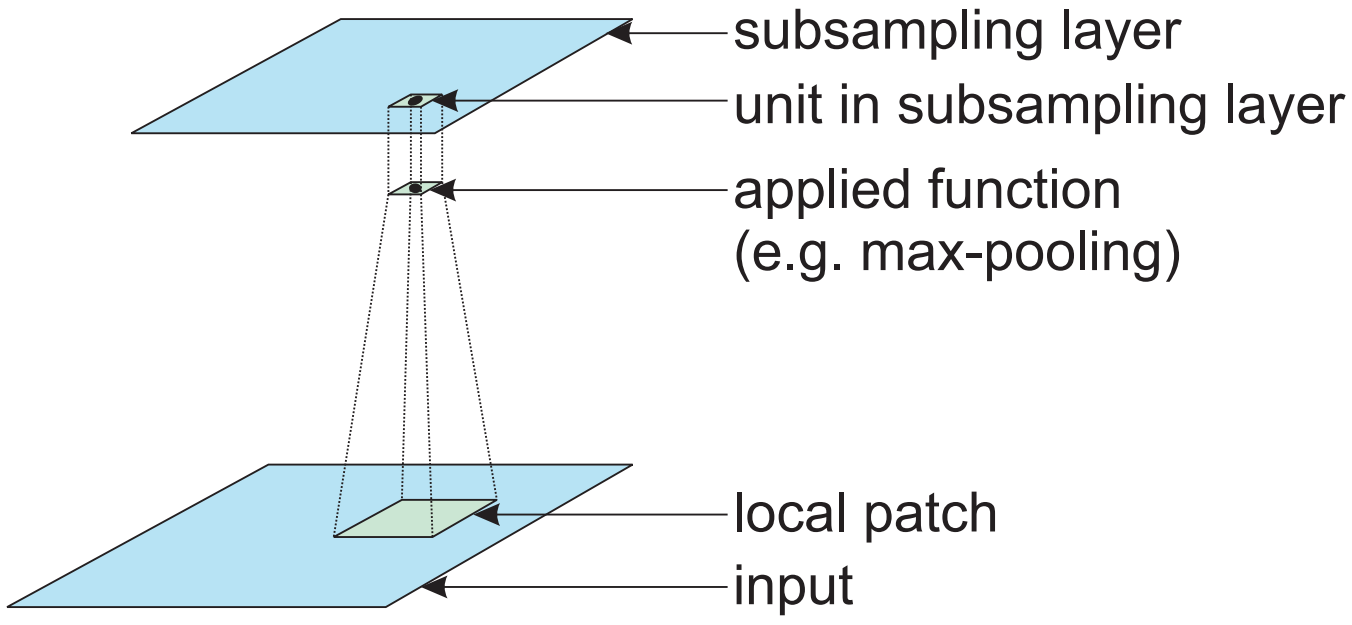

**Figure 2.5:** Subsampling step in convolutional networks

On top of the convolutional/subsampling structure we put fully connected layers (as in the standard neural networks). Although the number of fully connected layers depend on complexity of the problem, a frequent configuration is one to two fully connected hidden layers and an output layer on top of them. In case of classification, the output is put through a smoothing function, for example soft-max. Note that these fully-connected layers can result in a huge increment of the parameters of the network if the input planes have big dimensions. Therefore, a good practice is to build a fully connected layer on top of feature maps with smaller dimensions in order to reduce the number of parameters as much as possible.

A few other techniques are applied to improve the performance of convolutional networks even more. One such method is to purposely create distorted images and add them to the train set. This will enable the network to be even more invariant to such distortions in the input data. In addition, since the number of parameters is substantially lower compared to standard fully connected neural networks, it is a good practice to have trained the network on more epochs[8], which is possible since the training is faster. Other improvements include usage of contrast normalization layers, which subtract the mean and divide by the variance for each pixel (generally, it can be an output from any layer). The result of such a procedure is that the network does not take into account (small) variations in the color, which can be a huge problem in vision tasks.

[8] Epoch represents a single pass of the input images through the neural network

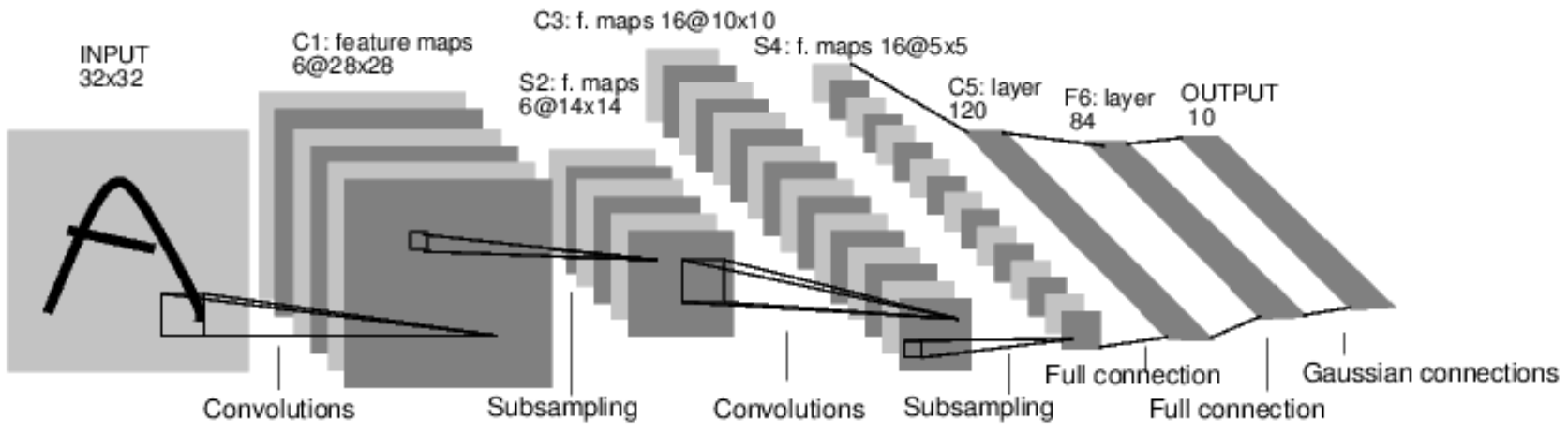

**Figure 2.6:** Architecture of the LeNet5 convolutional neural network

If we put more convolutional and subsampling layers in alternation order, the model will be able to recognize even more complex features, resulting in more powerful models. One might claim that features closer to the edge of the image are harder to be extracted (recognized). Such a statement is indeed true, since the pixels in the center of an image are included in more patches compared to the pixels near the edge of an image (LeCun *et al.* , 1998). This problem is solved by padding the image with extra 'dummy' pixels in all directions and centering the real input image. To end the discussion about convolutional networks, the LeNet5 network (LeCun *et al.* , 1998) is given in Figure 2.6[9]. The network depicted consists of two convolutional and two subsampling layers in alternating order. On top of the last subsampling layer, there are two fully connected layers and an output layer, which computes the Euclidean radial basis function[10] (RBF).

## 2.2 Reinforcement Learning

When a newborn baby tries to do certain actions, it does not know if that action is beneficial or possibly harmful. In order to acquire such information, it does not have positive and negative examples[11] to help it decide if that action is 'good' or 'bad'. Instead, it has senses (vision, smell, touch, pain, etc.) and needs to try out that action in order to decide if the action is good in that specific situation. If the action is touching a hot plate from a stove, it will feel pain and she will know that such action is 'bad'. Therefore, it would not attempt to do the same action in that same situation (in this example, when the plate is hot). Contrary to this, if the action is eating a chocolate bar, after executing that action, the baby would realize that such action is 'good' and it will repeat it in the future[12]. This is an example of how humans learn in a nutshell. However, it is very hard to represent a similar problem of learning using a supervised or unsupervised machine learning paradigm. Therefore, for such tasks we use an approach which is called *reinforcement learning*. Before defining the approach we will introduce a framework for it and the basic terms needed. For a more detailed theory about reinforcement learning, Sutton & Barto (1998) is recommended reading.

[9] Image taken from LeCun *et al.* (1998)
[10] For output $y_i$, this function is defined as $y_i = \sum_j (x_j - w_{ij})^2$
[11] Which is usually the case in the supervised learning approach.

For the purposes of this section, we shall assume an environment where an artificial agent tries to accomplish a predefined goal by taking actions from a given set of actions, and observing the results of how such actions change the environment. The configuration of the environment at any given time is called *state*, denoted as $s$, s.t. $s \in S$, where $S$ defines the state space. In order to explore the environment, the agent executes *actions* $a \in A$, where $A$ defines action space. By taking action $a$ from a given state $s$, the agent changes that configuration, therefore moves from state $s$ to state $s'$, where the transition is given by $s \xrightarrow{a} s'$. For simplicity, we shall assume a fully observable and stochastic environment, which means that the agent has access to the current state at any time and there is a probabilistic *transition model*. A probabilistic transition model $P : S \times A \times S \rightarrow [0, 1]$ defines the probability for going to state $s' \in S$ by taking action $a \in A$ from state $s \in S$. In each of the states the agent receives some positive or negative signal, called *reinforcement* or *reward signal*. The reward signal is defined by a *reward function*. This function map states and actions to numerical values and needs to be constructed to encode the goal of the problem. In order to accomplish its goal, the agent follows a *policy*, denoted as $\pi$. A policy determines which action should be performed from a given state, defined over the state space, i.e. it is mapping from states and actions to probabilities, $\pi : S \times A \rightarrow [0, 1]$, hence the probability of choosing action

[12] Subject to the assumption that the newborn baby discussed above likes chocolate.

$a$ from state $s$ is given by $\pi(s, a)$. By following an *optimal policy*, denoted as $\pi_*$, the agent maximizes the cumulative reward obtained from visiting a sequence of states. The cumulative reward, denoted as $R$, is a discounted sum of the rewards starting from a specific state $s$ and all consecutive states by following a policy $\pi$, computed as $R = \sum_{t=0}^{\infty} \gamma^t r_t$, where $\gamma$ represent the discount factor and $r_t$ is the reward at time step $t$. Now that we set up some kind of basic framework for reinforcement learning, we shall define this learning approach.

Reinforcement learning is a goal-directed machine learning approach, where states are mapped into actions in order to maximize some utility - numerical value given by the cumulative reward signal (Sutton & Barto, 1998). There are few key properties that make this approach different compared to other machine learning paradigms. First, exploration of the state space is encouraged, which means that in order for an agent to learn more about its environment, how specific actions from given states impact the environment, and which actions yield big rewards, it needs to explore the state space on its own, without given directives, which is not usually the case for other machine learning approaches. This is also called trial-and-error because the agent needs to discover which actions produce big rewards for specific states. Other key property is that the immediate reward does not mean that a certain action is ‘good’ or ‘bad’ since there may be some future actions which will drastically change the cumulative reward. Therefore, in order to make the decision as to whether executing an action from a given state is beneficial, the agent needs to take into account not only the immediate reward, but also future possible rewards. This is called delayed reward and makes the problem of reinforcement learning substantially harder. A solution for this problem is given by the value functions, which will be discussed shortly. Now that we have defined some basic aspects of the reinforcement learning approach, we will introduce formalism for such problems.

### 2.2.1 Markov Decision Process

One of the key characteristic of reinforcement learning problems, is that the environment has the Markov property, which means that we do not need to take into account previous states, actions and rewards, in order to get the next action, state and the reward of that state (Sutton & Barto, 1998). This implies that such learning tasks can be represented as *Markov Decision Process* (MDP), introduced in Puterman (1994). A Markov Decision Process is defined as

$$MDP = \{S, A, P, R\}, \tag{2.16}$$

where $S$ is a set of states (state space), $A$ is a set of actions (action space), $P$ are the transition probabilities and $R$ are the rewards. Lets define $S_t, A_t$ and $R_t$ as random variables that represent the state, action, and reward at time step $t$, respectively. The transition probabilities are defined as

$$P(s'|s,a) = Pr\{S_{t+1} = s'|S_t = s, A_t = a\}, \tag{2.17}$$

where $s, s' \in S$ and $a \in A$, such that by performing action $a$ from state $s$ the agent will go to state $s'$. The expected next reward is given by

$$R(s,a,s') = R(s,a) = \mathbb{E}[R_{t+1}|S_t = s, A_t = a]. \tag{2.18}$$

Section of MDP is given in Figure 2.7[13]. The solution to this MDP is given by the path $S0, S1, S2, ...$

In order to solve MDP, a policy that maximizes some utility function needs to be found. As it was said, the immediate reward is not a good indicator whether some action is desired. Therefore, this utility function should not be based solely on the immediate reward, but on the value functions which take into account future rewards as well.

[13] Retrieved from `http://artint.info/html/ArtInt_224.html` on 30.08.2014

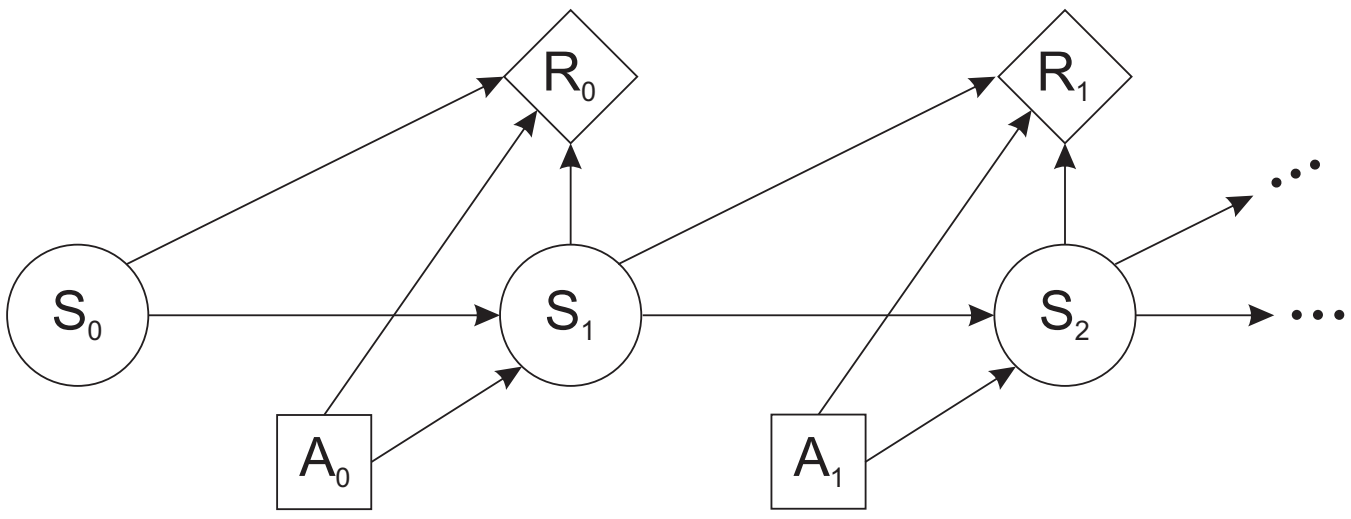

**Figure 2.7:** Decision network that represent a part of Markov Decision Process

### 2.2.2 State-Value Function

Let $\pi$ be a policy defined as $\pi(s,a) = Pr\{S_t = s, A_t = a\}$ s.t. $\forall a \in A, \forall s \in S$, which represents mapping from states and actions to transition probabilities. The value function for state $s$ under policy $\pi$ is defined as

$$V^\pi(s) = \mathbb{E}_\pi[\sum_{t=0}^{\infty} \gamma^t R_t | S_0 = s], \tag{2.19}$$

which is called *state-value function* for policy $\pi$, where $\gamma$ is a discount factor, $R_t$ is the reward at timestep t, and $\mathbb{E}_\pi$ is the expected value when the agent follows $\pi$. If we rewrite Equation 2.19 such that we separate the immediate reward given by $R_0$ from the future rewards, we get

$$V^\pi(s) = \mathbb{E}_\pi[R_0 | S_0 = s] + \mathbb{E}_\pi[\sum_{t=0}^{\infty} \gamma^t R_{t+1} | S_0 = s, S_1 = s'] \tag{2.20}$$

$$= \sum_a \pi(s,a) R(s,a) + \gamma \sum_a \pi(s,a) \sum_{s'} P(s'|s,a) V^\pi(s') \tag{2.21}$$

$$= \sum_a \pi(s,a)[R(s,a) + \gamma \sum_{s'} P(s'|s,a) V^\pi(s')]. \tag{2.22}$$

The intuition behind Equation 2.22 is that the value function for one state equals to the sum of the reward signal from that state and the future rewards starting from the consecutive state. Therefore, Equation 2.22 gives a connection between the value functions of two consecutive states. This is known as *Bellman equation* (Bellman, 2003) for the value function, according to which for state $s$ and successor state $s'$, if we know the value for state $s'$ and the dynamics[14] of the MDP, we can compute the value for state $s$. The notation $V^{\pi}$ can be interpreted as value function $V$ under policy $\pi$ (i.e. when policy $\pi$ is followed).

One might have noticed that when we introduced the basic terminology we stated that the policy represents a mapping from states to actions, while in this subsection we define it as mapping from states and actions to translation probabilities. The reason is that usually, we use greedy policy, that is, a policy where for a given state we choose the most probable action as desired. Therefore, the connection between those two definitions is given by

$$\pi(s) = \operatorname*{argmax}_a \pi(s,a) \quad \forall s \in S,\ a \in A. \tag{2.23}$$

From Equation 2.23 we can rewrite the Bellman equation given in Equation 2.22 as

$$V^{\pi}(s) = \max_a [R(s,a) + \gamma \sum_{s'} P(s'|s,a) V^{\pi}(s')]. \tag{2.24}$$

It should be evident that all possible policies are not equally good. We say that one policy is better than another if its expected cumulative reward is bigger compared to that reward of the other policy. The best policy[15] is also called *optimal policy*, defined as

$$\pi^*(s) = \operatorname*{argmax}_a \pi(s,a). \tag{2.25}$$

The value function defined in Equation 2.19 under optimal policy is called *optimal value function*

$$V^*(s) = V^{\pi_*}(s) = \max_{\pi} V^{\pi}(s) \quad \forall s \in S. \tag{2.26}$$

[14] The dynamics of the system represent the transition probabilities P, called the model of the environment
[15] Note that there may be more than one policy that yields the same highest expected cumulative reward

From Equation 2.24, we define the Bellman optimality equation for Equation 2.26 as

$$V^*(s) = \max_a [R(s,a) + \gamma \sum_{s'} P(s'|s,a) V^\pi(s')]. \tag{2.27}$$

### 2.2.3 Action-Value Function

In order to find optimal policy using the state-value function the agent will need to know (or approximate) the model of the environment because in order to solve the Bellman equation defined in Equation 2.24 for a given state, the transitional probabilities must be known. Now, we will define another value function which will not require a model to retrieve the policy. This function is called *action-value function*, denoted as $Q$, which maps states and actions to values. The $Q$ function relation to the $V$ function is given by

$$V^\pi(s) = \max_a Q^\pi(s,a) \quad \forall s \in S,\ \forall a \in A. \tag{2.28}$$

Taking into account Equation 2.28 and using Equation 2.19, we can define this function as

$$Q(s,a) = \mathbb{E}[\sum_{t=0}^{\infty} \gamma^t R_t | S_0 = s, A_0 = a]. \tag{2.29}$$

Analogously, the Bellman equation for the $Q$ function can be derived using Equation 2.24 as

$$Q(s,a) = R(s,a) + \gamma \sum_{s'} P(s'|s,a) \max_{a'} Q(s',a'). \tag{2.30}$$

From Equation 2.28 and Equation 2.30, the connection between both value functions is given by

$$Q(s,a) = R(s,a) + \gamma \sum_{s'} P(s'|s,a) V(s'). \tag{2.31}$$

### 2.2.4 Finding Optimal Policy

In order to solve the MDP and therefore solve the reinforcement learning problem, we need to find the optimal policy for that MDP. Note that most of the time finding the optimal policy is impossible, so we must settle for sub-optimal policies. The policy can be retrieved using the value functions as

$$\pi^*(s) = \underset{a}{\operatorname{argmax}} \sum_{s'} P(s'|s,a)V^*(s'), \tag{2.32}$$

$$\pi^*(s) = \underset{a}{\operatorname{argmax}}\, Q^*(s,a). \tag{2.33}$$

There are two approaches for learning policy: *adaptive dynamic programming* (ADP) and *temporal-difference* (TD). In the ADP methods, at every iteration, the value function is updated for each of the states. This update is given through Equation 2.27 or Equation 2.30. It is proven that the current estimate of the optimal policy will be closer to the real optimal policy after each iteration (Sutton & Barto, 1998). TD methods exploit the fact that two neighboring states are connected by the Bellman equations. Therefore, if we use the state-value function to retrieve the optimal policy, the update is given by

$$V^\pi(s) \leftarrow V^\pi(s) + \alpha \left[ \underbrace{\overbrace{R(s,a) + \gamma V^\pi(s')}^{V^\pi_{new}(s)} - V^\pi(s)}_{\Delta V^\pi(s)} \right], \tag{2.34}$$

where $V^\pi_{new}(s)$ is the new calculation of $V^\pi$ after state $s'$ is observed, $\alpha$ is a learning rate, and $\Delta V^\pi(s)$ represents the difference between the new and the old value of $V^\pi(s)$, also called temporal difference. The learning rate $\alpha$ dictates the impact of the new calculation, $V^\pi_{new}(s)$, when updating $V^\pi(s)$. This means that at each iteration, only the value functions of the consecutive states visited at that iteration are computed. As one might notice, for such an approach the underlying MDP does not need to be solved, which results in faster policy retrieval. The case when the action-value function is used to retrieve the policy in TD approach is called Q-Learning and will be discussed next.

#### 2.2.4.1 Q-Learning

*Q-Learning* is a model-free, off-policy, TD based method where the action-value function $Q$ is used to retrieve the optimal policy $\pi^*$, introduced by Watkins & Dayan (1992). This means that for such an approach, no explicit model of the environment P is needed and also no specific policy needs to be followed during the learning process. From the definition of the TD approach in Section 2.2.4, the update of the Q function can be written as

$$Q(s,a) \leftarrow Q(s,a) + \alpha \left[ \underbrace{\overbrace{R(s,a) + \gamma \max_{a'} Q(s',a')}^{Q_{new}(s,a)} - Q(s,a)}_{\Delta Q(s,a)} \right]. \tag{2.35}$$

From Equation 2.35 we can see that the new $Q$ value is equal to the sum of the old $Q$-value and the difference between the learned and the old value of $Q$, discounted by the learning rate $\alpha \in (0,1)$. If $\alpha$ takes a value close to 1, it will force the learned value as a new value for $Q$, without taking account of the old value for $Q$. A small value of $\alpha$ close to 0, will result in very slow learning, where the learned value has a very minor impact on the new value of $Q$. After we learn the $Q$-function we can retrieve the optimal policy by Equation 2.33.

$Q$-Learning is one of the most popular approaches for solving reinforcement learning problems as a consequence of the fact that the $Q$ function is learned completely independent from the policy. This property of $Q$-Learning contributes to a huge simplification of the reinforcement learning problem.

### 2.2.5 Parametrization

In scenarios where the state space is enormous, the convergence of the value function (be it action-value or state value function) will be very slow. Moreover, for such state space, the memory requirement for saving the computed values will be huge, which will result with not being able to save such data. Therefore, one solution is to approximate this function. One way for approximation is to *parametrize* this function i.e. the value function can be represented as a function of parameters $\theta$. The task then is to learn the parameters, in order to retrieve a value function near to the optimal one. Such an approximation of the Q function can be presented as

$$\hat{Q}_\theta(s,a) = g(\theta, f(s,a)) \quad s.t. \quad \theta \in \mathbb{R}^d, \tag{2.36}$$

where $g$ is some function (it can be both linear or non-linear) of the parameters $\theta$ and $f = f_1, .., f_{d-1}$ is a function of the states and actions, which represent a set of features or basis functions. One such parametrization where we apply logistic non-linearity would be

$$\hat{Q}_\theta = sigm(\theta^T f(s,a)) = sigm(\theta_0 + \theta_1 f_1(s,a) + \theta_1 f_1(s,a) + ... + \theta_{d-1} f_{d-1}(s,a)). \tag{2.37}$$

The connection between the approximated function and the optimal action-value function can be written as

$$\hat{Q}_{\theta^*}(s,a) \approx Q(s,a) \quad s.t. \forall s \in S, \ \forall a \in A, \tag{2.38}$$

where $\theta^*$ are the parameters that minimize loss function given by

$$L(Q_\theta) = \frac{1}{2} \sum_s \sum_a (\hat{Q}_\theta(s,a) - Q(s,a))^2. \tag{2.39}$$

We can rewrite Equation 2.39 as an optimization problem

$$\min_\theta L(Q_\theta). \tag{2.40}$$

In order to find the parameters $\theta_i$ for $i = 0, .., d-1$, that solve the optimization problem given in Equation 2.40, we use gradient descent method as discussed in Section 2.1.2. The update of parameter $\theta_i$ is given by

$$\theta_i \leftarrow \theta_i - \gamma \frac{\partial L(Q_\theta)}{\partial \theta_i}. \tag{2.41}$$

If we use a parameterized $Q$ function defined in Equation 2.36, the update in case of $Q$-Learning (defined in Equation 2.35) will be given by

$$\theta_i \leftarrow \theta_i + \alpha \left[ R(s,a) + \gamma \max_{a'} \hat{Q}_\theta(s',a') - \hat{Q}_\theta(s,a) \right] \frac{\partial \hat{Q}_\theta(s,a)}{\partial \theta_i}. \tag{2.42}$$

As some might think, we can use a neural network to find the parameters $\theta$ that minimize the loss (error) function given by Equation 2.39. In such a case, the weights of the neural network represent the parameters that need to be learned. Note that, the same approach applies to parametrization of the state-value fuction $V(s)$. We can use the same procedure to parametrize even the policy itself as $\pi_\theta$ if we try to directly find an estimate of the optimal policy.

The number of parameters will be much smaller than the number of states. As a consequence it will be much easier to learn those parameters than to learn the value function for the whole state space, which presents a huge reduction in the time needed for solving a reinforcement learning problem. Moreover, not only can such an approximation be learned fast, but it will also allow generalization to the unknown states. The intuition is that for a given unknown (not visited) state $s$, the value for the $V$ or $Q$ functions *can be approximated to other known states which are close to state $s$ in the parameters space*. This fact is fundamental for our work.

### 2.2.6 Exploration versus Exploitation

One of the biggest challenges in reinforcement learning tasks is the exploration vs. exploitation problem. In terms of reinforcement learning, using an *exploration* approach means to encourage the agent to go in states not seen before, in order to learn if those states provide big rewards. On the other hand, an *exploitation* approach directs the agent to visit states which are already visited and for which it has been learned that they yield big rewards. The problem is that some of the states that have big rewards may not be explored, so by strictly using an exploitation approach the agent cannot attain an optimal solution to the task. On the contrary however, it may be a case where many of the unseen states provide small rewards, and by forcing strict exploration the agent will collect insignificant rewards from the unexplored states instead of visiting states for which it is known that they provide significant rewards. Therefore, we must find some balance between both approaches in order to obtain optimal results. Such a balance cannot be generally defined, since it depends on the problem itself. We should know that initializing the value function with greater values will stimulate more exploration and vice versa. Another technique for controlling this balance is by a probability factor $\epsilon$. Such policy is called $\epsilon$-greedy policy. It means that an agent following that policy will perform the most beneficial action (will be greedy) with probability $1 - \epsilon$, but it will execute random action with probability $\epsilon$. Although using $\epsilon$-greedy policy will propagate exploration, the convergence time will be very long.

## 2.3 Apprenticeship Learning

The reinforcement learning approach offers a well defined framework for specifying and solving learning problems. However, this approach is not well suited for all types of problems. Assume for example, an agent, a very complex one, whose task is to obtain a DPhil in Computer Science. Some of the actions that this agent can perform include reading papers, writing reports, conducting research, and discussions with the supervisor, among others. Although it is easy to define that procrastination, for example, is not beneficial, and conducting research is probably helpful for attaining the final goal, it is very hard to rate how important each of the possible factors[16] related to this task can be, which means that it is very difficult, or almost impossible to construct a reward function for this problem. Some might argue that this is a very complex problem, so we will give one more example as presented in Abbeel & Ng (2004). Assume an agent who has a task to learn how to drive a car. There are many variables in driving car, such as driving speed, distance between other cars, amount of gas, whether there are pedestrians on the road, and so on. It is hard to define which of the variables are more important for successful driving and to what extent. Once again, constructing a reward function for a problem such as driving is almost impossible. Since, the availability of a reward function is a fundamental assumption in reinforcement learning, we rely on other approaches for solving such problems called apprenticeship learning.

*Apprenticeship learning* is a learning approach, which assumes that the reward function is unknown, but instead, expert knowledge is available. This expert knowledge is in the form of sequences of states and actions, trajectories, where the goal state is achieved. The idea is to use these trajectories to retrieve the policy used to create them in the first place, called *expert policy*. Note that the retrieved policy does not need to be the expert policy, so we settle for close approximation as well. There are two common approaches to retrieve the expert policy - imitation learning and inverse reinforcement learning - which will be discussed shortly.

Indeed, if we contemplate the driving example, we can conclude that the process of learning to drive starts with an expert (parent or instructor) showing the student how to drive. Therefore this approach is more natural for certain problems. We must agree that many learning tasks of humans can be seen as apprenticeship learning tasks, one example being a child who imitates what his/her parents do and how they behave.

[16] The factors for this problem includes the number of hours spent on conducting research, the number of published papers, the psychological and physical state etc.

### 2.3.1 Imitation Learning

Imagine a game-playing environment, where the problem is defined as in reinforcement learning, but with one important difference: there is no reward function available. Instead, we are given a set of expert trajectories, samples of game-plays, explaining the expert behaviour for that particular game. It is obvious that this task belongs to the apprenticeship learning paradigm, where we need to retrieve the expert policy given the expert trajectories. For this instance, we shall assume a deterministic game, where the environment does not change, the starting state is same for different tries, and the dynamics of the game are fixed. In such a case we can use a standard classifier in order to directly infer the policy. This approach is called *imitation learning* and is one of the methods for doing apprenticeship learning.

In such a case, the expert behaviour is given as a set of trajectories created following the expert policy, $D = \{s_0^{(i)}, a_0^{(i)}, s_1^{(i)}, ...\}_{i=0}^{N}$, where $N$ is the number of trajectories. For more convenience, we will merge those trajectories into one dataset consisting of state-action pairs $D' = \{(s_t, a_t)\}_{t=0}^{T}$. The policy $\pi$ is parametrized with parameters $\theta \in \mathbb{R}^d$, denoted as $\pi_\theta$. These parameters need to be optimized in order to minimize a loss function. Neu & Szepesvári (2012) define this loss function as

$$J(\pi_\theta) = \sum_{s \in S} \sum_{a \in A} \hat{\psi}(s)(\pi_\theta(a|s) - \hat{\pi}_E(a|s))^2, \tag{2.43}$$

where $\hat{\psi}$ and $\hat{\pi}_E$ are the empirical occupation frequencies under the expert policy and the empirical estimate of that policy, respectively, defined as

$$\hat{\psi}(s) = \frac{1}{T+1} \sum_{t=0}^{T} \mathbb{I}\{S_t = s\}, \tag{2.44}$$

$$\hat{\pi}_E(a|s) = \frac{\sum_{t=0}^{T} \mathbb{I}\{S_t = s, A_t = a\}}{\sum_{t=0}^{T} \mathbb{I}\{S_t = s\}}. \tag{2.45}$$

The intuition behind the loss function given by Equation 2.43 is that by minimizing it, we minimize the squared difference between the found policy and the expert policy. Therefore, when we find the optimal estimate of the parameters $\hat{\theta}$, from Equation 2.43, it follows that $J(\pi_{\hat{\theta}}) \approx J(\pi_E)$, which means that the learned policy $\pi_{\hat{\theta}}$ will be an approximation of the expert policy $\pi_E$. As a result, the agent will learn how to mimic the behaviour of the expert, which is useful when we have a simple problem, with relatively small state space, and expert trajectories that cover most of those states. However, if we drop one of the assumptions made for the example above, such an approach will not work well. In that case the agent will learn how to copy the expert, but will fail to generalize on the expert's knowledge and use it, for example, when the starting state is different from the one of the expert. For that reason we can also refer to this approach as *behaviour cloning*. Since some of the states will not be covered by the expert trajectories, for those states we cannot compute estimates of the frequencies $\psi$ as in Equation 2.44.

### 2.3.2 Inverse Reinforcement Learning (IRL)

*Inverse reinforcement learning* is an approach for attacking the apprenticeship learning problem, first introduced by Ng & Russell (2000). Since the reward function is "*the most succinct, robust, and transferable definition of a task*" (Abbeel & Ng, 2004), it seems reasonable to learn this function in order to retrieve the expert policy. Such an approach was proposed by Abbeel & Ng (2004). They define the reward function as a linear combination of predefined, known features that describe the problem which needs to be solved. The idea is to learn the importance of each of the features, so that the policy which is optimal for reward function of those features, will be (near) optimal to the expert policy. Indeed, it is much easier to define which features are important for solving a problem than giving an explicit reward function based on those features. The reward function is parameterized as

$$R(s) = \boldsymbol{w}^{\boldsymbol{T}} \boldsymbol{\phi}(\boldsymbol{s}) \quad s.t. \quad \boldsymbol{\phi} : S \to [0, 1]^k, \tag{2.46}$$

where $\boldsymbol{\phi}$ is the features vector, $\boldsymbol{w} \in \mathbb{R}^k$ is a weight vector which need to be learned, and $k$ is the number of features. The features vector is predefined to reflect our knowledge of the problem. Another vector called feature expectations, denoted as $\mu$, is defined as

$$\boldsymbol{\mu}(\boldsymbol{\pi}) = E[\sum_{t=0}^{\infty} \gamma^t \boldsymbol{\phi}(\boldsymbol{s_t})|\pi] \in \mathbb{R}^k. \tag{2.47}$$

We iteratively compute Equation 2.47 for the current policy (starting by an arbitrary chosen policy) and compare it to the feature expectations for the expert policy $\boldsymbol{\mu_E} = \boldsymbol{\mu}(\boldsymbol{\pi_E})$. In order to do so, we need to find the estimate of $\boldsymbol{\mu_E}$, which is computed as

$$\hat{\boldsymbol{\mu}}_{\boldsymbol{E}} = \frac{1}{m} \sum_{i=0}^{m} \sum_{t=0}^{\infty} \gamma^t \boldsymbol{\phi}(\boldsymbol{s}_{\boldsymbol{t}}^{(i)}), \tag{2.48}$$

where $\{s_0^{(i)}, s_1^{(i)}, ...\}_{i=1}^m$ represent the set of expert trajectories. At each iteration, the weights vector $\boldsymbol{w}$ is optimized in order to attain $\boldsymbol{\mu}$ which is as close as possible to $\boldsymbol{\mu_E}$. Next, the reward function is recomputed as defined in Equation 2.46, followed by the retrieval of the policy using a standard reinforcement learning method. This procedure is repeated until the found policy is close enough to the expert policy. The optimization of the weights vector can be done by using max margin or projection methods. Those methods are out of the scope of this dissertation, but more on them can be found in (Abbeel & Ng, 2004). Although this approach does not have a problem to generalize on the expert knowledge, which was not the case for imitation learning, there are some other potential problems. Neu & Szepesvári (2012) claim that for successful retrieval of the reward function the features need to be properly scaled, which is a non-trivial task. Furthermore, the policy learned cannot outperform the expert policy. Syed & Schapire (2007) provide a small change in the original method in order to allow learning a reward function, for which the extracted policy can be even better compared to the expert policy. Many approaches were suggested for solving the inverse reinforcement learning problem (for example, see Syed & Schapire, 2007), but here we will limit our discussion only to those that are in some way connected to our method.

#### 2.3.2.1 Supervised Learning Based Approach to IRL

One possible way of solving the IRL problem is to use a supervised approach in order to retrieve the reward function given the expert trajectories. However, without having any information on the reward function it is impossible to retrieve it directly. Therefore, Klein *et al.* (2013) propose dividing this problem into two tasks. Firstly, using any classification method a score function $q$ is learned. Secondly, regression is used to retrieve a reward function based on the learned $q$-function.

The score function, which is defined over states and actions, rates how much given action is desired to be executed from a specific state based on the expert trajectories. Let $D_c = \{s_t, a_t = \pi_E(s_t)\}_{t=0}^{T}$ denote a set of state-action pairs obtained from the expert trajectories. We shall use $D_c$ as input to any classifier in order to learn the $q$-function $q : S, A \to \mathbb{R}$. In order to obtain a policy $\pi_C$ given the learned score function, we use

$$\pi_C(s) = \operatorname*{argmax}_a q(s, a). \tag{2.49}$$

As we can see, Equation 2.49 is very similar to Equation 2.33. Therefore, we can say that the score function $q$ represents the optimal action-value function for the policy $\pi_C$. This means that instead of solving MDP to find the optimal policy $\pi_*$, we can find the policy $\pi_C$ using the classification approach, which is much faster.

We shall assume that there exists a function $R_C$, that is a reward function for which the classification policy $\pi_C$ is optimal. The connection between the reward function, the optimal action-value function, and the optimal policy, is given by the Bellman equality (Equation 2.30). Therefore, based on Equation 2.30 and taking into account the fact that the score function represents an optimal action-value function for the classification policy, we can rewrite the Bellman equation as

$$q_{\pi_C}(s,a) = R_C(s,a) + \gamma \sum_{s'} P(s'|s,a) q_{\pi_C}(s', \pi_C(s')). \tag{2.50}$$

Next, using basic arithmetic operations on Equation 2.50, we can define the reward function $R_C$ as

$$R_C(s,a) = q_{\pi_C}(s,a) - \gamma \sum_{s'} P(s'|s,a) q_{\pi_C}(s', \pi_C(s')). \tag{2.51}$$

Since the score function $q$ is learned, for estimating $R_C$, only the dynamics of the environment $P$ need to be learned. In order to do so, we can use samples acquired from any policy. Klein *et al.* (2013) define another dataset $D_R = \{(s_n, a_n, a'_n\}_{n=0}^{N}$, which is created by sampling random play, since the interest is not in the underlying policy, but in the dynamics of the system. Next, another dataset $D_F = \{(s_n, a_n), \hat{r}_n\}_{n=0}^{N}$ is created, where $\hat{r}_n$ represent the estimate of the reward $R(s_n, a_n)$ defined as

$$\hat{r}_n = q(s_n, a_n) - \gamma q(s'_n, \pi_c(s'_n)) \quad s.t. \quad (s_n, a_n, s'_n) \in D_R. \tag{2.52}$$

Finally, a regression step is applied on $D_F$ as input, in order to retrieve the reward function $R_C$. Since $R_C$ is the reward function for which the policy $\pi_C$ is optimal, and $\pi_C$ is an approximation of the expert policy $\pi_E$, it follows that $R_C$ will be an approximation to the reward function $R_E$[17], subject to the existence of such a function. Note that we do not use the classification policy $\pi_C$ directly as a solution, since such an approach will be imitation learning as discussed in Section 2.3.1, accompanied by all the flaws in those methods. By obtaining the underlying reward function, we can obtain a generalization of the policy on states not visited. The whole procedure is called *cascaded supervised inverse reinforcement learning*, which is given in Algorithm 2.

According to Klein *et al.* (2013), one of the biggest advantages of this approach is that

[17] $R_E$ is reward function for which the policy $\pi_E$ is optimal

**Algorithm 2** Cascaded Supervised Inverse Reinforcement Learning (Klein *et al.* , 2013)

***Given***

Dataset of expert trajectories $D_C$ defined as $D_C = \{(s_t, a_t = \pi_E(s_t))\}_{t=1}^{T}$,

Dataset of random play $D_R$, defined as $D_R = \{(s_i, a_i, s_i')\}_{i=1}^{N}$.

***Train*** a score function-based classifier on $D_C$ to obtain:

- score function $q$, $q : S, A \to \mathbb{R}$,
- decision policy $\pi_C$ using $\pi_C(s) = \text{argmax}_a\, q(s,a) \quad s.t. \quad s \in S, a \in A$.

***Create*** a dataset $D_F = \{(s_n, a_n), \hat{r}_n\}_{n=0}^{N}$,

- $\hat{r}_n = q(s_n, a_n) - \gamma q(s_n', \pi_c(s_n'))$,
- $(s_n, a_n, s_n') \in D_R$.

***Learn*** a reward function $\hat{R}_C$ using regression method on $D_F$.

***Output*** the reward function $\hat{R}_C$ .

the MDP which represents the problem is not solved, compared to other approaches (for example, Neu & Szepesvári, 2012), that solve the underlying MDP on every iteration, which can be time-consuming. The idea is that the constraints enforced by the MDP can be introduced using the Bellman equality. Other advantages include the fact that we do not need to know almost anything about the underlying reward function, which means that a problem such as scaling features encountered in other approaches (for example, Abbeel & Ng, 2004) is not an issue here. Because of the regression step, this method can infer a good approximation of the reward function using less expert trajectories compared to other approaches.

#### 2.3.2.2 Gradient Based Approach to IRL

Another way of solving the IRL problem is by parameterizing the reward function instead of the policy. Then we try to find estimates of the parameters such that the policy which is optimal to this optimized reward function is in close approximation of the expert policy. Such an approach, called *gradient based method for inverse reinforcement learning*, is presented by Neu & Szepesvári (2012).

We shall assume a family of reward functions $r_\theta \in \Theta$. The idea is to find the policy $\pi_\theta$ that is optimal for the reward function $r_\theta$ and it is in close approximation to the expert policy i.e.

$\pi_\theta \approx \pi_E$. The optimization problem is given by

$$\min_{\theta} J(\pi_\theta), \tag{2.53}$$

where $J(\pi_\theta)$ is a loss function as defined in Equation 2.43 and the policy is defined as

$$\pi_\theta(a|s) = S(Q_\theta(s,a)) = \frac{\exp \beta Q(s,a)}{\sum_{a'} \exp \beta Q(s,a')}, \tag{2.54}$$

where $S$ is a soft-max function[18] and $\beta$ is a parameter that dictates how greedy the policy will be in respect to $Q$. Policy as defined in Equation 2.54 is called Boltzmann action-state policy. This policy is near-greedy to the action-state function $Q$, according to its parameter $\beta$. The reason why such a policy is used in this approach, is that if a strict greedy policy is employed the policy would not be differentiable in respect to the parameters $\theta$, and therefore the respective gradients could not be calculated.

In order to find the gradient of the loss function $J(\pi_\theta)$ in respect to the parameters $\theta$, we need to derive the gradient of $\pi_\theta$ in respect to $\theta$. Since $\pi_\theta$ is function of $Q^*$ as defined in Equation 2.54, for such calculation we shall calculate the gradient of $Q^*$ in respect to the parameters. To calculate that gradient we will first derive the gradient of the parametrized reward function $r_\theta$ in terms of its parameters. Therefore, to compute the gradient of the loss function we shall propagate it to the policy followed by the action-value function, followed by the reward function. We will discuss computing the gradient of the parameters vector $\theta$ by deriving the gradients of each component of $\theta$ (parameters) given by $\theta_i$ such that $\theta = \{\theta_i\}_{i=0}^{d}$, where $d$ is the number of parameters i.e. $\theta \in \mathbb{R}^d$. The first gradient $\frac{\partial r_\theta}{\partial \theta_i}$ is trivial to be computed since $r_\theta = f(\theta)$. In order to calculate the gradient $\frac{\partial Q^*}{\theta_i}$ we need to solve the fixed point equation given by $\phi_\theta(s,a) = \frac{\partial BE(s,a)}{\partial \theta}$, where $BE$ is Bellman operator

[18] $S$ can be any smoothing function. We define it as soft-max based on the approach of Neu & Szepesvári (2012)

**Algorithm 3** Gradient Based Method for IRL (Neu & Szepesvári, 2012)

***Given***

Dataset of expert trajectories $D_E$ defined as $D_E = \{(s_t, a_t = \pi_E(s_t))\}_{t=1}^T$.
Parameters $\theta \in \mathbb{R}^d$, initialized by random values,
Small error value $\epsilon$,
Parameterized reward function $r_\theta$, defined as $r_\theta = f(\theta)$.

**while** $J(\pi_\theta) = \sum_s \sum_a \hat{\psi}(s)(\pi_\theta(a|s) - \hat{\pi}_E(a|s))^2 > \epsilon$ **do**
  Compute the gradients of $r_\theta$ in respect to $\theta_i$ s.t. $i = 1, .., d$
  Compute the gradients of $Q^*_\theta$ in respect to $\theta_i$ using the fixed point equation (Eq. 2.56)
  Compute the gradients of $\pi_\theta$ in respect to $\theta_i$ where $\pi_\theta$ is defined as in Eq. 2.54
  Compute the gradients of $J(\pi_\theta)$ in respect to $\theta_i$ where $J(\pi_\theta)$ is defined as in Eq. 2.43
  Update the parameters $\theta_i$ s.t. $i = 1, .., d$ using the update rule $\theta_i \leftarrow \theta_i - \eta \frac{\partial J(\pi_\theta)}{\partial \theta_i}$
**end while**
**Output** $\pi_\theta$

(see Equation 2.30) defined over states and actions as

$$BE(s, a) = r_\theta(s) + \gamma \sum_{s'} P(s'|s, a) \max_{a'} Q(s', a'). \tag{2.55}$$

From Equation 2.55 we can rewrite the fixed point equation $\phi_{\theta_i}(s, a) = \frac{\partial BE(s,a)}{\partial \theta_i}$ as

$$\phi_{\theta_i}(s, a) = \frac{\partial r_\theta}{\partial \theta_i} + \gamma \sum_{s'} P(s'|s, a) \sum_{a'} \pi(a'|s')\phi_{\theta_i}(s', a'). \tag{2.56}$$

We will not delve into more detail about the fixed points theorem. It is important to know that when we fix $\phi_{\theta_i}(s', a')$ we can compute $\phi_{\theta_i}(s, a)$. By iteratively repeating this procedure it has been proved that the $\phi_{\theta_i}(s, a)$ will converge toward $\frac{\partial Q^*(s,a)}{\partial \theta}$ (Sniedovich, 2010). Moreover, Neu & Szepesvári (2012) use natural gradient in order to force a unique solution for the parameters $\theta$. Discussion about natural gradient is out of the scope of this dissertation (see Amari, 1998) The whole procedure is given in Algorithm 3.

## 2.4 The Big Picture of Reinforcement and Apprenticeship Learning

Next, we will summarize all of the approaches introduced in this chapter and we will present how reinforcement and apprenticeship learning are connected. In a nutshell, the reinforcement learning approach is based on a known reward function which needs to reflect the goal that should be obtained. Using this function, an optimal policy is retrieved. That policy solves the MDP which formally describes the task. On the other hand, apprenticeship learning assumes that such a reward function is not known. Instead, an expert behaviour is given in advance in the form of set of trajectories created by following an optimal (or near optimal) expert policy, which is not known. The idea is to try to retrieve this expert policy, or a close approximation of it. There are two possible ways for retrieving the policy. One approach is to directly infer the policy based on a classification method used on the expert trajectories. This method is called imitation learning, due to the fact that for most tasks, the agent which follows such policy will only know how to mimic the expert's behaviour, without successful generalization of unknown (not visited) states. An even better approach is to learn a reward function for which some policy is optimal. In order to retrieve that policy, standard reinforcement learning algorithms are used. For this task, called inverse reinforcement learning, different techniques may be employed, such as gradient methods, classification and regression methods (or their combination), maximum entropy methods, maximum margin methods, and projection methods. A high-level overview of the possible approaches to do apprenticeship learning is given in Figure 2.8. Here we can also see how reinforcement learning is connected to the apprenticeship learning paradigm. Note that, there are two possible ways to arrive to "Policy" when starting from "Expert's Demonstrations". These correspond to the two approaches for doing apprenticeship learning, discussed above.

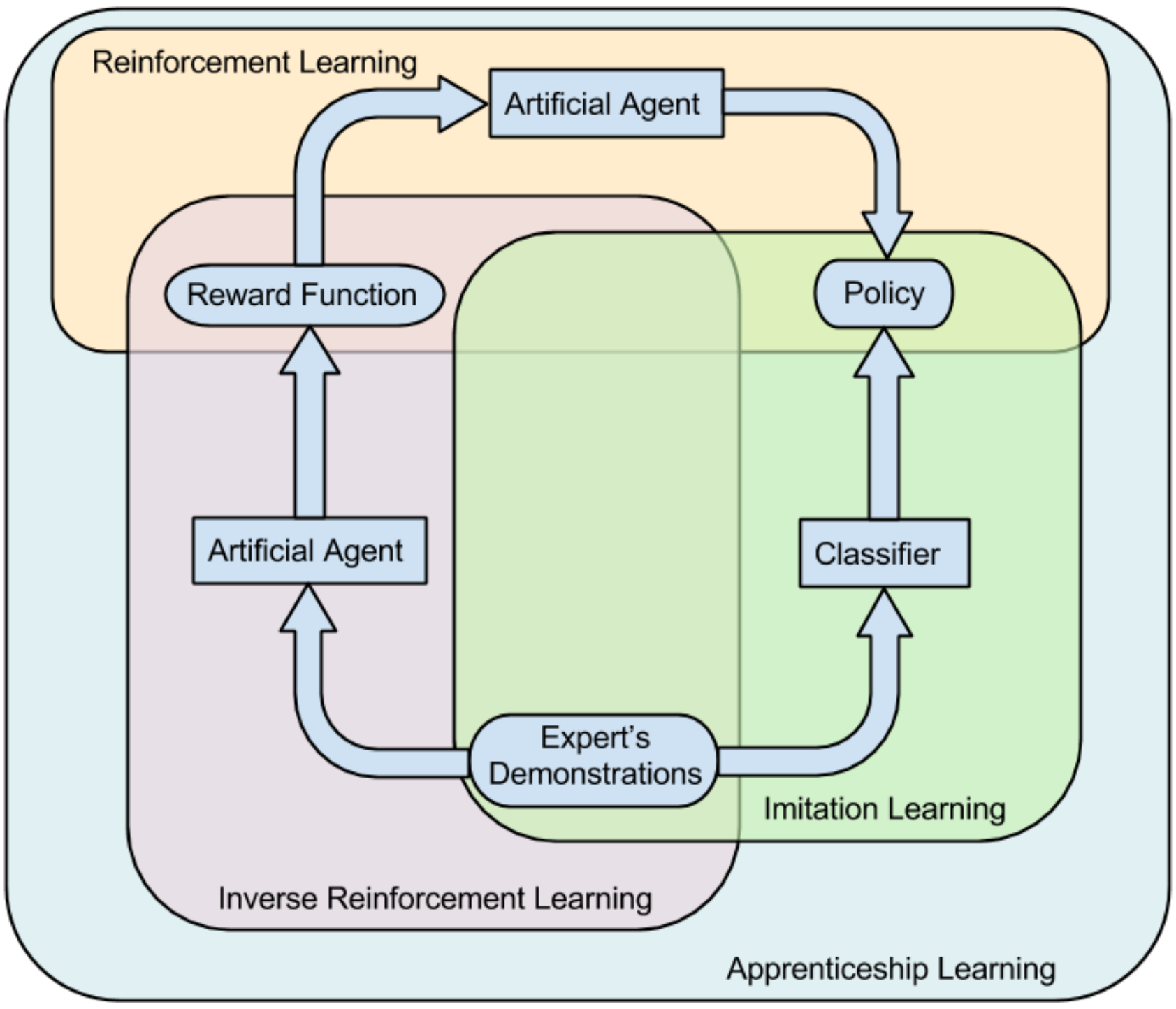

**Figure 2.8:** Summary of Approaches for Doing Apprenticeship Learning

## 2.5 Deep Learning

As we already mentioned in Section 2.1.3, higher layers of neural networks learn more abstract features. Therefore, for complex problems where the input is multidimensional, such as for example images, videos or speech sequences, ordinary machine learning models are not powerful enough to fit the data. In such cases we can use neural networks consisting of more layers in order to recognize more abstract features. Neural networks that have several hidden layers are called *deep neural networks* (Bengio & LeCun, 2007). The paradigm of building more complex models is called *deep learning*. It needs to be mentioned that deep learning as an approach is used with models other than neural networks as well. However, it is used most frequently for neural networks, in addition to the fact that the majority of successful applications and scientifically important advances in this area include deep neural network models. Therefore, in this dissertation, by deep models we will refer exclusively to deep neural networks.

Historically, deep neural networks date back from more than thirty years ago (Fukushima, 1980). However, since there had been many problems pertaining to the training of such networks, the first working approach is less than a decade old (Hinton *et al.* , 2006). In this approach, the authors propose an unsupervised step before the training phase for deep belief networks, called pre-training. They suggest iteratively building the network starting from one hidden layer, which is seen as a restricted Boltzmann machine (Smolensky, 1986). Then, after the weights for that layer are learned i.e. this simple network is trained, a new hidden layer is added on top of the old, and the weights for the new layer are learned while the weights for the old layer are fixed. This procedure continues by iteratively stacking layers until the whole network is built. At the end, backpropagation is used for fine tuning of the network.

One of the biggest challenges in training deep network was the so called "vanishing gradient" problem. In a nutshell, since the network has possibly many layers, when propagating the gradients, their values approach zero very fast, resulting with convergence of the parameters to not optimal values. Traditionally the logistic function, given in Equation 2.1, was used as non-linearity. The problem with this function is that it takes values between $[0, 1]$, which means that the difference between the gradients for extreme values will be very small. This contributes to the "vanishing gradient" problem, especially when the networks consist of more hidden layers. In recent years, this problem is solved by using a rectifier as an activation function. With their approach, Hinton *et al.* (2006) solve this issue. Other problem is the fact that deep networks have many (sometimes thousands of millions) parameters, which results in very slow training. In the last few decades the processing power has increased dramatically, shortening the training time to an acceptable duration. Especially important, in terms of performance, is the advancement of graphical processing architectures (GPUs).

In the last few years deep learning methods produced very promising, state-of-the-art results in many areas, such as vision (Krizhevsky *et al.* , 2012; Farabet *et al.* , 2013; Eigen *et al.* , 2013; Sermanet *et al.* , 2013), speech recognition (Hinton *et al.* , 2012; Graves *et al.* , 2013; Deng *et al.* , 2013), and natural language processing (Huang *et al.* , 2012; Socher *et al.* , 2013; Grefenstette *et al.* , 2014) among others. Despite its popularity, there is little research on using deep learning for solving reinforcement learning problems, although it may seem as a reasonable idea. The only such application is by Google's DeepMind[19] (Mnih *et al.* , 2013). Their approach employs a gradient based variant of Q-learning where a deep convolutional neural network is used. We will discuss this research shortly. To our knowledge, there is no research on solving apprenticeship learning problems, by using deep neural networks. Therefore, our work presents the first approach in that direction, which we will call *deep apprenticeship learning*, for obvious reasons.

### 2.5.1 Deep Reinforcement Learning

In this section, we will discuss the work of Mnih *et al.* (2013). Briefly, they aim to train an agent to play a game, given an appropriate reward function and video of the gameplay. One of the most important contributions of their work is that they show that "sparse, noisy, and often delayed reward signals" (Mnih *et al.* , 2013) can be successfully used in order to train a deep convolutional neural network which learns the Q-function in reinforcement learning problems when the input is multidimensional. This method is called *deep reinforcement learning*. Although the learning approach is different than ours, the configuration of the problem is very similar and the application is almost identical. Therefore we will discuss not only the method used, but also the structure of their network and the domain in which it is applied.

Their goal is to teach an agent how to play Atari 2600 games only by having raw video frames from the game play, known reward and termination signals, and the set of possible actions for that game. The agent does not have any information about the internal state of the game. Instead, it communicates with an emulator in order to find the game dynamics and which states produce bigger rewards. Note however, that this problem is not fully observable, which means that the environment does not have the Markov property, and therefore the Bellman equation given in Equation 2.30 does not hold up. Therefore instead of taking only one game state as a state in the MDP for the task, they take a finite sequence of game states (video frames) and actions. Such sequence will be denoted as $s = x_1, a_1, ...x_T$, where $x_i$ is a video frame at time $i$, $a_i$ is the action performed at that time, and $T$ is the termination time of the sequence[20].

[19] `www.deepmind.com`

The goal is to infer an optimal (or near optimal) Q-function as defined in Equation 2.30 in order to retrieve the optimal policy. Since for each of the sequences the Q-function will be different without achieving any generalization, a slightly changed variation of the Q-Learning method presented in Section 2.2.4.1 is used. Instead of using the Bellman optimality equation given by Equation 2.35 as iterative update for the Q-function, they use a neural network in order to approximate $Q_\theta(s, a) \approx Q_*(s, a)$, where $\theta$ are the weights (parameters) of the neural network. They call this network Q-network, which is basically a deep convolutional neural network so it can handle complex video input. The loss function $J_i(\theta_i)$ at iteration $i$ is given as

$$J_i(\theta_i) = \mathbb{E}_{(s,a)\sim p}[(y_i - Q_{\theta_i}(s, a))^2], \tag{2.57}$$

where $p$ represents the dynamics of the system $p : S, A \rightarrow [0, 1]$, $\varepsilon$ is the emulator, and $y_i$ is the target value for the Q-function, defined as

$$y_i = \mathbb{E}_{s'\sim\varepsilon}[R(s, a, s') + \gamma \max_{a'} Q_{\theta_{i-1}}(s', a'|s, a)]. \tag{2.58}$$

[20] Note the change of notation. Such notation will be used for the rest of this section.

Mnih *et al.* (2013) suggest that there are few challenges when doing reinforcement learning using a deep learning approach. One important issue is that deep learning models assume that consecutive input samples are not correlated, which is not true for reinforcement learning, since near states are highly associated. They solve this problem by using a method called experience replay, where on each iteration, a constant number of states are sampled randomly from the previously available states, and used for training. Moreover, they break the correlation between consecutive states of a sequence by randomizing the order of the state-action pairs in that given sequence. The full algorithm is given in Algorithm 4, taken from Mnih *et al.* (2013). We will not go into more details in the algorithm, since it is mainly included because of our discussion about short-term future direction for work on our project in Chapter 5, and not as a base for our approach.

In terms of the structure, they base the network on Krizhevsky *et al.* (2012). Their network, called Deep Q-Network (DQN), is a convolutional neural network which consists of three hidden layers. The input represents four images (video frames) of the game, with dimensions of $84 \times 84$ pixels. The first hidden layer is a convolutional layer with 16 feature maps, $8 \times 8$ filters with stride $s = 4$. The second layer is a convolutional layer with 32 feature maps, $4 \times 4$ filters with stride $s = 2$. The third layer is a fully connected layer with 256 hidden units. Finally, the output layer is fully connected with units corresponding to the number of possible actions in the game. They use the same structure and approach for seven games, out of which state-of-the-art results are achieved in six of them.

**Algorithm 4** Deep Q-learning with Experience Replay (Mnih *et al.* , 2013)

Initialize replay memory $\mathcal{D}$ to capacity $N$
Initialize action-value function $Q$ with random weights
**for** episode $= 1, M$ **do**
  Initialize sequence $s_1 = \{x_1\}$ and preprocessed sequenced $\phi_1 = \phi(s_1)$
  **for** $t = 1, T$ **do**
    With probability $\epsilon$ select a random action $a_t$
    otherwise select $a_t = \max_a Q^*(\phi(s_t), a; \theta)$
    Execute action $a_t$ in emulator and observe reward $r_t$ and image $x_{t+1}$
    Set $s_{t+1} = s_t, a_t, x_{t+1}$ and preprocess $\phi_{t+1} = \phi(s_{t+1})$
    Store transition $(\phi_t, a_t, r_t, \phi_{t+1})$ in $\mathcal{D}$
    Sample random minibatch of transitions $(\phi_j, a_j, r_j, \phi_{j+1})$ from $\mathcal{D}$
    Set $y_j = \begin{cases} r_j & \text{for terminal } \phi_{j+1} \\ r_j + \gamma \max_{a'} Q(\phi_{j+1}, a'; \theta) & \text{for non-terminal } \phi_{j+1} \end{cases}$
    Perform a gradient descent step on $(y_j - Q(\phi_j, a_j; \theta))^2$
  **end for**
**end for**

CHAPTER 3

# Methodology

*"Everything must be made as simple as possible. But not simpler."*

– Albert Einstein, *1934*

In this chapter we present in detail the method that is used in our research, as well as its application. The goal is for the reader to understand the intuition behind our approach and to explain how we build up on previous research in the area which was covered in the previous chapter. In a detailed manner we aim to describe the network used in our method and to discuss how it is trained. In addition, an overview of the data construction process for the project is as well given.

We start by describing how the input data was created. A short overview of the platform and the games used for creating the data is provided, including discussion about the framework used for data collection. Next, we present our methodology for creating the input dataset and discuss some of the potential problems that can arise pertaining to the data. What follows is an explanation of the procedure used for preprocessing the data in order to be suitable as input to our model. We divide our method into two parts. The first one regards training a neural network to obtain a policy based on the expert trajectories. This is basically a network similar to the Deep Q-Network in Mnih *et al.* (2013), in which we retrieve the Q-function. Then the policy is obtained in a greedy fashion as defined in Equation 2.33. So far we can see this part as a type of classification based imitation learning. The generalization of the policy is achieved in the second part of our method, where another dataset is used in a regression method in order to obtain the reward function. Following this outline, we will first expound on the network used for learning the Q-function and how we trained it. Furthermore, there will be a discussion about the problems that arose during training the network and how they were solved. Then we will detail the procedure used for obtaining the reward function via regression, including a short discussion on the other dataset needed for this task. At the end, we will give a short overview of the environment in which we worked and the tools that were used during this research.

## 3.1 Construction of Dataset

For apprenticeship learning tasks, one of the biggest problems is the availability of data. Such an approach requires a huge amount of expert input in order to be able to retrieve the underlying policy. In order to acquire expert knowledge, for the majority of tasks, the human input is used, which is both time and resource consuming. Additionally, when the task is to learn how to play a game, the process of constructing the data will consist of choosing episodes of game play where the human expert achieves a very good score (cumulative reward). Therefore, hours may be needed only to create a single acceptable episode of a few minutes, which additionally makes the task of collecting data even more time-consuming and challenging.

For this project, we chose to use games from the Atari 2600 platform as one of the most popular home video gaming consoles from the eighties. The games which we apply our method on are one of the most known Atari games. All games are relatively simple and do not require very complex long-term planning.

### 3.1.1 Explanation of the Games

#### 3.1.1.1 Freeway

In Freeway, the player controls a yellow figure that needs to cross a very busy highway as many times for a total of 2 minutes and 15 seconds. This figure is located in the lower end of the highway and needs to make its way to the upper end, relative to the screen. There is a total of ten driving lanes, out of which the lower five are for driving in the direction of left to right, while the others are for driving from right to left. The cars in the lanes closer to the sides of the highway drive slower compared to the cars in the lanes more central i.e. more distant from the sides. If the figure hits a car, it is pushed approximately two lanes backwards, and then stunned for a very short time (1-2 seconds). The cars follow a deterministic pattern, such that same configuration is achieved a few times in a single game episode. The only way for game termination is after the allocated time passes. Although we do not use reward function in our approach, we need the score in order to compare the performance of our agent to those of a human player or other agents. The score is defined as the total times of crossing the highway in one play. The score is an integer, which means there is no credit for partially crossing the highway. A sample screen from this game is given in Figure 3.1a. Note that, although there are two players, we used only the player on the left of the game screen. There is a total of three actions for controlling the figure: going up, going down, and doing nothing.

#### 3.1.1.2 Space Invaders

Space Invaders is an arcade game where the player controls a laser cannon to prevent alien invasion. The cannon is moved horizontally in the bottom of the screen and can fire one laser beam at a time. In order to kill an alien, the player needs to hit it with a laser beam. The aliens are distributed in six rows and six columns on the screen, moving horizontally from left to right (and vice versa), and slowly descending toward the cannon. They are

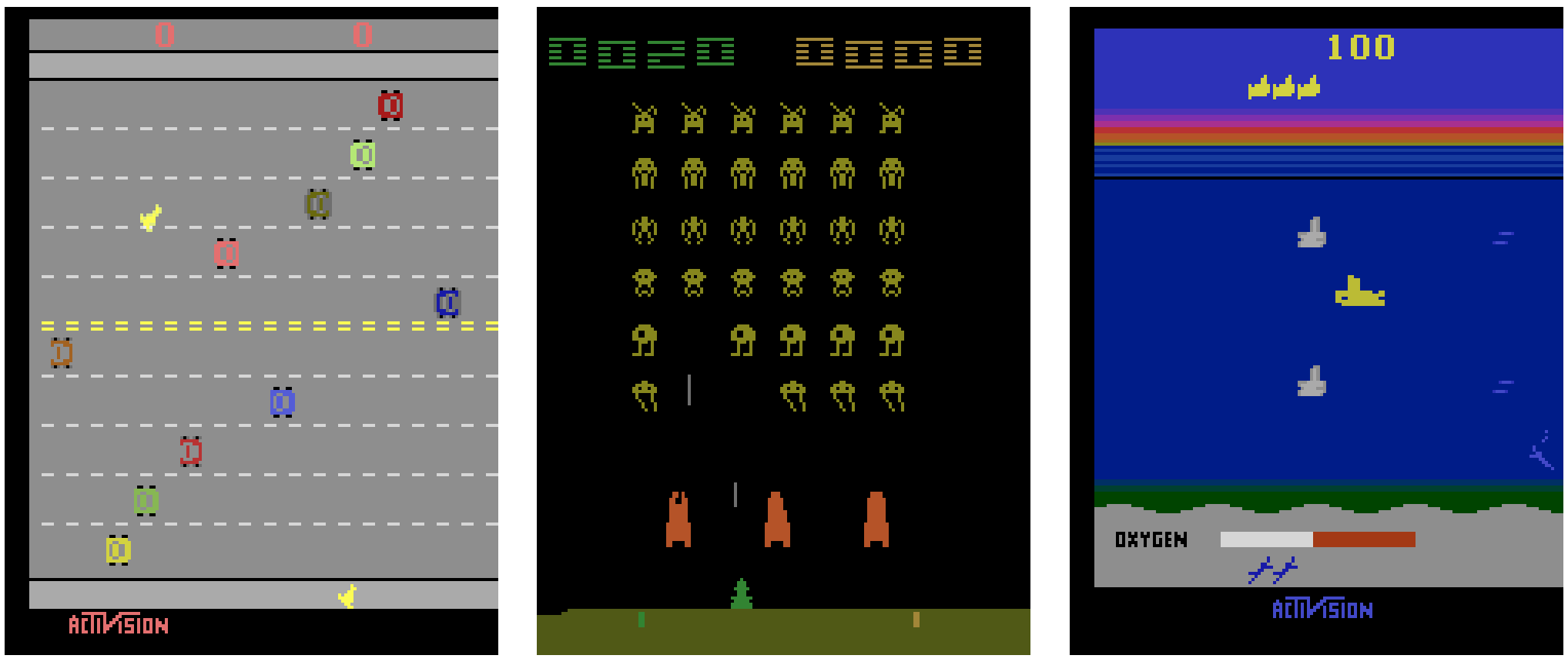

**(a)** Freeway **(b)** Space Invaders **(c)** Seaquest

**Figure 3.1:** Sample Screen from the Atari games used

also equipped with laser beams and shoot toward the cannon in order to destroy it. Besides loosing all three lives, game termination can occur if an alien reaches the same horizontal position as the cannon. The goal is to destroy all the aliens, after which another level starts where they are even closer to the cannon. Two conditions influence the faster movement of the enemies. One of them is the distance between the aliens which are closest to the cannon, and other is the total number of aliens not destroyed. For each destroyed alien, the player gets points, so that aliens farther from the canon give a higher score compared to destroying the closer enemies. Also, once in a while, an alien spaceship appears at the top of the screen, which if destroyed, yields a high number of points. The total score represents the total number of points achieved during one game play. There are six possible actions: doing nothing, firing, moving to the left, moving to the right, going left and firing at the same time, and going right and firing at the same time. A sample screen of this game is given in Figure 3.1b.

#### 3.1.1.3 Sea Quest

Seaquest is a game where the player controls a submarine which needs to collect divers while destroying or avoiding sharks and other submarines. The game screen consists of a cross-section of water surface divided into four imaginary horizontal tracks in which divers, sharks, and submarines, move. Initially, the submarine is at the top of the screen (the surface) and needs to move both horizontally and vertically in order to collect divers. The sharks can be killed (and the hostile submarines destroyed) by shooting torpedoes toward them. The goal is to collect six divers and take them to the surface. However, as time passes the oxygen level in the submarine decreases, which means that there is a limit as to how long the submarine can be underwater. If that time runs low the submarine needs to go to the surface, after which the timer is restarted. Lives can be lost in case a shark or an enemy submarine reaches the player's submarine, a hostile torpedo hits the submarine, or the oxygen timer expires. After the submarine goes to the surface to refill its oxygen level, the enemies advance, meaning that they are faster and harder to avoid. When the submarine collects all the divers needed and brings them to the surface, the player advances to a new level. Points are awarded after passing each level. Moreover, points can be obtained by killing enemies, although much less compared to the first case. As a player advances to a higher level, the points awarded for collecting people increases drastically. The points for destroying submarines and killing sharks, however, remains constant. Possible actions for this game are: doing nothing, moving up, moving down, moving to the left, moving to the right, and firing combined with all the other possible moves. A sample image is given in Figure 3.1c.

### 3.1.2 Arcade Learning Environment

In order to create the data needed for this project, we used the Arcade Learning Environment (ALE) (Bellemare *et al.* , 2012). ALE is an object-oriented framework for developing AI agents for Atari games. It is built on top of the Stella[1] emulator, which is an open source, multiplatform emulator with support for almost all Atari games and peripheral devices. One of the biggest advantages of this framework is that the agents and the games are encapsulated from the emulator, which means that low level emulation details are hidden from the end user. In order to start emulation, only the appropriate game and agent need to be specified. Although, the framework is written in C++, agents can be implemented in any programming language and then they can communicate with the emulator through FIFO pipes. ALE offers default support for approximately 50 Atari games, but there is an interface so other games can be added only by defining the reward function and the termination condition for the appropriate game. ALE communicates with Atari through Stella by sending game commands and receiving screens and RAM states. Therefore, it is straightforward to obtain screen images and actions, needed for our purposes.

[1] `http://stella.sourceforge.net/`

### 3.1.3 Data Collection

The data used for this research was gathered in the period between May 2014 and August 2014. It consists of 73 episodes of Freeway, 97 episodes of Space Invaders and 37 episodes of Seaquest. For Freeway, each of the episodes has a score of more than 30, with some of the episodes attaining a score of 32. We think that such a score can be considered to be a near optimal for this game. One episode consists of 8,700 screen images, equivalent to 2 minutes and 15 seconds of game play, for a total of 500,000 frames. For Space Invaders, the score varies, but the mean is about 3,000 points. The length of the episodes is not fixed as in Freeway. All episodes summed up have 800,000 frames. Seaquest as well does not have constant episode length. We collected a total of 750,000 Seaquest frames, where the mean of the total score of each episode is around 20,000 points. For all the games, each image is $210 \times 160$ pixels and has a 128-color palette. Additionally, a set of actions is known for each episode. As expected, we do not save any information pertaining to the reward function.

In order to obtain reward function in the regression step, we use another data set. It has the same structure as the input data set but it contains as many as possible game states in order to achieve generalization for the states which are not explored in the expert trajectories. For this data set the policy followed is not relevant. Therefore, we used random policy to obtain as many as possible game states. Note that for simple games, such as Freeway, which have only a few valid actions and relatively small state space, the generalization will not be as important as in more complex games.

### 3.1.4 Data Pre-processing

Generally, color images are more complex because objects with identical form but different colors will need to be learned as two distinct features. For example, in Freeway, we are interested in learning the object car, despite its color. It does not make any difference if the figure will hit the blue or the red car. Moreover, for most Atari games, there is a flickering effect such that in consecutive frames the color of an object varies in order to achieve an animation effect. Although the color image encodes more information about the game state, in the case of the Atari games that we used, such information is not important and complicates the learning process. Therefore, the first step in our pre-processing procedure is to convert all the screens from color to gray-scale images, as given in Figure 3.2b. Our network is designed to work with input images of the same dimensions (height and width) because of the requirements of the convolutional networks implementation used. Therefore, the next step is to reduce the input dimensions from $210 \times 160$ to $83 \times 83$ pixels (Figure 3.2c. We chose the number 83 because it is suitable for all the future convolution and pooling operations. The background of a screen does not contribute to any valuable information regarding the learning process, on contrary, it contributes to more complex data. Therefore, we simplify the input even more by stripping the background, which is given in Figure 3.2d. In order to do so, for each pixel, we take an average of the values for that pixel in all the images from one episode. For example, in Freeway, since the cars and the figure are the only important object that we need, by doing such an operation we will strip all the unnecessary objects for further simplification. The resulting image after such an operation is given in Figure 3.2e. Next, in order to further simplify the image we do a kind of contrast normalization, where we divide each pixel (its RGB values) by 255. After this operation, the resulting image will be different shades of the color black. Figure 3.2f represent the image after such operation, where the colours are encoded for better viewing. The complete pre-processing procedure can be seen in Figure 3.2g.

### 3.1.5 Problems

Next we will discuss a few problems that arose during the data collection process. One of the most important decisions in regards to this step is the frame rate of data generation while the human (expert) player plays. This is the number of frames per second created during playing. If this value is very high, then we will end up with a huge amount of data for a relatively short effective playing time. On the contrary, a very low rate could potentially cause problems with the learning process because some important behaviour will not be captured by the expert data. This could range from skipping the transition process between states to serious issues, such as missing whole state-action pairs. This parameter needs to be set in the ALE. Pertaining to ALE, another important issue is that when an episode ends (game termination occurs), ALE continues to save screen images and actions until the process is not directly terminated. For example in Freeway, from our experience, most of the episodes end while the figure is in the middle of the highway. Therefore, if the process is not terminated instantly we could end up with having thousands of screens from the last state before termination accompanied with the same number of 'nop'[2] actions. This could influence the training process and overfit the model on those states that the game usually ends with. Basically, it could means that the agent will learn that in specific states, it is better to do nothing(instead of going up or down), which is not true, but it is a result of the problem described above. Hence, a safe solution is to limit all the episodes to a fixed number of frames, and to cut off the rest of the images. However, Space Invaders and

[2] The action 'nop' directs the agent not to take any moving action i.e. to stay in the same position.

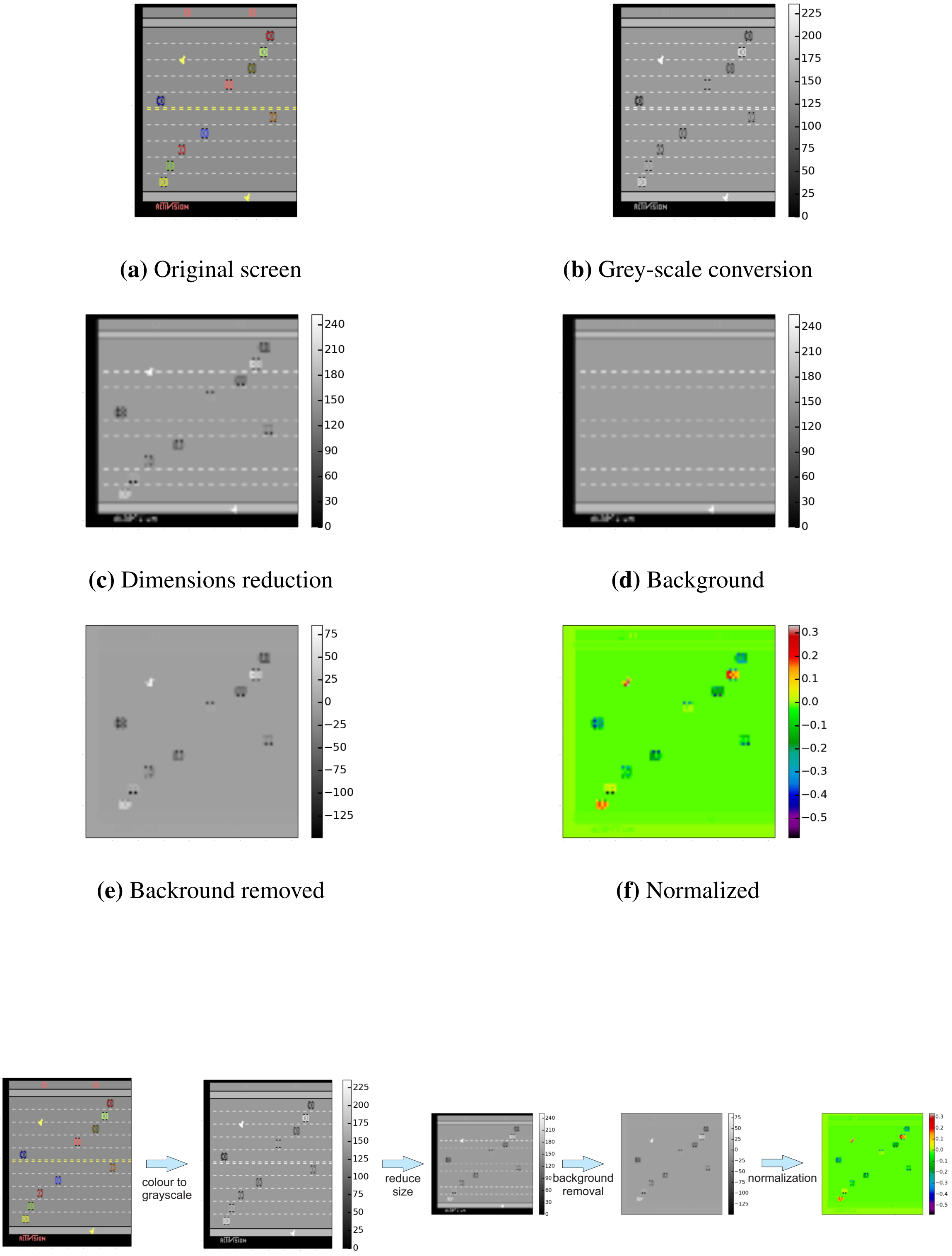

**(a)** Original screen

**(b)** Grey-scale conversion

**(c)** Dimensions reduction

**(d)** Background

**(e)** Backround removed

**(f)** Normalized

**(g)** The preprocessing procedure of a frame from the game Freeway

**Figure 3.2:** Preprocessing of input data

Seaquest do not have episodes of a fixed length. Because of that, the cut-off for these games is determined by the final score. Concerning the second dataset, it is important that this dataset is created in the same settings as the input data set. Even if the frame rates differ we can still obtain good results in some cases, but it is still better to use the same frame rates in order to avoid possible issues when extracting the reward function. Generally, most of the problems at this point were somehow connected to the arcade environment (ALE). Since we used two versions of ALE, some of the options from the latest version were not available in the older version. For example, we also collected datasets for the games Enduro and Breakout, but due to emulation problems they are not included. Additionally, these two versions of ALE use a different version of the emulator Stella, resulting in different outputs for the same Atari ROM. Such a problem prevents us from learning a successful model for Space Invaders. Therefore, anyone who plans to recreate our results or work in the reinforcement or apprenticeship learning methods to be identically applied as in our project must know that many problems connected to the emulation process will arise. Although most of them, if not all, can be solved, their resolution requires plenty of time.

## 3.2 Deep Apprenticeship Q-Network (DAQN)

The first step of our method is to learn a score function that rates actions for states ($q$-function), given the expert trajectories. In order to do so we use a convolutional neural network, similar to the Deep Q-Network presented in (Mnih *et al.* , 2013). We will call this network *Deep Apprenticeship Q-Network* (DAQN). As input to the network we provide one to four frames from the expert dataset, which are pre-processed as described in Section 3.1.4. The training is done using the AdaGrad algorithm (Duchi *et al.* , 2011), which is a variation of the stochastic gradient descent method. The weights are not updated for each instance, instead, for every iteration, 32 samples are passed forward on the network as a batch.

### 3.2.1 DAQN Structure

The *Deep Apprenticeship Q-Network* is a neural network which consists of three hidden layers, out of which two are convolutional and one is fully connected. The first convolutional layer has 16 feature maps, which means it can learn 16 distinct features. Since each feature map has tied weights, there is a total of 16 filters. Each unit in a feature map takes $8 \times 8$ patches from the input image, therefore the filters are with those dimensions. The stride is 4, which means that two neighbouring patches are 4 pixels apart. At the end, a max-pooling of the size $4 \times 4$ is applied. Therefore, each of the feature maps will have the dimensions of $19 \times 19$ units. The second convolutional layer consists of 32 feature maps, each with dimensions of $8 \times 8$. There are also 32 filters with dimensions of $4 \times 4$. Max-pooling with stride 4 is applied at the top. Each of the units (neurons) from the feature maps in the second convolutional layer is connected to the third hidden layer, which has 256 neurons. As will be discussed in Chapter 4, best results are obtained by using hyperbolic tangent (tanh) as activation function on the neurons of the hidden layer. The output layer consists of one neuron for each of the possible actions for a game. Each of the output neurons behaves as a logistic regression model and apply the soft-max function on the input given by

$$S[q(s,a)] = \frac{\exp q(s,a)}{\sum_{a'} \exp q(s,a')}. \tag{3.1}$$

By applying such non-linearity the score function will be normalized to values in the interval $[0,1]$. Moreover, the sum of the values of the score function for a given state and all of the possible actions is equal to one. Therefore, such approach is suitable for representing probability distribution. We need such approach because we want to extract policy afterwards, which can be seen as probability distribution of actions over states. The output represents the values of the score function $q$ for the state represented by the input sequence of game frames and each of the possible actions. This network can be seen in Figure 3.3.

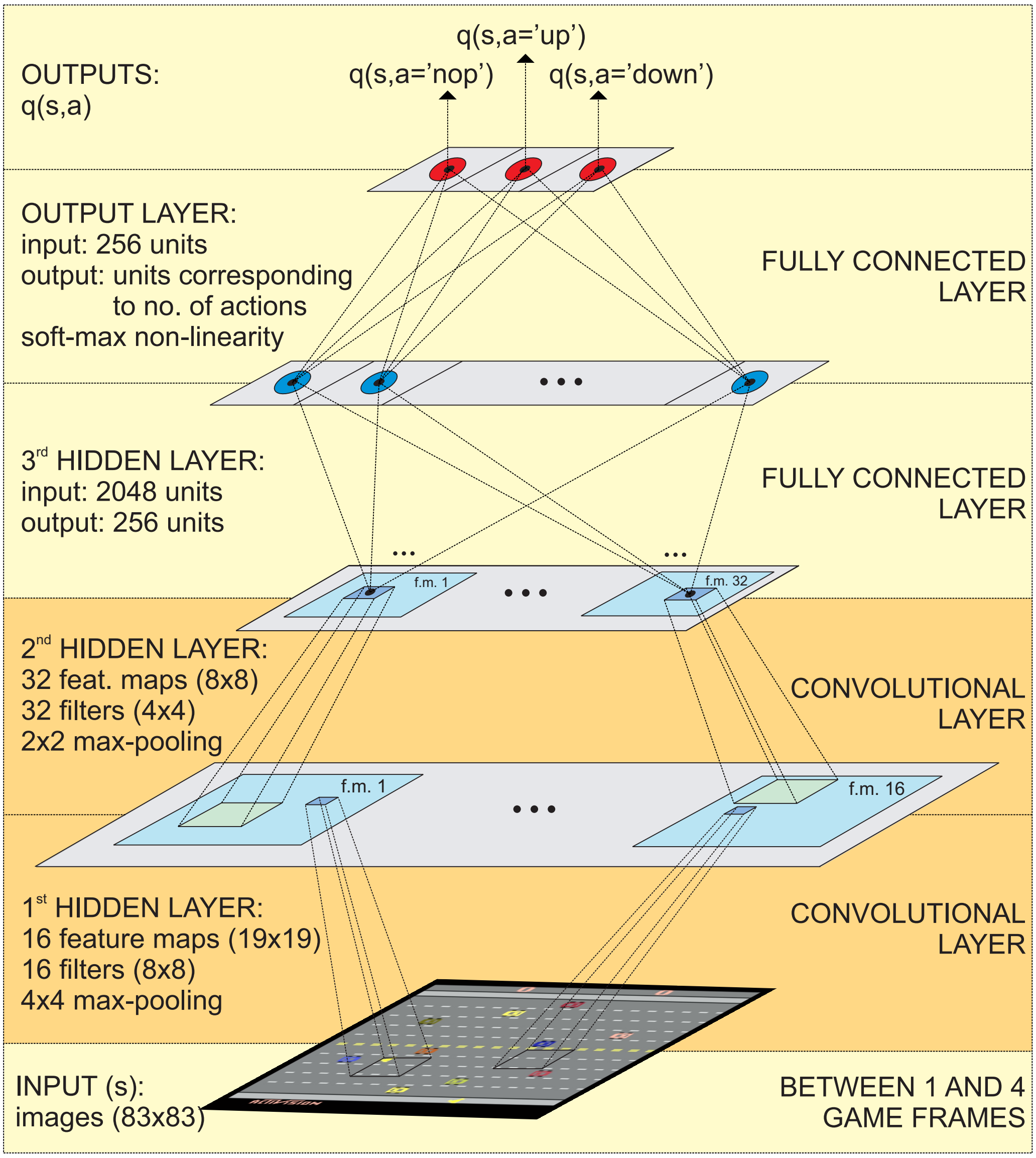

**Figure 3.3:** Deep Apprenticeship Q-Network Structure

### 3.2.2 DAQN Training

The score function $q$ which DAQN learns is actually the action-value function $Q$ for the expert policy. Therefore, by learning this function we can obtain a policy optimal for the $q$-function based on Equation 2.33 as

$$\pi^*(s) = \max_a q(s, a). \tag{3.2}$$

The learned policy will approximate the expert policy i.e. $\pi^* \approx \pi_E$.

In order to train a neural network we need to optimize the loss function $J$ as given in Equation 2.8. For such a loss function with parameters $\mathbf{w}$, the update of one parameter (weight) $w_i$ when using gradient descent for optimization is given by

$$w_i^{t+1} = w_i^t - \eta \nabla_{w_i} J(w), \tag{3.3}$$

where $w_i^t$ is the value of the weight $w_i$ in time step $t$ s.t. $t = 1, ..., T-1$ and $i = 1, ..., I$, $I$ is the number of parameters (total sum of all the layers), $\nabla_{w_i} J(\mathbf{w})$ is the gradient of the loss function $J(\mathbf{w})$ in respect to $w_i$, and $\eta$ is the learning rate. The loss function $J(\mathbf{w})$ is given by

$$J(\mathbf{w}) = \sum_a \left[q_{\mathbf{w}}(s, a) - \hat{q}(s, a)\right]^2, \tag{3.4}$$

where $q_{\mathbf{w}}(s, a)$ is the output of the DAQN network and $\hat{q}(s, a)$ is the expected output. For input data $(s = s_n, a = a_n) \in D_E$, the expected output $\hat{q}$ is given by

$$\hat{q}(s, a) = \begin{cases} 1 & \text{if } a = a_n \\ 0 & \text{if } a \neq a_n \end{cases}. \tag{3.5}$$

The learning rate $\eta$ in Equation 3.3 is a hyper-parameter which dictates how fast the weights will converge toward their optimal values (values that minimize the loss function $J(\mathbf{w})$ given in Equation 3.4). Regular stochastic gradient descent methods as described above have only one learning rate parameter for all weights. Such an approach, however, does not produce fruitful results in our case. Therefore, we use another algorithm which is based on the stochastic gradient descent, but where the learning rate for each parameter is defined separately. This method is called AdaGrad and will be discussed shortly. Pre-training of the model helps in many deep learning approaches. In our case we tried using the RICA method to pre-train the filters of the first convolutional layer, but such an approach does not result in better overall results. However, we will discuss this pre-training method as well.

#### 3.2.2.1 AdaGrad

Having one learning rate for all the features can pose a problem when some of the features are less frequent (or with different scaling) compared to others. Therefore, Duchi *et al.* (2011) present an algorithm called *AdaGrad*, which has a different learning rate for each of the parameters (weights). The learning rate for each parameter is iteratively calculated in order to reflect the data seen so far by the model. This represents a more informative approach compared to the stochastic gradient descent as described above. The learning rate for parameter $w_i$ at time step $t$ is given by

$$\eta_i^t = \frac{\eta}{\sqrt{G_{i,i}^t}} \tag{3.6}$$

where $\eta$ is some initial learning rate, $G$ is a diagonal matrix defined as

$$G_{ij}^t = \begin{cases} \sum_{t'=0}^{t} g_{t',i}^2 & \text{if } i = j \\ 0 & \text{if } i \neq j \end{cases} \tag{3.7}$$

such that $g_{t,i}$ represents the gradient of parameter $w_i$ at time step (iteration) $t$. Therefore, the learning rate for a given feature and time is calculated by dividing the initial learning rate by the square root of the sum of the squares of all previous gradients for that particular feature. This will result in a lower value for the learning rate of the more frequent features and proportionally, higher values for the learning rate of rarely observed features. The consequence will be that even features which are not regularly observed can be recognized because albeit scarce, they may be important for the learning process.

#### 3.2.2.2 RICA Filters

One of the breakthrough results that increased the interest in deep learning was the work of Hinton *et al.* (2006) and Bengio *et al.* (2007). Both approaches are about a pre-training phase in unsupervised fashion for deep neural networks. The problem is that for deep networks regular supervised training may not work well. Erhan *et al.* (2009) show that for different initial parameter values, gradient based methods converge to different local optima. Bengio *et al.* (2007) claim that unsupervised pre-training acts as a regularizer and therefore the result is better generalization. Briefly, a pre-training unsupervised phase for a neural network with one hidden layer consists of constraining the weight matrix of the output layer to be transposed of the weight vector in the hidden layer and have the same number of neurons in the output layer as in the input. By training such a network, the hidden layer will learn a feature representation of the input. To learn more abstract representations, this procedure can be applied multiple times by adding the same structure between a hidden layer that is trained and the output layer as between the hidden layer and the output, such that the weight matrices in the upper layers are transposed to the corresponding weight matrices in the lower layers. One way of doing unsupervised pre-training is by using *Independent Component Analysis* (ICA), which forces sparse representation. The optimization in such a case is given as

$$\min_{W} ||Wx||_1 \qquad s.t. \qquad WW^T = I \tag{3.8}$$

where $W$ is the weight matrix of the hidden layer,$x$ is the input, $I$ is the unit matrix, and $\|\cdot\|_1$ is $l_1$ norm. The condition $WW^T = I$ forces the features to be orthogonal in order to obtain sparser representation. However, the problems appear when the features are overcomplete, which means that "the number of features can exceed the dimensionality of the input data" (Le *et al.* , 2011). They propose a modification of this method, called *Reconstructed ICA* (RICA), which we tried out in our work for pre-training the first convolutional layer in the DAQN network. This method proposes using smooth penalty for the orthogonality instead of using hard constraint, as in ICA. The optimization task is then given as

$$\min_W \left( \lambda \|Wx\|_1 + \frac{1}{2}\|W^T W x - x\|_2^2 \right) \tag{3.9}$$

where $\|\cdot\|_2$ is $l_2$ norm and $\lambda$ is a constant which tells how important the orthogonality constraint is. When $\lambda \to \infty$ the RICA method is identical to ICA. However, such a method does not achieve better results in our network. The reason probably lies in the fact that convolutional networks can recognize features even if their scaling differs, so such a step is not required. We will not go into more detail on RICA or unsupervised pre-training since it is out of the scope of this dissertation. In any case though, a great source for unsupervised pre-training and its benefits can be found in Bengio *et al.* (2007) and Erhan *et al.* (2010).

## 3.3 Extraction of Reward Function

After we learn the Q-function using the DAQN network as described above, we can obtain a policy by using a greedy approach as defined in Equation 3.2. However, as discussed in Section 2.3.1, such policy may not generalize well on unexplored states in the expert trajectories. Technically, if we obtain a perfect feature extraction for the problem, the network will produce the desired output even for unseen states. Nevertheless, such an assumption is very optimistic, especially when the input is a relatively complex sequence of video frames. Accordingly, we first try to obtain a reward function such that the policy which is optimal for this function will generalize well. For that task, we use an approach similar to the one described in Section 2.3.2.1.

We start by creating another dataset, as described in Section 3.1.3. The purpose of this dataset is to contain as many as possible game states. Since how the policy is created is not of any importance, we use an agent that follows random policy to create such a dataset. As expected, the instances of that dataset are pre-processed in the same manner as the expert trajectories, described in Section 3.1.4. Afterwards, this dataset is presented as input to another neural network, which learns a reward function $r$ based on the score function $q$.

### 3.3.1 DARN Structure

We refer to the network used for learning the $r$ function as *Deep Apprenticeship Reward Network* (DARN). DARN has the same structure as the DAQN network already discussed. The convolutional layers are initialized with the same values as the optimal parameters of the DAQN network. Such initialization acts as a pre-training phase, since this network is fed with input frames from the same game and settings as in the DAQN. Therefore, the process of feature extraction is the same. The only difference is that this network does not learn the score function $q$, but uses this function in order to learn a reward function $r$.

### 3.3.2 DARN Training

The function that DARN learns, denoted as $r$ function, tries to recover the reward function $R$ for which the policy $\pi_E$ is optimal. Therefore, they can be seen as the same function i.e. we try to learn the $r$ function for which the expert policy $\pi_E$ is optimal. The calculation of the reward function is implemented using the trick for representing the reward in terms of the Q-function, as given in Equation 2.51, which was proposed by Klein *et al.* (2013). For our network, given input instance $(s, a, s') \in D_g$, we can rewrite Equation 2.51 as

$$r(s, a) = q(s, a) - \gamma q(s', \pi(s')) \tag{3.10}$$

Since we are using a greedy policy, we can rewrite Equation 3.10 as

$$r(s, a) = q(s, a) - \gamma \max_{a'} q(s', a') \tag{3.11}$$

The score function $q(s, a)$ in Equation 3.11 represents the value of the output neuron that corresponds to action $a$ from the DAQN network for input $s$. More specifically, we take the values before applying soft-max smoothing, meaning that we take the pre-synaptic value of the corresponding output neuron (instead of the post-synaptic value). We will denote such values as $q^{PRESOFT}(s, a)$ Therefore, we can rewrite Equation 3.11 as

$$r(s, a) = q^{PRESOFT}(s, a) - \gamma \max_{a'} q^{PRESOFT}(s', a') \tag{3.12}$$

The loss function of the DARN network that needs to be optimized can be written as

$$J_r(\mathbf{w}) = \|r_{\mathbf{w}}(s, a) - \hat{r}(s, a)\|_2 \tag{3.13}$$

where $\| \cdot \|_2$ is $L_2$ norm, $r_{\mathbf{w}}(s, a)$ is the output of the DARN network, and $\hat{r}(s, a)$ is the expected reward, computed as in Equation 3.12. We can rewrite the complete expression for the loss function that needs to be optimized as

$$J_r(\mathbf{w}) = \|r_{\mathbf{w}}(s, a) - \Big(q^{PRESOFT}(s, a) - \gamma \max_{a'} q^{PRESOFT}(s', a')\Big) \|_2 \tag{3.14}$$

The whole structure needed to extract the reward is given in 3.4. Note that the weights of the DAQN networks are fixed while we train the DARN network.

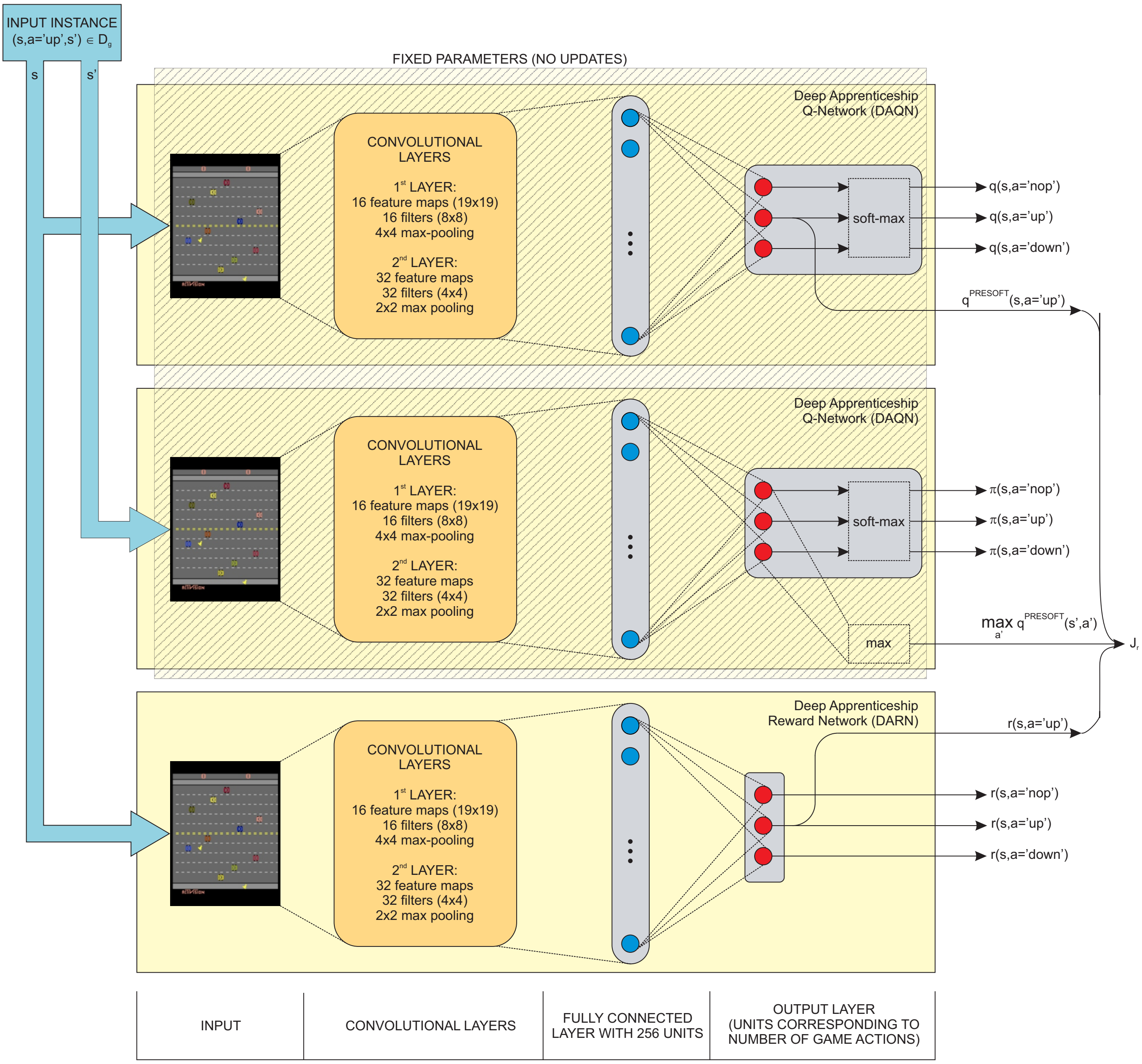

**Figure 3.4:** Reward Extraction in Deep Apprenticeship Learning

## 3.4 Generalized Policy

While the reward that is learnt will capture the expert knowledge, it will not behave well in states not visited by the expert since the policy we follow is undefined for such states. Considering that in most situations we will not be able to supply expert trajectories for all the game states, in addition to the fact that even if such amounts of data are provided, some states may not be visited at all because they are not beneficial for the task, we need to extend our method to unseen states. Since we already have a reward function, the problem reduces to standard reinforcement learning task where we need to learn a policy given a reward by incorporating the dynamics of the environment so that our policy is well defined on the whole state space. In order to do so, we use the approach of (Mnih *et al.* , 2013) because it is state-of-the-art at the moment for reinforcement learning tasks for complex input such as video.

## 3.5 Deep Apprenticeship Learning (DAL)

Since to our knowledge there is no other work on using apprenticeship learning for complex and multidimensional input, we will call this approach *Deep Apprenticeship Learning* (DAL). The algorithm for our approach is given in Algorithm 5. A short summary of the method follows.

We start by generating expert trajectories, which are merged into one data set, denoted as $D_E$. The next step is generating another dataset $D_G$, which consists of trajectories by following random policy. The idea of this dataset is for it to be used for discovering the dynamics of the environment and to help in the generalization for states not included in the expert data. We first train the DAQN network in order to learn the score function $q$ which approximates the action value function $Q$, where the AdaGrad algorithm is used for optimization instead of the regular stochastic gradient descent. This function is solely based on the expert input, which means that for beneficial states not explored, the agent will not know about the high rewards that could be obtained by visiting such states. Therefore, we try to retrieve the reward function which will generalize on unknown states. We use another network, DARN, which basically has the same structure as the DAQN network. We use $D_G$ as input to this network in order to learn the reward function $r$. For such a task, we are exploiting the Bellman equation in order to compute the reward, given the values of the action-value function $Q$ for two consecutive states, which are known from the DAQN network. In the end, we obtain the final policy in a greedy manner by taking the actions which produce the highest reward. Such policy will not generalize well on unvisited states. Therefore, we use the approach of (Mnih *et al.* , 2013) to learn new, improved policy, that will generalize on unseen states.

---

**Algorithm 5** Deep Apprenticeship Learning

***Given***

Dataset of expert trajectories $D_E$ defined as $D_E = \{(s_t, a_t = \pi_E(s_t))\}_{t=1}^{T}$,
Dataset of random play for generalization $D_G$, defined as $D_G = \{(s_i, a_i, s_i')\}_{i=1}^{N}$,
Hyper-parameters $\gamma, \eta$,
Randomly initialized parameters $\mathbf{w}$ and $\mathbf{v}$,
Matrix $G$ with dimensionality $T \times |\mathbf{w}|$ s.t. $G[a, b] = g_{a,b} = 0, \forall a, b$,
Matrix $H$ with dimensionality $T \times |\mathbf{v}|$ s.t. $H[a, b] = h_{a,b} = 0, \forall a, b$.

***Train*** DAQN network on $\forall(s_E, a_E) \in D_E$ that learns score function $q(s, a)$.

- the loss function is given by $J(\mathbf{w}) = \sum_a \left[q_{\mathbf{w}}(s_E, a) - \hat{q}(s_E, a)\right]^2$
- the estimate $\hat{q}(s, a)$ is given by by $\hat{q}(s, a) = \begin{cases} 1 & \text{if } a = a_E \\ 0 & \text{if } a \neq a_E \end{cases}$
- the weights update is given by $w_{t+1,i} = w_{t,i} - \frac{\eta}{\sqrt{\sum_{t'=0}^{t'=t} g_{t,i}^2}} \nabla_{w_i} J(w)$

***Train*** DARN network on $\forall(s_g, a_g, s_g') \in D_G$ that learns reward function $r(s, a)$.

- the loss function is given by $J_r(\mathbf{v}) = \|r_{\mathbf{v}}(s_g, a_g) - \hat{r}(s_g, a_g)\|_2$
- The reward estimate $\hat{r}(s_g, a_g)$ is computed as:
  $\hat{r}(s_g, a_g) = q^{PRESOFT}(s_g, a_g) - \gamma \max_{a'} q^{RPESOFT}(s_g', a')$
- the weights update is given by $v_{t+1,i} = v_{t,i} - \frac{\eta}{\sqrt{\sum_{t'=0}^{t'=t} h_{t,i}^2}} \nabla_{v_i} J(v)$

***Generalize*** learn new policy $\pi$ using ***Algorithm 4***

***Output*** policy $\pi$

---

## 3.6 Environment and Tools

In terms of implementation, for this project we mainly used Python environment. For working with ALE there is a Python interface, but if new games are added a small part of the programming needs to be done in C++. For generating game plays, we used Java interface built on top of the ALE's C++ implementation. We extensively used Theano and the Numpy package for Python. For understanding the source code given in Appendix A, the reader needs to have a basic knowledge of Python and an understanding of Theano. We will assume an understanding of basic Python so there will be no further discussion about this programming language. As Theano is a newer library, which is especially used in deep learning research, we will give a short introduction.

### 3.6.1 Theano

*Theano* is library for Python introduced by Bergstra *et al.* (2010), and used for large-scale computations. It allows optimizing and evaluating expressions of scalars, vectors, matrices, and tensors on CPUs and CPUs. Theano is built on top of the Python library Numpy[3]. In order to write a function in Theano, the user should first define the expression for that function, compile it so a symbolic graph is created and then use it. Theano uses a functional programming paradigm, compared to the imperative programming in Python. Because of this, Theano can compute a gradient much easier compared to pure Python. By defining the symbolic expression for functions, Theano builds symbolic graphs for those functions, and therefore in case a function is changed, the gradient will be directly computed for the new function since the symbolic graph is changed as well. This is a tremendous advantage of Theano, especially when we work with complex, deep networks, since in case of changes in the network, the gradients need not be recomputed manually (which can easily take days, even weeks, for complex networks). Moreover, the computations are much faster compared to pure Python because it uses g++ (for CPUs) and CUDA (for GPUs). Even though it belongs to the functional programming paradigm, Theano allows variables that have explicit values, called shared variables. One of the biggest strengths of Theano is that it can use GPUs for computations. If we take into account the fact that GPUs are much faster than CPUs, we can arrive to the conclusion that Theano is much faster compared to some C implementations for attacking similar problems. Its documentation[4] states that GPU implementations are as much as 140 times faster than CPU implementations.

[3] `http://www.numpy.org/`

[4] https://pypi.python.org/pypi/Theano

CHAPTER 4

# Results and Analysis

*"I pass with relief from the tossing sea of Cause and Theory to the firm ground of Result and Fact."*

– Sir Winston Churchill, *1898*

In this chapter, we will discuss the results of our work. Since this is still work in progress as stated later in Section 5.1, we will present the result from implementing the approach explained in Chapter 3. We start by an analysis of the network, in terms of the hyper-parameters and structure, which produce the best results. Then, we continue with a discussion about the learned reward and interpretation of the cumulative reward graphs for few specific strategies of game play. What follows is an analysis of the influence maps, which are kinds of heat maps and that tell us which pixels are more important for the learning process, based on the change of the gradient for the parameters.

## 4.1 Analysis of the Network

We tried different settings in terms of the structure of our network, such as varying the number of neurons (units) in the fully-connected layers, varying the filters size and the number of feature maps in the convolutional layers. Therefore, here we will present only the

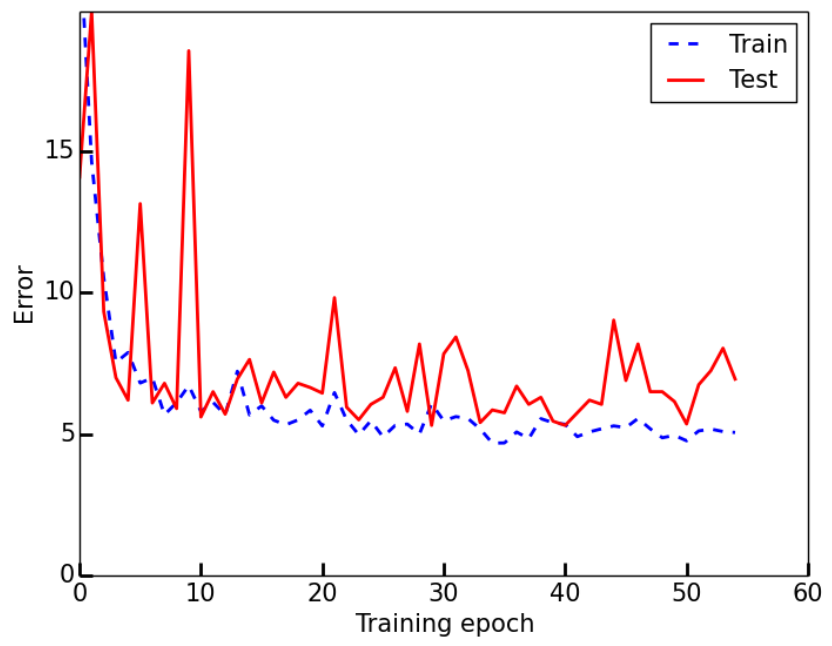

**(a)** Freeway

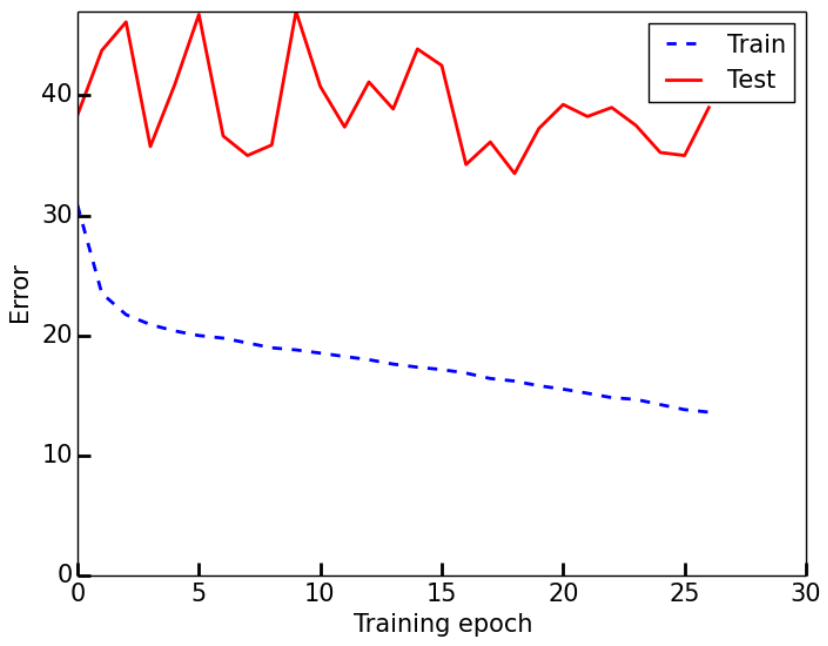

**(b)** Space Invaders

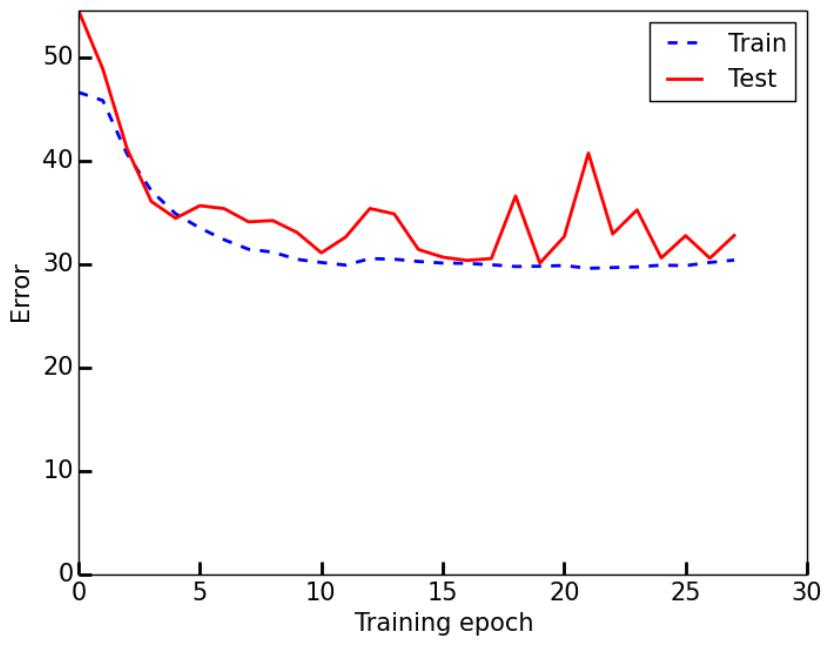

**(c)** Seaquest

**Figure 4.1:** The change of the training and testing error with the number of epochs for each of the games

results when using the network structure for which we obtained the best results, as discussed in Chapter 3. Still, we will discuss the implications of the choice of activation function to the training and testing error of the model. We will also analyse how the length of the input sequences given by the number of game frames in one sequence, can influence the results.

The training and testing error for each of the games is given in Figure 4.1. The results for Freeway are much better compared to the other two games. It is mainly because of the fact that this game has a really simple state space. Because of that, much less data is needed in order to learn a relatively good model. Therefore, we are not expecting much improvement for this game only by increasing the amount of expert trajectories. The training and test error for this game can be seen in Figure 4.1a. In Space Invaders, as can be seen in Figure 4.1b, the training error is still decreasing with time. Therefore, we will need to perform longer training phase for this game in order to obtain satisfactory results. The high values for the test error can be explained by the possibility of choosing game episodes which are not suitable for the testing phase. For Seaquest, as can be seen in Figure 4.1c, it is evident that more data is needed for better performance. The rationale for such statement is that this game is considerably more complex compared to the previous two and requires substantial planning in advance.

**Table 4.1:** Error on the test set for each of the games

| Input | Freeway | Seaquest | Space Invaders |
|---|---|---|---|
| one game frame | 5.3% | 60.2% | 60.2% |
| two game frames | 4.3% | 34.0% | 34.0% |
| four game frames | 5.5% | 29.9% | 29.9% |

The test error for each of the games for different number of game frames as input is summarized in Table 4.1. As we can see, for Freeway the test error decreases with a lower number of game frames in the input sequence. This is because of the fact that for that game, one frame is enough to capture the current state. For Seaquest and Space Invaders this is not the case. The reason is that for these games one frame is not enough to capture the game state at any given moment. For example in Seaquest, one frame does not give information about the direction in which the submarine is headed, or if it is not moving at all at that given time. A similar rationale applies to Space Invaders as well, where one frame is not enough to infer the directions of the objects and the bullets.

Next, we discussed the results in terms of the total score obtained. Note that, only the results for Freeway will be analysed since the other two games do not achieve any score. We present only the best results which are obtained when using tanh non-linearity. As can be noted from Table 4.2 our approach outperforms all other artificial methods. By random play, the score is zero because the agent will not reach the other side of the highway, since the probability for each of the possible actions is equal. Using Sarsa, (Bellemare *et al.* , 2012) reports a score of 11, which is considerably less than our approach, deep apprenticeship learning, which obtains score of 17. Results of using deep reinforcement learning, as in

**Table 4.2:** Comparison of obtained scores for Freeway using different approaches. The Sarsa scores are obtained from Bellemare *et al.* (2012)

| Method | Reported score |
|---|---|
| random agent | 0 |
| human player | 32 |
| Sarsa | 11 |
| DAL | 17 |

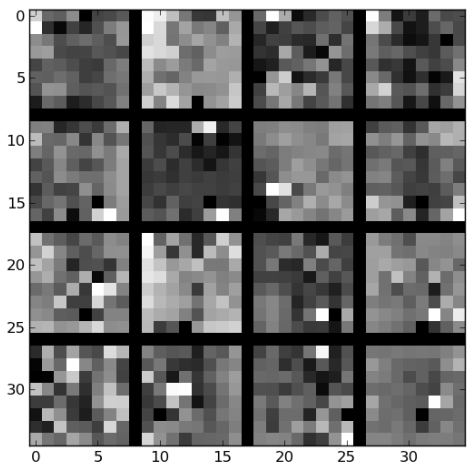

**(a)** AdaGrad

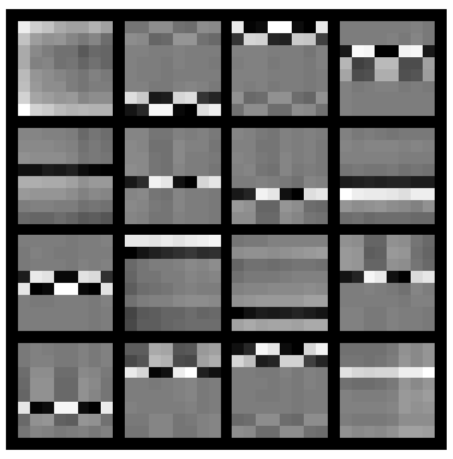

**(b)** RICA

**Figure 4.2:** The filters of the first conovlutional layer for Freeway

Mnih *et al.* (2013), are not reported in their paper. We suspect that it might be the case that $\epsilon$-greedy policy cannot achieve the behaviour wanted for this game. More specifically, since the probability of choosing an action is equal for all possible ones, the agent most likely would not move considerably more upwards than downwards. Because of the definition of score in Freeway, such an approach would not be fruitful. One exception would be if the parameters in their network are randomly pre-initialised so that the agent moves upwards all the time, which is a very unlikely possibility.

As we discussed in Chapter 3, we tried pretraining the network using RICA filters, which was not beneficial for our network. The comparison between the filters pre-trained with RICA and trained using AdaGrad for Freeway is given in Figure 4.2. Since these are the filters of the first convolutional network they recognize the most basic forms in an input image. Indeed, as can be seen in Figure 4.2 the filters recognize basic lines, which represent the edges of the objects, in our case the cars and the figure in Freeway.

## 4.2 Reward Function Analysis

What follows is discussion of the results pertaining to the DARN network as explained in Chapter 3. Since we obtained best result for Freway, in terms of learning a policy by the score function, we performed reward training only for this game. We will discuss the cumulative reward in time for four specific scenarios. In the first scenario, the player is not taking any action i.e. it is staying all the time in the initial position. The cumulative reward obtained can be seen in Figure 4.3a. The second scenario is when the agent is going upwards and downwards all the time, but without crossing the highway. The cumulative reward for such case is given in Figure 4.3b. We can see that the cumulative reward for those two cases is dropping significantly with time. For the other two cases, which represent a scenario when the agent is moving upward all the time and a scenario where the agent is playing optimally, respectively, the cumulative reward is much better, as can be seen in Figure 4.3c and Figure 4.3d. As can be noted, the case where the agent is moving upwards is more beneficial compared to the case where expert play is performed. This can be expected since the model learns that going upwards for such a game is a much better action compared to all the other possible ones. Therefore, a policy which consists of moving strictly in this direction will obtain good score. Moreover, our agent learnt that if cars are occasionally avoided the cumulative score may get even better. Because of this, we can conclude that such an approach for Freeway is indeed successful. We claim that our approach can outperform all the others reinforcement learning methods, since in such methods learning the specific policy beneficial for this task is almost impossible.

## 4.3 Influence Maps Analysis

In order to get information if the agent recognizes features in the input frames, we used influence maps which represent the gradient of the outputs in terms of the input. Such maps,

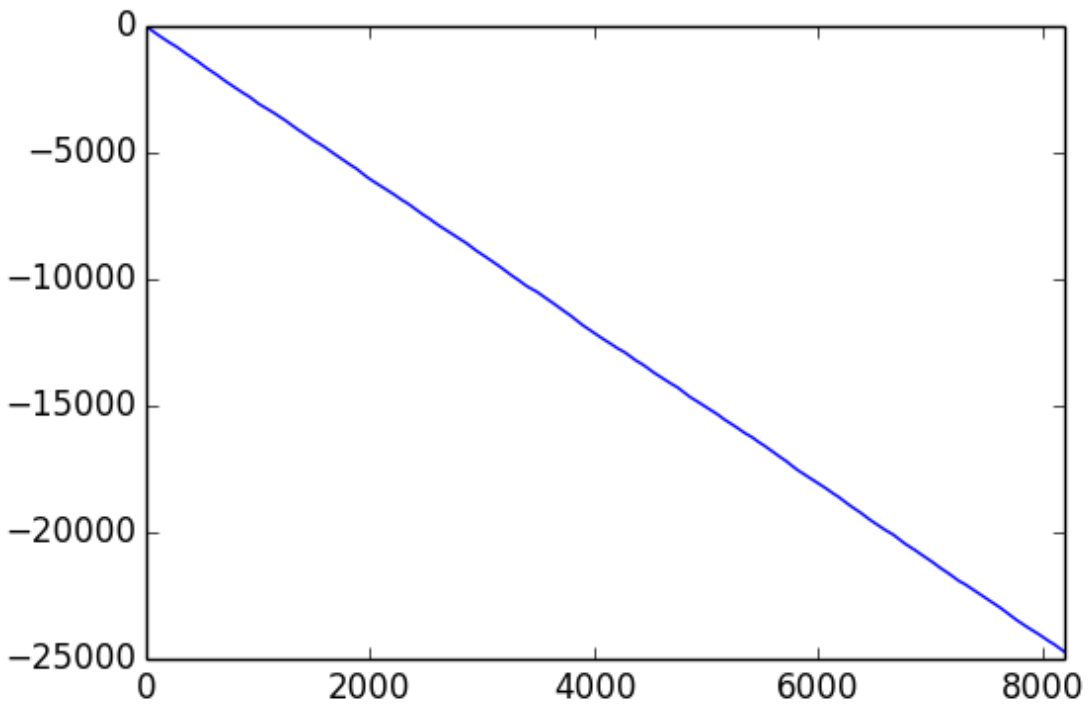

**(a)** No movement

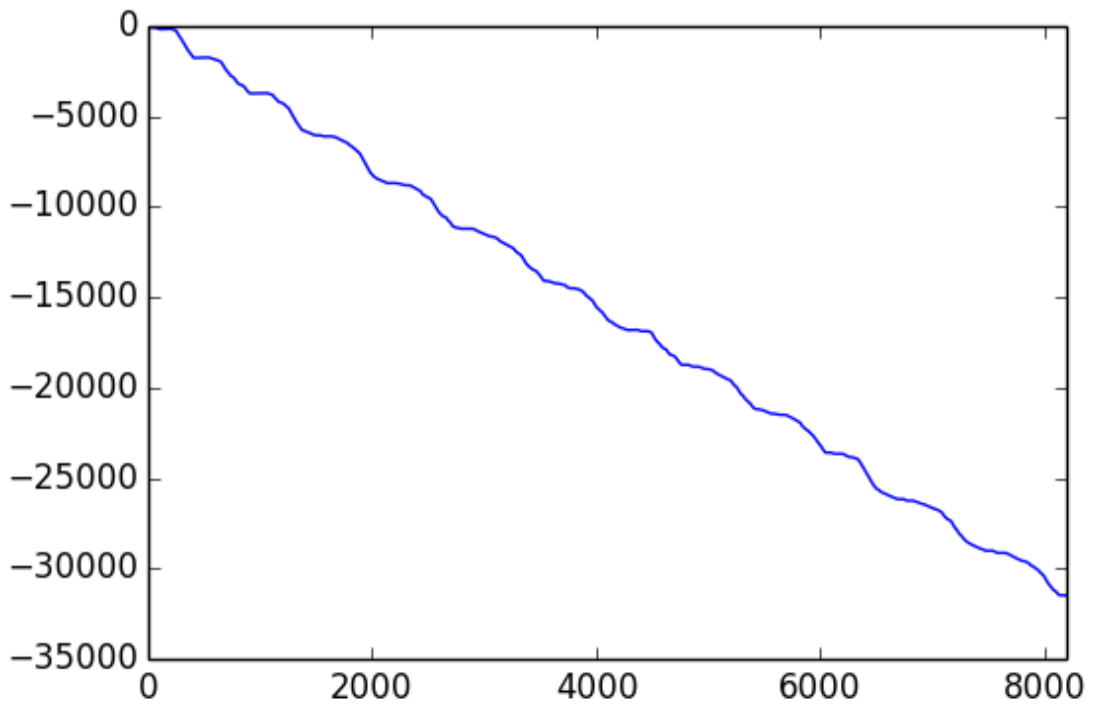

**(b)** Moving up and down without crossing

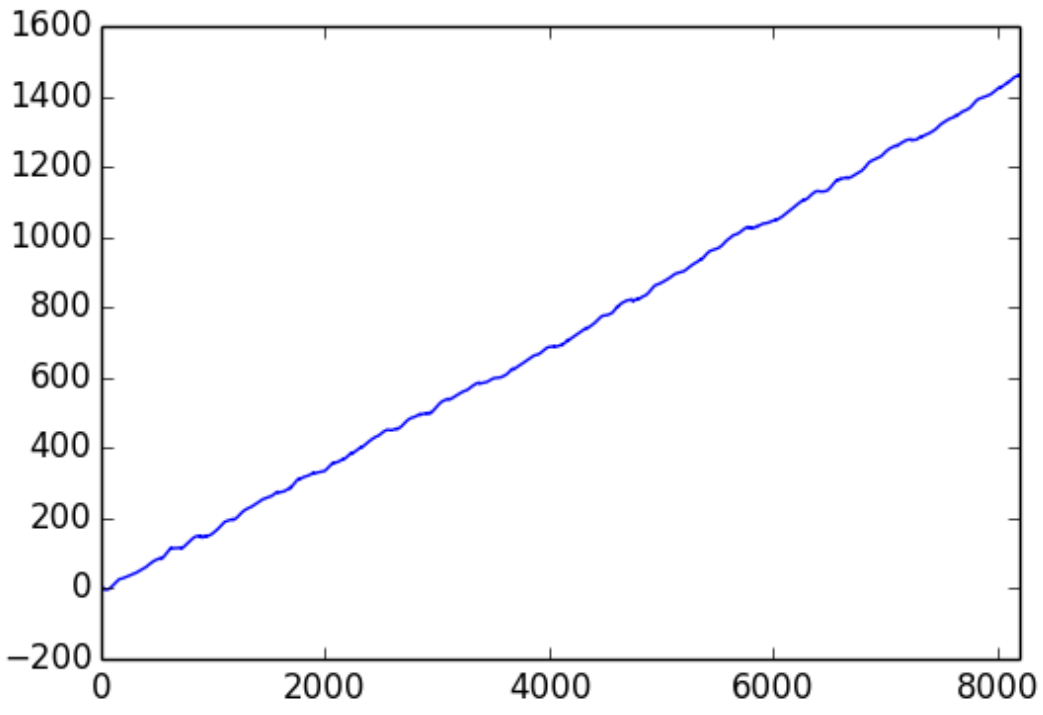

**(c)** Moving upwards all of the time

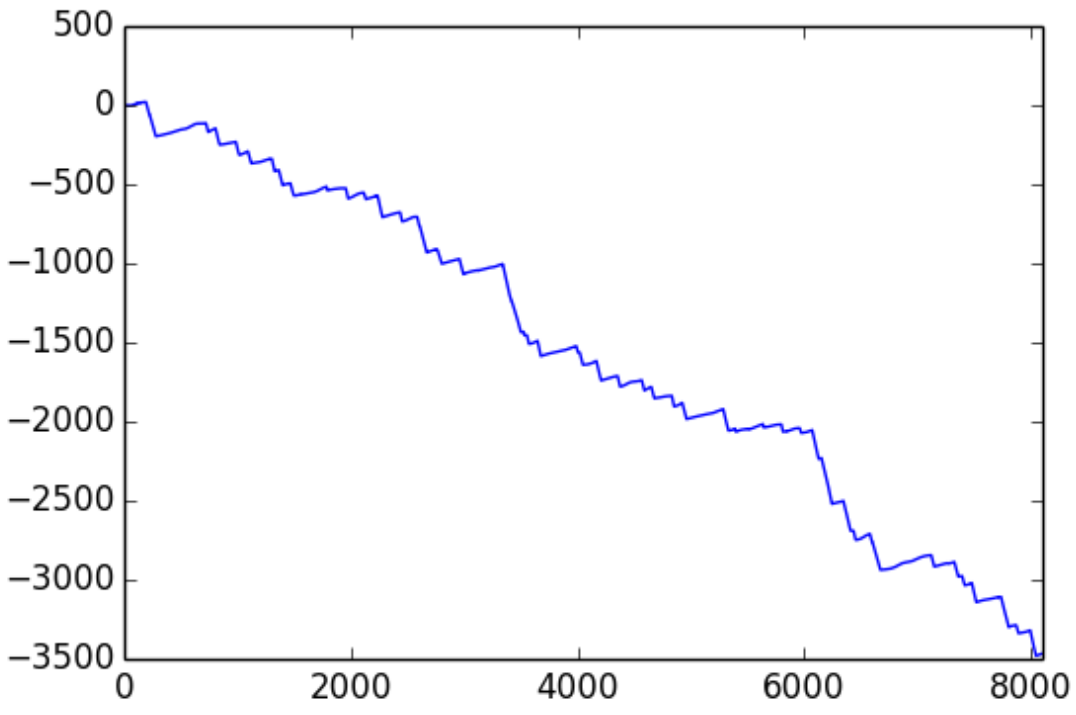

**(d)** Expert play

**Figure 4.3:** Cumulative reward

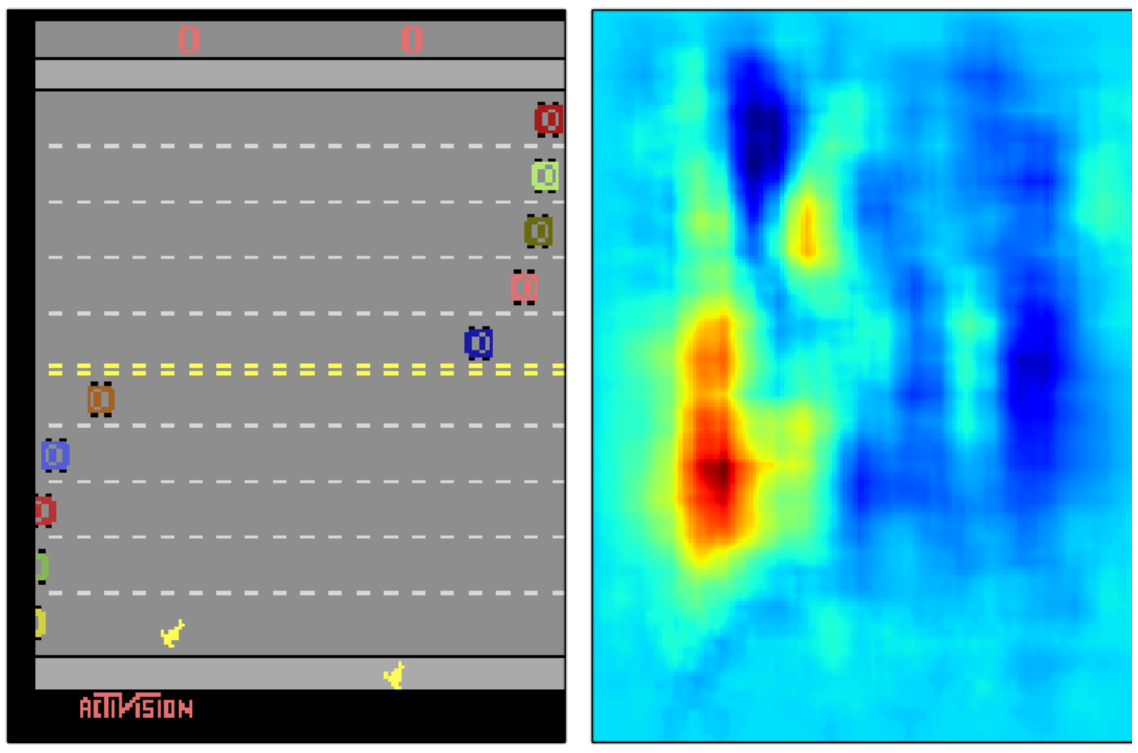

**(a)** Situation when the figure is at the bottom of the screen

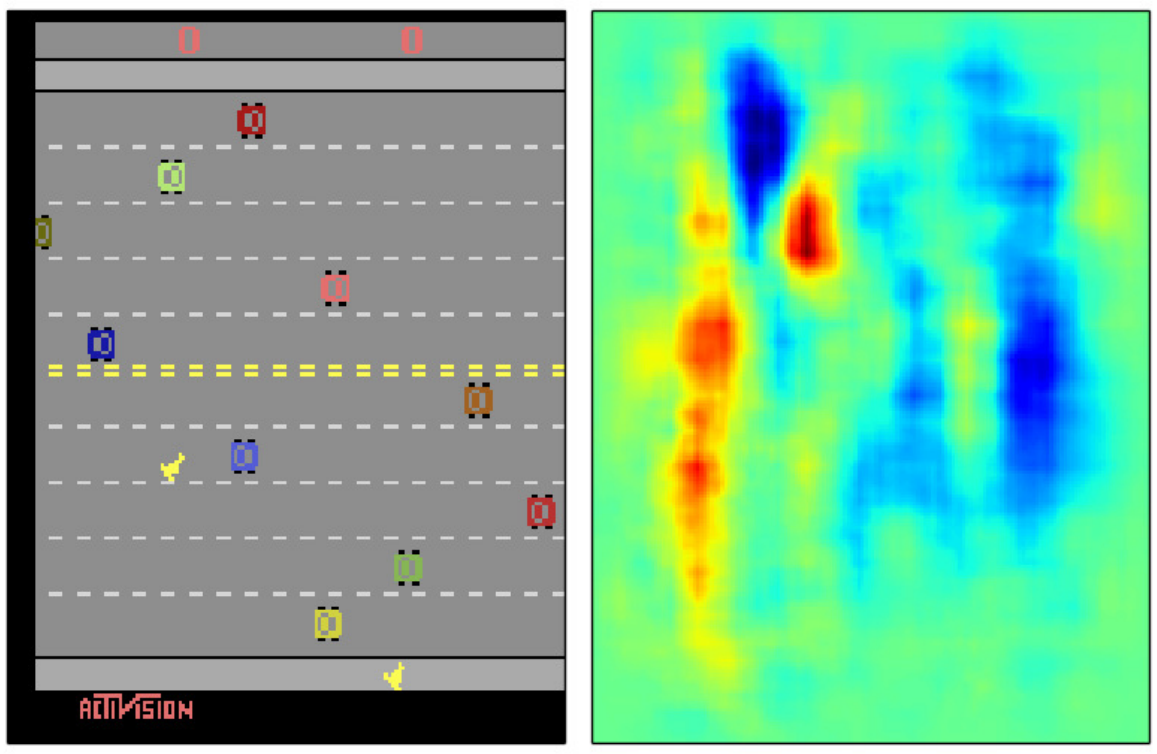

**(b)** Situation when the figure is at the middle of the screen shown in Figure 4.4, give us visualizations of the part of the input image which influence changes of the parameters during training, which in turn can be seen as patches of the image which are important for the learning task. Note that the colours pertain to how positive or negative the given value is, which is not important in our analysis. Instead the topology of these maps is important. Figure 4.4a shows a Freeway game state where the figure is on the bottom side of the highway respective to the screen, and is about the start crossing. In such a case, the cars in immediate proximity are headed in the direction from left to right, which means that the agent will need to evaluate that part of the image. As we can see in the Figure 4.4a the agent is showing exactly that behaviour, with the cars on his left being very important for the task compared to the further cars which are not relevant. In another case, when the figure is in the middle of the road it should change its behaviour and start looking on its right side since the cars are coming from that side. It can be noticed in Figure 4.4b that the agent is demonstrating the fact that the cars on the right are more important for the learning task compared to the part in the section of the road which is already crossed by the agent.

**Figure 4.4:** Influence maps for Freeway showing the importance of the objects in the frame in two situations

Therefore, we can conclude that the agent learnt to look left of the figure when crossing the first half of the road and accordingly, right, when crossing the rest of the highway. Such observation means that by using our method the features important for the learning task can be recognized. Although such a recognition is not ideal, possibly trying a different structure of the convolutional layers, or even a different number of convolutional layers may produce even better results.

CHAPTER 5

# Future Work and Conclusions

> *"We often think that when we have completed our study of one we know all about two, because 'two' is 'one and one'. We forget that we still have to make a study of 'and'. "*
>
> – Sir Arthur S. Eddington, *1928*

In this chapter we discuss plans for future work on this project, possible research directions for future projects, and conclusions drawn from our experience with the project so far. The work covered in this dissertation is only part of a bigger idea for doing apprenticeship learning. Therefore, a substantial portion of the discussion about future work will be directly related to implementing the other planned parts in order to offer a more complete apprenticeship learning approach. We start by discussion on reproducing the work of Mnih *et al.* (2013) and the reasons for embarking on such a task. We continue by introducing the concept of corrections as a phase that can probably improve this method even more. Then we put all that together into one unified apprenticeship learning cycle, which can be iteratively applied for even better results. We also discuss the importance of data in apprenticeship learning, and our goals in terms of increasing the amount of data and the number of games that the method is applied to. Furthermore, we discuss the future directions for this project. We start by a discussion on the possibility of applying such an approach in a much broader sense than only for playing games. Then we discuss forward models for distal supervised learning as well as recurrent networks and their possible capabilities and advantages if used in deep apprenticeship learning. We finish this chapter with conclusions drawn from our work.

## 5.1 Short-Term Planned Work

We will present several ideas which we plan to implement in the following few months. Note that, although they are an integral part of our idea for apprenticeship learning, due to the time constraint for this project, we were unable to implement them fully.

### 5.1.1 Reproducing DeepMind

In our method, after retrieving the reward function we obtain policy in a greedy manner by choosing the actions that produce the highest rewards. This approach, however, has weaknesses, as were already discussed in Section 2.2. Therefore, a better approach would be using the deep reinforcement learning method (Mnih *et al.* , 2013).Consequently, one important task that we plan to undertake in the near future is to reproduce the work of DeepMind. Since by obtaining a reward function our problem transforms into a reinforcement learning task, their approach seems reasonable for that task. There are a few reasons for replicating the work of DeepMind. First of all, we want to juxtapose our results to theirs. Since the closest research in spirit to this project is that of Mnih *et al.* (2013), our goal is to compare both approaches, which will effectively be a comparison between the reinforcement and the apprenticeship learning approach for the same application. We are aware of the fact that apprenticeship learning is probably a harder problem because the reward function is not given, but the idea is to see if it is possible to attain near optimal results by knowing the expert behaviour compared to the reinforcement learning approach, where the reward is known in advance, but no information on expert behaviour is given. We also want to analyse if using the deep reinforcement learning approach after we obtain the reward function will result in better policy compared to the greedy policy that we create based on the reward function itself.

The second phase in this direction is to try to improve the approach of DeepMind even more. Firstly, we want to do a comprehensive analysis of the Deep Q-Network. After such an analysis, if possible, we might change some aspects of their approach to improve the performance, such as: trying a different structure, trying a pre-training phase, and using AdaGrad for training[1]. One possible flaw in their approach is that by using Experience Replay to create the input sequence, the states are actually sampled randomly. The problem is that some states might be much more important for the cumulative reward of the sequence compared to other states, but the probability of including such states in the sequence is the same as the probability for including not desirable states. Therefore, a more intelligent approach to the sampling step must be employed. One such method is to use Bayesian Optimization (see Murphy, 2012), so that more important states get higher probability for sampling. This is an instance of the exploration versus exploitation problem as described in Section 2.2.6.

### 5.1.2 Correction of Actions

After successful reproduction (and possible improvement) of the DeepMind method (Mnih *et al.* , 2013), we want to implement a correction phase. In this phase, the agent behaviour will be corrected when it plans to execute actions from a given state, which are not desired by the expert policy. To better explain the necessity of such phase, we will give a short example of how humans learn tasks. Let us suppose that a child needs to learn how to play football. In a nutshell, the process starts in such a way that the trainer shows the child

[1] We are not sure if they use a different algorithm than the stochastic gradient descent for training, since such information is not given in their paper

the basic moves that need to be learned. Then the child tries to execute the same moves. The process when the child tries to recreate the trainer's moves can be compared to an imitation learning task. Moreover, the process of demonstration by the trainer followed by imitation by the child can be compared to an apprenticeship learning task. However, the child will probably do some moves differently than the trainer. Therefore, the trainer needs to correct such moves. Note that the trainer would not start the learning procedure from the very beginning, by repeated demonstration of all the previously shown moves, but rather, he/she will try to explain to the child only the moves that are done differently and will demonstrate those moves again. This can be seen as correction to the initially learned policy. Therefore, it seems natural to exploit such an approach in apprenticeship learning tasks as well. Inspired by the explained behaviour in how humans learn, we discuss two different methods for implementing corrections in the learning process.

The first, more rudimentary way, is to add a fixed number of corrected state-action pairs in the initial dataset, when an agent makes a mistake. We can identify a mistake as a situation where the agent which follows the learned policy executes an action for a given state which is not desirable by the expert policy. The new state-action pairs will reflect the most beneficial action from the given state. The problem here is to choose how many such pairs should be added. Since such a decision depends on the size of the dataset, as well as on the specific state-action pair, it is hard to efficiently implement the correction step using this approach. Although this method may work to some extent, a more sophisticated approach would be to fit Gaussian Process over the reward function. When working with neural networks, one (or few) examples can hardly change the training process so that it reflects the knowledge of those examples. This however, is not the case with Gaussian Processes, which will change not only for that one (or few) state-action pairs, but also for the nearest neighbourhood as well.

### 5.1.3 Apprenticeship Learning Cycle

We can use both Deep Q-Network (DQN) and the correction phase as described above to create a cycle so that the apprenticeship learning process can be executed iteratively many times for obtaining better policy. This cycle will start by using DAQN to obtain the $Q$ function. Then, DARN network is used to retrieve reward function $R$, which is input to the DQN so that a better $Q$ function can be learned. Next, the policy is extracted from the newly learned $Q$ function, followed by a correction process where it is corrected to reflect the behaviour of the expert policy. The modified dataset is used as input to the DAQN network, and the whole procedure is repeated iteratively. In every iteration, the policy converges more toward the expert policy, and the reward function will be more generalized. Note that for this claim we do not provide mathematical justification, but rather we just present our intuition for such an approach. This so called *Apprenticeship Learning Cycle* is given in Figure 5.1.

### 5.1.4 Data and Games

In terms of the games which our approach is applied to, we will first try to create expert data for all the games that Mnih *et al.* (2013) used in their research, so we are able to compare both of the approaches as discussed above. In terms of data, we will create more expert episodes for the games we already use. Note that although this dissertation covers only three games which our approach is applied to, we have expert data for two other Atari games, which are not included here because of problems with the ALE as discussed in Section 3.1.5. Of highest importance in the near future is to experiment with how much data is needed for a successful apprenticeship learning approach in deep settings. We want to investigate the relation between the amount of expert data available and the performance of the model. The result will be that we can estimate, for games with different complexity, how much data is needed and if that amount of data is (relatively) easy to collect. Another important task may involve the creation of a new game for our purposes on which the level of complexity can be customized. By creating such a game, we can experiment on the data needed in terms of the complexity of the game in order to achieve satisfactory results.

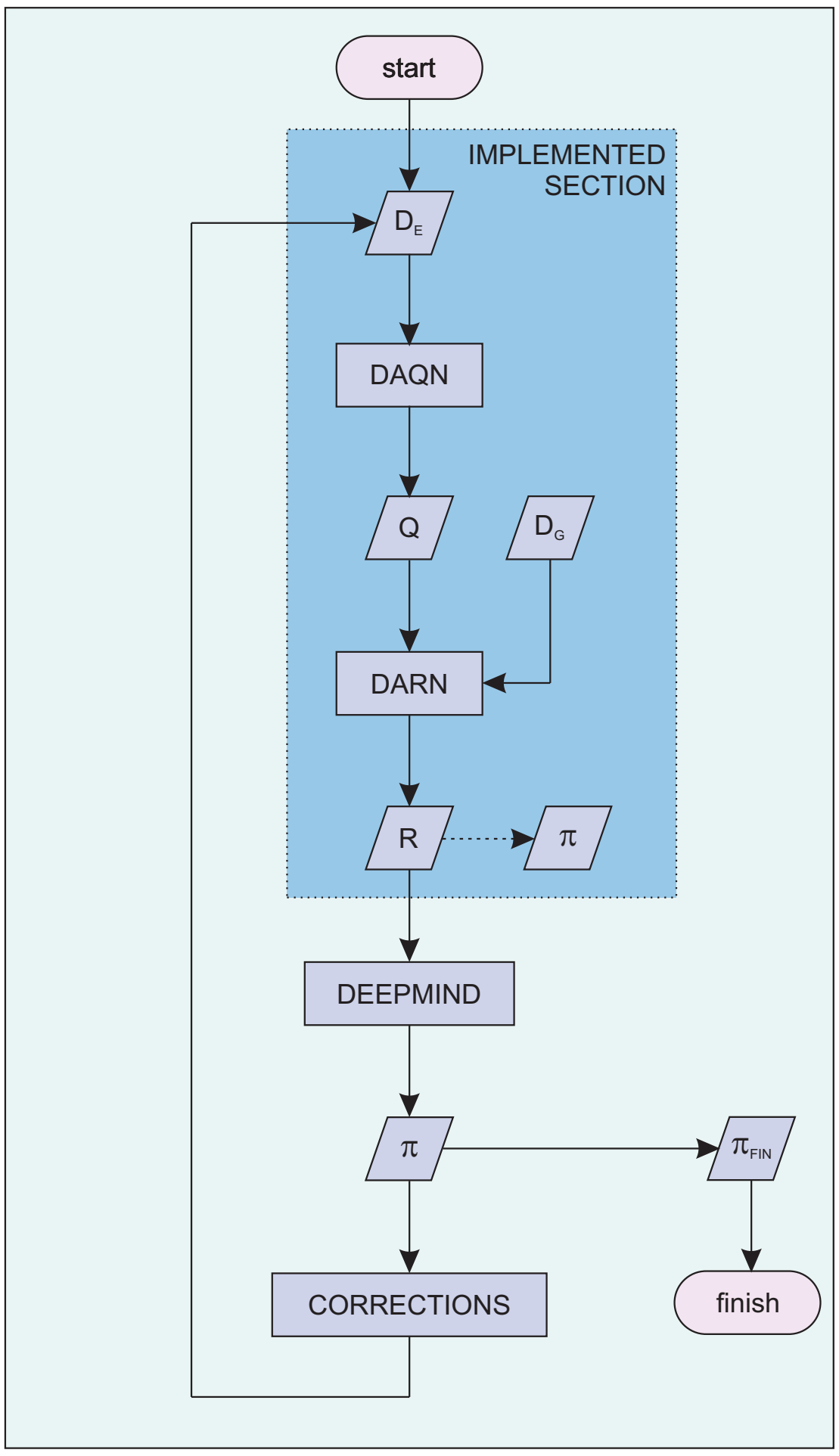

**Figure 5.1:** Deep Apprenticeship Learning Cycle

## 5.2 Future Directions

### 5.2.1 Broader Application

In our project we apply apprenticeship learning for the game playing domain. Although it may seem that such a task is complex, our application domain is much simpler compared to many other potential application areas. For example, apprenticeship learning can be applied for tasks such as driving a car. In that direction the apprenticeship learning approach can be used for many similar applications, such as learning tasks related to flying planes or helicopters (Abbeel, 2008). As Google already created autonomous, self-driving cars (Fisher, 2013), it would be interesting to see if such an approach can improve their product. Since they claimed that their cars drove more than 700,000 miles in total (Urmson, 2014), if video footage for those trials is available, an apprenticeship learning approach can have a high potential.

Learning by demonstration is probably one of the sub-goals in order to fully implement artificial intelligence. Therefore, apprenticeship learning methods can be used for many tasks that involve robots, which need to learn to perform some process or task if expert knowledge in the form of visual data is available for that task. Ultimately, if this approach will make a progress in the years (or decades) to follow, it can be used for general-purpose robots, which will be able to learn to perform any task, by observing an expert for that task. Although such a goal is far from attainable at this moment, theoretically, it can be achieved if progress is made in the apprenticeship learning paradigm as well as in computer vision and technology in general.

### 5.2.2 Forward Models and Distal Supervised Learning

Next, we discuss one possible extension of apprenticeship learning based on the forward models introduced by Jordan & Rumelhart (1992). Regular feed-forward neural networks have given dataset consisting of pairs of input and desired output. The idea is to learn parametrized function such that when applied on the input it will yield the desired output (or a close approximation of it). This function is non-linear in its parameters, which are the weights in the neural network. We use such an approach for obtaining the action given input state as defined in our problem. In these settings, we can specify a high level goal for the problem which can be achieved by a sequence of states and actions[2]. If we use such a formulation, the idea is to retrieve a sequence of states and actions that achieve the goal. Therefore, the action will depend not only on the current state, but also from the high level goal that needs to be achieved. In such a way, we can generalize our problem to be able to learn different strategies for learning a task in one model. To transfer the discussion into the apprenticeship learning paradigm, if we have expert trajectories for different strategies of learning a task, we can create a general model that will learn how to achieve the task based on the strategy that needs to be followed. Note that, those high level strategies can also be sub-routines of the task. For example in a given game, instead of directing each step of the agent we may define goals such as ‘reach given object’, or ‘avoid some enemy’, without specifying the low level details to perform that sub-routine. Then the final goal may be described by a sequence of sub-routines.

Jordan & Rumelhart (1992) present a model of learning, called learning with distal teacher, that enables learning sub-routines as part of one main routine. They present *forward models*, which take as input state and action and produce the next state. After such a model is trained, it is used as a component in an approach called *distal supervised learning*, where the agent first learns the forward model, followed by learning a mapping from states and plans (routines or sub-goals) to actions. Such a network is given in Figure 5.2. Note that each of the components can be represented as neural network with few layers, if needed. In case of the state components, when training the forward model, in our case the network is a convolutional network with layers as described in Chapter 3, in order to extract feature representations. When the forward model is used as a component, only the higher layer (the feature representation of a state) is taken. The only exception is the initial state, which is given as raw input (video frames for example) to the model, and needs to be passed through the convolutional structure in order to obtain abstract feature representation, corresponding to the last layer of the forward model as described above. Note that $s[t]$ denotes the state at time $t$, $a[t]$ denotes the action at time $t$, and $g[t]$ is the sub-goal at time $t$. Usually, the sub-goal does not change in each time step. This structure can be seen as a network with $g$ and $s[0]$ as input that produce sequence of actions $a[t]$ in order to achieve the sub-goal $g$ when the initial state is given by $s[0]$.

[2] Note that there may be more than one possible sequences that achieve the same goal

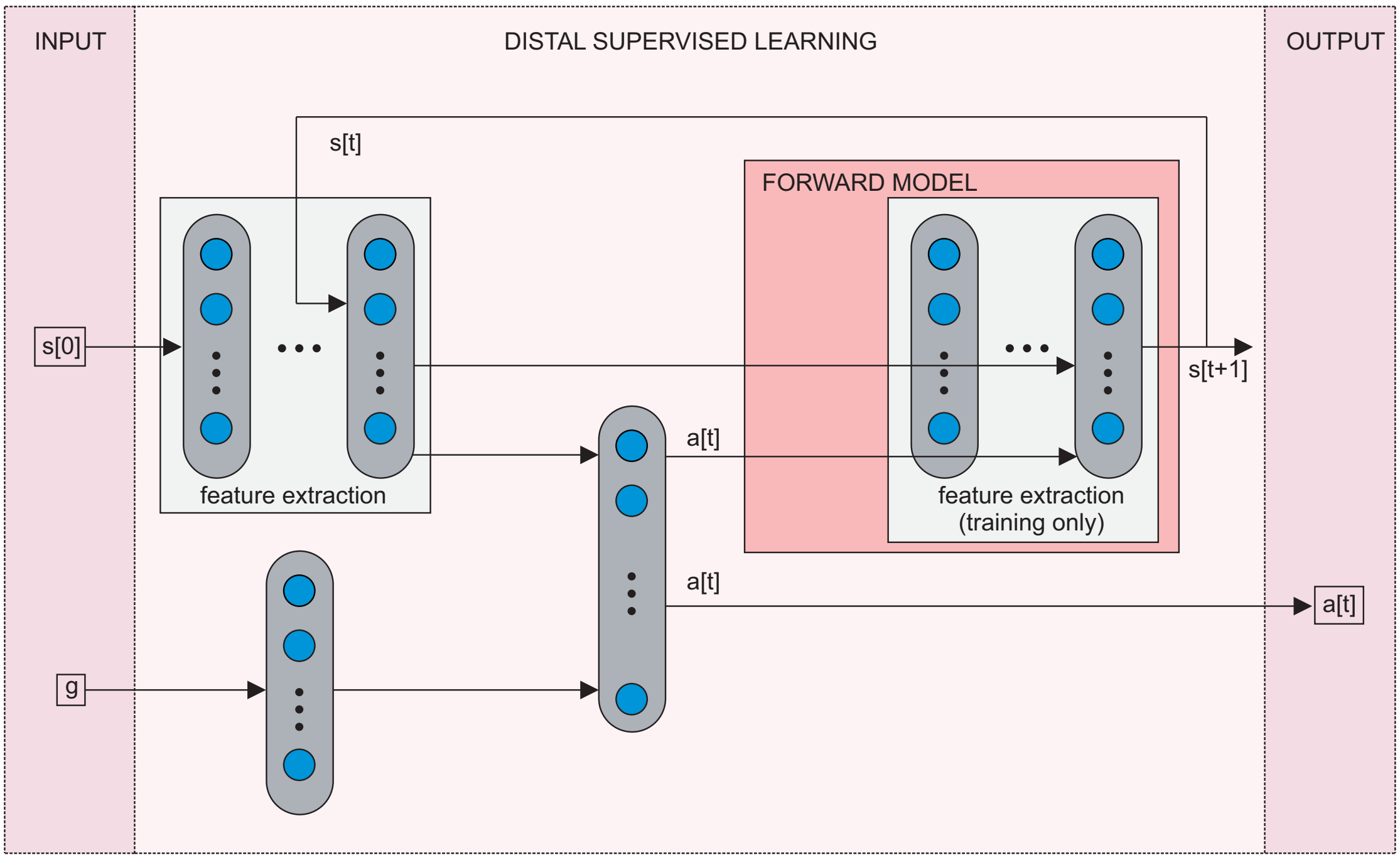

**Figure 5.2:** Deep Apprenticeship Learning via Distal Supervised Learning

### 5.2.3 Recurrent Networks

Another approach for learning sequences of actions that achieve sub-goals is by using recurrent neural networks. Unlike standard feed-forward networks, recurrent networks have directed cycle connections. Because of that structure, they are very suitable for problems that can be represented as a function of time. For our case, in every time step, one action is performed from a given state. This means that such a network can be used so that the output action at time $t$ is the action that needs to be performed at time step $t$. One such type of network, which has had success in the last few years for different tasks, are called Long Short Term Memory (LSTM) (Hochreiter & Schmidhuber, 1997). The biggest advantage of this type of network is that it solves the gradient vanishing problem, which is not the case for regular recurrent networks. For recurrent networks, variation of the backpropagation algorithm is used, where the gradients are also propagated through time. For using the LSTM method in the reinforcement learning paradigm see Bakker (2002). In terms of our application, with a recurrent network the model will learn sequences of actions that achieve the goal (or sub-goals) when starting from a given state. Recurrent networks are becoming really popular in the machine learning society, therefore investigating such an approach will be one of our main objectives in the future.

## 5.3 Conclusions

The contribution of our work is that the apprenticeship learning approach is used for the first time for learning tasks with complex representations. In spite of the fact that this project is based on the previous research on inverse reinforcement and imitation learning, by combining those approaches we avoid their weaknesses and show that our method can be applied for complex tasks such as learning to play games by having only the video frames of an expert.

The combination of Atari games as an application area and ALE as a development environment can be an excellent choice for evaluating apprenticeship learning methods, based on the fact that the games for Atari are relatively simple and the development in ALE is well structured and straightforward. One important fact that should be taken into account in respect to this discussion, is that ALE is best exploited when the development is in C++. Since most machine learning projects use Python as an unofficial standard for programming, ALE may not be the best choice. There are two possible solutions to alleviate this problem. One is to write a Python wrapper for the newest version of ALE, and the other is to create a customized testbed for evaluating machine learning methods applied to console games. As one might see, both approaches require substantial time for programming. It may be a good idea for a standardized environment to be created that supports Python and which can be used by many researchers to evaluate and compare their methods for reinforcement and apprenticeship learning.

In terms of training of the neural network, standard stochastic gradient methods are not effective for our approach. Therefore, one way to overcome that problem is by using the AdaGrad method instead. While pre-training is demonstrated to be effective for deep neural networks, in our case such an approach produced worse results. One contradictory conclusion, compared to the majority of research in the area, is that using hyperbolic tangent non-linearity leads to better results compared to rectifier non-linearity. We suspect that it might be the case that our network is not very complex, so there is no need for rectifier for successful training.

As reported in Chapter 4, our method outperforms all other previously used methods for the game Freeway. If we take into account that we are not assuming knowledge of reward function nor exact information of the internal game space, but only expert trajectories which can be relatively easier to collect compared to other data sources, the significance of our research increases drastically. Although we did not have success in training more complex games, such as Space Invaders and Seaquest, if we leave out the problems with the emulator, we are confident that having more data and trying different architectures for the neural networks can lead to fruitful results even for such complex games.

In terms of comparison with different types of agents, we showed that for simple games, such as Freeway, other strategies such as the random one or SARSA can be significantly outperformed. The human expert score, however, cannot be outperformed but our agent still achieves a score that is at least comparable to the expert's which is still a huge success. We suspect that different network structures, more data, and longer training using the method proposed in our work can result with a cumulative reward higher than the human's score.

Even though our model cannot learn the optimal (or near-optimal) policy, it is evident that the learning process is successful to some extent. The main question should not be if learning a given task is possible in these settings, but rather, how much expert data is needed for such learning to be effective. The problem of data sparsity will diminish in the future, taking into account the fact that technological advancement and the digital era contribute to more data available for such use. Therefore, as years pass, the plausibility of successful application of apprenticeship learning methods will increase even more.

The visualizations of the gradients of the output in terms of the input, shows us that important features can be indeed learnt successfully. Such observation is very important since we can be confident that our method does recognize important features of the input and use them to model the learning process.

Although the cumulative rewards obtained as described in Chapter 4 are not ideal, we can conclude that the reward extraction step is well designed since the results obtained correspond to the expected results from the learned policy using the score function. Therefore, we are confident that combination of more data, trying different network structures, and completing the deep apprenticeship learning cycle can produce state-of-the-art results which can be successfully used in many applications where other types of methods are unsuitable.

The reinforcement learning was motivated by the fact that some tasks are not natural to be represented as supervised or unsupervised learning problems. Although apprenticeship learning relies on the theory of reinforcement learning, the same parallel can be drawn between these two approaches as well. Some learning tasks cannot be represented as reinforcement learning problems. Therefore, it is crucially important to invest time (and resources) in apprenticeship learning research, since it can lead to breakthroughs in scientific research. Ultimately, apprenticeship learning can be considered as a possible step that can contribute toward achieving the goal of artificial intelligence.

***We know very little, and yet it is astonishing that we know so much, and still more astonishing that so little knowledge can give us so much power.***

- Bertrand Russell, third Earl Russell

# Appendix A

# Code

*"Programs must be written for people to read, and only incidentally for machines to execute."*

– Harold Abelson, *1996*

What follows is the most relevant portion of the code which is needed in order to replicate our work. The whole code is included in the DVD accompanying this dissertation.

```
import numpy as np

from conv_layer import LeNetConvPoolLayer
from logistic_sgd import LogisticRegression
from mlp import HiddenLayer

import theano.tensor as T
import theano

def relu(x):

    return T.switch(T.lt(x, 0), 0, x)

def identity(x):

    return x

class DeepMindNetwork:

    HIDDEN_LAYER_SIZE = 256

```

```
    def __init__(
        self, x, num_actions, batch_size, rng, history_size, non_linearity,
        W0 = None, b0 = None,
        W1 = None, b1 = None,
        W2 = None, b2 = None,
        W3 = None, b3 = None):

        self.layer0 = LeNetConvPoolLayer(
            rng = rng,
            input = x,
            filter_shape = (16, history_size, 8, 8),
            image_shape = (batch_size, history_size, 83, 83),
            poolsize = (4, 4),
            W = W0,
            b = b0,
            relu = (non_linearity == 'relu'))

        self.layer1 = LeNetConvPoolLayer(
            rng = rng,
            input = self.layer0.output,
            filter_shape = (32, 16, 4, 4),
            image_shape = (batch_size, 16, 19, 19),
            poolsize = (2, 2),
            W = W1,
            b = b1,
            relu = (non_linearity == 'relu'))

        layer2_input = self.layer1.output.flatten(2)

        if non_linearity == 'relu':
            activation = relu
        else:
            activation = T.tanh
        self.layer2 = HiddenLayer(
            rng = rng,
            input = layer2_input,
            n_in = 2048,
            n_out = DeepMindNetwork.HIDDEN_LAYER_SIZE,
            activation = activation,
            W = W2,
            b = b2)

        self.layer3 = LogisticRegression(
            input = self.layer2.output,
            n_in = DeepMindNetwork.HIDDEN_LAYER_SIZE,
            n_out = num_actions,
            W = W3,
            b = b3)
```

```
        self.params = self.layer3.params + self.layer2.params + self.layer1.params + self.layer0.params

    def saveToFile(self, fileName):
        np.savez_compressed(
            fileName,
            W0 = self.layer0.W.get_value(),
            b0 = self.layer0.b.get_value(),
            W1 = self.layer1.W.get_value(),
            b1 = self.layer1.b.get_value(),
            W2 = self.layer2.W.get_value(),
            b2 = self.layer2.b.get_value(),
            W3 = self.layer3.W.get_value(),
            b3 = self.layer3.b.get_value())

    @staticmethod
    def readFromFile(x, num_actions, batch_size, rng, fileName, history_size, non_linearity):
        W0 = np.load(fileName)['W0']
        b0 = np.load(fileName)['b0']
        W1 = np.load(fileName)['W1']
        b1 = np.load(fileName)['b1']
        W2 = np.load(fileName)['W2']
        b2 = np.load(fileName)['b2']
        W3 = np.load(fileName)['W3']
        b3 = np.load(fileName)['b3']
        return DeepMindNetwork(
            x, num_actions, batch_size, rng, history_size, non_linearity,
            W0 = theano.shared(np.asarray(W0, dtype=theano.config.floatX), borrow=True),
            b0 = theano.shared(np.asarray(b0, dtype=theano.config.floatX), borrow=True),
            W1 = theano.shared(np.asarray(W1, dtype=theano.config.floatX), borrow=True),
            b1 = theano.shared(np.asarray(b1, dtype=theano.config.floatX), borrow=True),
            W2 = theano.shared(np.asarray(W2, dtype=theano.config.floatX), borrow=True),
            b2 = theano.shared(np.asarray(b2, dtype=theano.config.floatX), borrow=True),
            W3 = theano.shared(np.asarray(W3, dtype=theano.config.floatX), borrow=True),
            b3 = theano.shared(np.asarray(b3, dtype=theano.config.floatX), borrow=True))


    @staticmethod
    def compareNetworkParameters(net1, net2):
        if not np.array_equal(net1.layer0.W.get_value(), net2.layer0.W.get_value()):
            return False
        if not np.array_equal(net1.layer0.b.get_value(), net2.layer0.b.get_value()):
            return False
        if not np.array_equal(net1.layer1.W.get_value(), net2.layer1.W.get_value()):
            return False
        if not np.array_equal(net1.layer1.b.get_value(), net2.layer1.b.get_value()):
            return False
        if not np.array_equal(net1.layer2.W.get_value(), net2.layer2.W.get_value()):
```

```
            return False
        if not np.array_equal(net1.layer2.b.get_value(), net2.layer2.b.get_value()):
            return False
        if not np.array_equal(net1.layer3.W.get_value(), net2.layer3.W.get_value()):
            return False
        if not np.array_equal(net1.layer3.b.get_value(), net2.layer3.b.get_value()):
            return False
        return True
```

code/deepmind_network.py

```
import theano
import theano.tensor as T
import numpy as np

from mlp import HiddenLayer
from convolutional_mlp import LeNetConvPoolLayer
from theano import shared
from logistic_sgd import LogisticRegression

from deepmind_network import DeepMindNetwork
import deepmind_static_training as dst

class DoubleNetwork:

    def __init__(self, num_actions, gama, batch_size, rng, history_size, learning_rate = 0.1):
        self.batch_size = batch_size

        curr_state = T.tensor4('curr_state')
        next_state = T.tensor4('next_state')
        reward = T.lvector('reward')
        action = T.lvector('action')
        terminal = T.lvector('terminal')

        small_curr_state = T.tensor4('small_curr_state')

        self.net1 = DeepMindNetwork(
            x = curr_state, num_actions = num_actions,
            batch_size = batch_size, rng = rng,
            history_size = history_size,
            W0 = None, b0 = None,
            W1 = None, b1 = None,
            W2 = None, b2 = None,
            W3 = None, b3 = None)

        net2 = DeepMindNetwork(
            x = next_state, num_actions = num_actions,
```

```
        batch_size = batch_size, rng = rng,
        history_size = history_size,
        W0 = self.net1.layer0.W, b0 = self.net1.layer0.b,
        W1 = self.net1.layer1.W, b1 = self.net1.layer1.b,
        W2 = self.net1.layer2.W, b2 = self.net1.layer2.b,
        W3 = self.net1.layer3.W, b3 = self.net1.layer3.b)

    small_net = DeepMindNetwork(
        x = small_curr_state, num_actions = num_actions,
        batch_size = 1, rng = rng,
        history_size = history_size,
        W0 = self.net1.layer0.W, b0 = self.net1.layer0.b,
        W1 = self.net1.layer1.W, b1 = self.net1.layer1.b,
        W2 = self.net1.layer2.W, b2 = self.net1.layer2.b,
        W3 = self.net1.layer3.W, b3 = self.net1.layer3.b)

    y = T.switch(T.eq(terminal, 1), reward, reward + gama * T.max(net2.layer3.output, axis = 1))

    y_pred = self.net1.layer3.output[np.arange(batch_size), action]
    cost = T.sum(T.sqr(y - y_pred))

    q = self.net1.layer3.output[np.arange(batch_size), action]
    q_prime = T.max(net2.layer3.output, axis = 1)
    l = y - y_pred

    self.params = self.net1.layer3.params + self.net1.layer2.params + self.net1.layer1.params + self.net1.layer0.params

    updates = dst.sgd_updates_adagrad(cost = cost, params = self.params, learning_rate = learning_rate, consider_constant = [y])

    self.minibatch_update = theano.function(
        [curr_state, next_state, reward, action, terminal],
        cost,
        updates = updates,
        allow_input_downcast = True)

    self.best_action_for_state = theano.function([small_curr_state], T.argmax(small_net.layer3.output, axis = 1), allow_input_downcast=True)

    self.get_debug_info = theano.function(
        [curr_state, next_state, reward, action, terminal],
        [q, q_prime, reward, l],
        allow_input_downcast = True)
```

code/deepmind_training_graph.py

```python
import cPickle
import gzip
import os
import sys
import time

import numpy
import numpy as np

from logistic_sgd import LogisticRegression, load_data
from mlp import HiddenLayer

import os.path
from PIL import Image
from matplotlib import pyplot as plt
from scipy.misc import imresize
from matplotlib import colors as clrf
from collections import OrderedDict

import theano
import theano.tensor as T

import generalUtils

from deepmind_network import DeepMindNetwork
from    os                  import      mkdir
from    os.path             import      exists

import argparse

def loadEpisode(database, episodeIndex, history_size):
    episode_path = database + str(episodeIndex) + '.npz'
    return load_episode_from_path(episode_path, history_size)

def create_multiframe_states(states, history_size):
    #print states.shape
    multiframe_states = np.zeros((states.shape[0], history_size, states.shape[1], states.shape[2]))
    #print multiframe_states.shape
    for i in range(history_size):
        multiframe_states[(history_size - i - 1):, i, :, :] = states[:(states.shape[0] - history_size + i +
      1)]
    return multiframe_states

def load_episode_from_path(episode_path, history_size):
    episode = np.load(episode_path)['x']
    # To handle 4 times faster frame rate when recording expert play
    episode = episode[::4]
    background = np.median(episode, axis=0)
```

```
    #episode -= background
    episode /= 255.0

    # TODO: Use "create_multiframe_states" here.
    x = np.zeros((episode.shape[0], history_size, 83, 83))
    for i in range(history_size):
        x[(history_size - i - 1):, i, :, :] = episode[:(episode.shape[0] - history_size + i + 1)]

    y = np.load(episode_path)['y']
    y = y[::4]

    return [x, y]

def load_episode_and_shuffle(database, episode_index, history_size):
    [x, y] = loadEpisode(database, episode_index, history_size)

    indices = numpy.random.permutation(x.shape[0])
    x = x[indices]
    y = y[indices]

    return [x, y]

def sgd_updates_adagrad(cost, params, learning_rate, consider_constant = [], epsilon=1e-10):

    accumulators = OrderedDict({})
    e0s = OrderedDict({})
    learn_rates = []
    ups = OrderedDict({})
    eps = OrderedDict({})

    for param in params:
        eps_p = numpy.zeros_like(param.get_value())
        accumulators[param] = theano.shared(value=eps_p, name="acc_%s" % param.name)
        e0s[param] = learning_rate
        eps_p[:] = epsilon
        eps[param] = theano.shared(value=eps_p, name="eps_%s" % param.name)

    gparams = T.grad(cost, params, consider_constant)

    for param, gp in zip(params, gparams):
        acc = accumulators[param]
        ups[acc] = acc + T.sqr(gp)
        val = T.sqrt(T.sum(ups[acc])) + epsilon
        learn_rates.append(e0s[param] / val)

    updates = [(p, p - step * gp) for (step, p, gp) in zip(learn_rates, params, gparams)]

    return updates
```

```
def trainDeepMindNetwork(
    game,
    history_size,
    non_linearity,
    database,
    output_suffix,
    learning_rate=0.1,
    numEpochs=1000,
    batch_size=32,
    maxEpisodes=500,
    use_rica_filters = False):

    game_settings = generalUtils.get_game_settings(game)
    num_actions = len(game_settings.possible_actions)
    [x, y] = load_episode_and_shuffle(database, 0, history_size)

    testSetX = theano.shared(numpy.asarray(x, dtype=theano.config.floatX), borrow=True)
    testSetY = theano.shared(numpy.asarray(y, dtype='int32'), borrow=True)

    [x, y] = load_episode_and_shuffle(database, 1, history_size)
    trainSetX = theano.shared(numpy.asarray(x, dtype=theano.config.floatX), borrow=True)
    trainSetY = theano.shared(numpy.asarray(y, dtype='int32'), borrow=True)

    index = T.lscalar()
    x = T.tensor4('x')
    y = T.ivector('y')

    rng = numpy.random.RandomState(23455)

    if history_size == 1 and use_rica_filters:
        W0 = np.loadtxt('W.txt', dtype='float32')

        W0_reshaped = np.zeros((16, 1, 8, 8), dtype='float32')
        for i in range(16):
            W0_reshaped[i][0] = W0[i].reshape(8, 8).T

        shared_W0 = theano.shared(W0_reshaped, borrow = True)

        net = DeepMindNetwork(
            x = x,
            num_actions = num_actions,
            batch_size = batch_size,
            rng = rng,
            history_size = history_size,
            non_linearity = non_linearity,
            W0 = shared_W0)
```

```
    else:
        net = DeepMindNetwork(
            x = x,
            num_actions = num_actions,
            batch_size = batch_size,
            rng = rng,
            history_size = history_size,
            non_linearity = non_linearity)


    cost = net.layer3.negative_log_likelihood(y)

    params = net.params

    updates = sgd_updates_adagrad(cost = cost, params = params, learning_rate = learning_rate)

    train_model = theano.function([index], net.layer3.errors(y), updates=updates,
          givens={
            x: trainSetX[index * batch_size: (index + 1) * batch_size],
            y: trainSetY[index * batch_size: (index + 1) * batch_size]})

    test_model = theano.function([index], net.layer3.errors(y),
             givens={
                x: testSetX[index * batch_size: (index + 1) * batch_size],
                y: testSetY[index * batch_size: (index + 1) * batch_size]})

    output_path = game + '/' + non_linearity + str(history_size) + output_suffix + '/'

    if not exists(game):
        mkdir(game)
    if not exists(output_path):
        mkdir(output_path)

    for epochIndex in range(numEpochs):

        all_batch_scores = None

        for episodeIndex in range(1, maxEpisodes):
            if (os.path.isfile(database + str(episodeIndex) + '.npz')):
                print 'episode: ' + str(episodeIndex)

                start = time.time()
                [x, y] = load_episode_and_shuffle(database, episodeIndex, history_size)
                trainSetX.set_value(numpy.asarray(x, dtype=theano.config.floatX))
                trainSetY.set_value(numpy.asarray(y, dtype='int32'))

                end = time.time()
                print 'time for loading episode: ' + str(end - start) + 's'
```

```
                start = time.time()
                trainSetSize = trainSetX.get_value(borrow=True).shape[0]

                batch_scores = [train_model(minibatchIndex)
                    for minibatchIndex in xrange(trainSetSize / batch_size)]

                if all_batch_scores == None:
                    all_batch_scores = batch_scores
                else:
                    all_batch_scores = np.append(all_batch_scores, batch_scores)

                end = time.time()
                print 'time for minibatch updates: ' + str(end - start) + 's'
                print ''

        net.saveToFile(output_path + str(epochIndex))

        # Testing saving and loading of the network

        #loadedNetwork = DeepMindNetwork.readFromFile(
        #    x = x, num_actions = num_actions, batch_size = batch_size, rng = rng,
        #    fileName = 'data/network' + str(epochIndex) + '.npz')
        #print 'comparison of networks: ' + str(DeepMindNetwork.compareNetworkParameters(net, loadedNetwork
    ))

        training_score = np.mean(all_batch_scores)
        training_file = open(output_path + 'training.txt', 'a+')
        training_file.write(str(training_score) + '\n')
        training_file.close()

        testSetSize = testSetX.get_value(borrow=True).shape[0]
        validationScore = numpy.mean([test_model(minibatchIndex)
            for minibatchIndex in xrange(testSetSize / batch_size)])
        validation_file = open(output_path + 'validation.txt', 'a+')
        validation_file.write(str(validationScore) + '\n')
        validation_file.close()

if __name__ == '__main__':
    parser = argparse.ArgumentParser(description='Generates gameplay from a given network')
    parser.add_argument('-g','--game', help='Which game', required=True)
    parser.add_argument('-s','--history_size', help='Number of frames in input', required=True, type = int)
    parser.add_argument('-n','--non_linearity', help='tanh or ReLU', required=True)
    parser.add_argument('-d','--database', help='Human gameplay database', required=True)
    parser.add_argument('-o','--output_suffix', help='Suffix for output folder', required=True)


    args = vars(parser.parse_args())
```

```
trainDeepMindNetwork(args['game'], args['history_size'], args['non_linearity'], args['database'], args[
 'output_suffix'])
```

code/deepmind_static_training.py

```
from    numpy               import      *
import numpy as np
from    sys                 import      exit
from    pylab               import      *
from    os                  import      mkdir
from    os.path             import      exists
from    random              import      randint
from    player_agent        import      PlayerAgent
from    samples_manager     import      SamplesManager
from    run_ale             import      run_ale
from    common_constants    import      actions_map
from    game_settings       import      AstrixSettings, SpaceInvadersSettings, \
                                        FreewaySettings, SeaquestSettings, BreakoutSettings
from memory_database import MemoryDatabase
from scipy.misc import imresize
from deepmind_network import DeepMindNetwork
import generalUtils
import theano
import theano.tensor as T

from scipy import stats

import argparse


class DeepMindAgent(PlayerAgent):
    """
        A player agent that acts randomly in a game, and gathers samples.
        These samples are later used to detect the game background, and
        also detect object classes.

        Instance Variables:
            - num_samples       Number of samples to collect
            - act_contin_count  Number of frames we should repeat a
                                randomly selected action
            - samples_count     Number of samples we have collected
                                so far
            - reset_count_down  sometimes we need to send reset
                                action for a number of frames
```

```
            - curr_action       The action taken in the previous step
            - curr_action_count Number of times we have taken this action
            - samples_manager   Instance of SampleManager class, responsible
                                for gathering samples
            - rand_run_only     When true, we will just do a random-run, without
                                actually gathering any samples

    """
    def __init__(self, game_settings, which_network, plot_histogram, non_linearity, play_for_frames,
                 action_continuity_count, epsilon, working_directory = ".",
                 rand_run_only = False, uniform_over_game_space = False,
                 greedy_action_selection = False, history_size = 4, batch_size = 32):
        PlayerAgent.__init__(self, game_settings, working_directory)
        self.color_map = generalUtils.generateColorMap()
        #TODO: Update this, to properly support the new GameSettings framework
        self.act_contin_count   = action_continuity_count
        self.curr_action        = None
        self.curr_action_count  = 0
        self.restart_delay      = 0
        self.initial_delay      = 100
        self.episode_status     = 'uninitilized'
        self.episode_counter    = 0
        self.episode_reward     =0
        self.curr_state             = []
        self.frame_count        = 0
        self.database          = MemoryDatabase(history_size = history_size)
        self.uniform_over_game_space = uniform_over_game_space
        self.greedy_action_selection = greedy_action_selection
        self.save_next_frame = False
        self.prev_state = None
        self.prev_action = None
        self.influence = np.zeros((256, 83, 83))
        self.history_size = history_size
        self.batch_size = batch_size
        self.batch = np.zeros((self.batch_size, self.history_size, 83, 83))
        self.plot_histogram = plot_histogram

        x = T.tensor4('x')
        y = T.ivector('y')
        rng = np.random.RandomState(23455)

        net = DeepMindNetwork.readFromFile(
            x = x,
            num_actions = len(game_settings.possible_actions),
            batch_size = self.batch_size,
            rng = rng,
            fileName = which_network,
            history_size = self.history_size,
```

```
            non_linearity = non_linearity)

        self.get_action = theano.function(
            [x],
            net.layer3.y_pred[0])

        self.get_probs = theano.function(
            [x],
            net.layer3.p_y_given_x[0]
        )
        self.epsilon = epsilon

        #self.background = np.load('data/background0.npz')['background']
        self.a = []

        i = T.iscalar('i')

        self.get_influence = theano.function([x, i], T.grad(net.layer3.p_y_given_x[0][i], x),
    allow_input_downcast=True)

        #self.get_influence = theano.function([x, i], T.grad(net.layer3.p_y_given_x[0][i], x),
    allow_input_downcast=True)           self.get_influence = theano.function([x, i], T.grad(net.layer3.
    p_y_given_x[0][i], x), allow_input_downcast=True)
        self.get_outputs = theano.function([x, i], net.layer1.output[0][i], allow_input_downcast=True)


    def export_image(self, screen_matrix, filename):
        "Saves the given screen matrix as a png file"
        try:
            from PIL import Image
        except ImportError:
            exit("Unable to import PIL. Python Image Library is required for" +
                 " exporting screen matrix to PNG files")
        plot_height, plot_width = screen_matrix.shape[0], screen_matrix.shape[1]
        rgb_array = zeros((plot_height, plot_width , 3), uint8)
        counter = 0
        for i in range(plot_height):
            for j in range(plot_width):
                rgb_array[i,j,0] = screen_matrix[i, j]
                rgb_array[i,j,1] = screen_matrix[i, j]
                rgb_array[i,j,2] = screen_matrix[i, j]
        pilImage = Image.fromarray(rgb_array, 'RGB')
        pilImage.save(filename)

    def export_color_image(self, screen_matrix, filename):
        "Saves the given screen matrix as a png file"
        try:
            from PIL import Image
```

```
        except ImportError:
            exit("Unable to import PIL. Python Image Library is required for" +
                 " exporting screen matrix to PNG files")

        pilImage = Image.fromarray(screen_matrix, 'RGB')
        pilImage.save(filename)

    def scale_array(self, rawpoints):

        high = 256.0
        low = 0.0

        mins = np.min(rawpoints)
        maxs = np.max(rawpoints)
        rng = maxs - mins

        scaled_points = high - (((high - low) * (maxs - rawpoints)) / rng)

        return scaled_points

    def add_to_batch(self, new_frame):
        # Shift previous frames
        self.batch[0, :self.history_size - 1] = self.batch[0, 1:]
        # Add new frame
        self.batch[0, self.history_size - 1] = new_frame

    def agent_step(self, screen_matrix, console_ram, reward = None):
        """
            The main method. Given a 2D array of the color indecies on the
            screen (and potentially the reward recieved), this method
            will decides the next action based on the learning algorithm.
            Here, we are using random actions, and we save each new
        """
        # See if we ar in the inital-delay period.
        if self.initial_delay > 0:
            # We do nothing, until the game is ready to be restarted.
            self.initial_delay -= 1
            print "Initial delay:", self.initial_delay
            return actions_map['player_a_noop']

        # At the very begining, we have to restart the game
        if self.episode_status == "uninitilized":
            if self.game_settings.first_action is not None:
                # Perform the very first action next (this is hard-coded)
                self.episode_status = "first_action"
            else:
                self.episode_status = "ended"
            self.restart_delay = self.game_settings.delay_after_restart
```

```
            return actions_map['reset']

        # See if we are in the restart-delaying state
        if self.restart_delay > 0:
            print "Restart delay:", self.restart_delay
            self.restart_delay -= 1
            return actions_map['player_a_noop']

        # See if we should apply the very first action
        if self.episode_status == "first_action":
            print "Sending first action:", self.game_settings.first_action
            self.episode_status = 'ended'
            return actions_map[self.game_settings.first_action]

        terminal = 0
        # See if we are the end of the game
        if self.game_settings.is_end_of_game(screen_matrix, console_ram):
            terminal = 1
            # End the current episode and send a Reset command
            print "End of the game. Restarting."
            if self.game_settings.first_action is not None:
                self.episode_status = "first_action"
            else:
                self.episode_status = "ended"
            self.restart_delay = self.game_settings.delay_after_restart
            return actions_map['reset']

        if reward is None:
            reward = self.game_settings.get_reward(screen_matrix, console_ram)
        self.episode_reward += reward

        if  self.episode_status == 'ended':
            print "Episode #%d: Sum Reward = %f" %(self.episode_counter,
                                                   self.episode_reward)
            self.episode_counter += 1
            self.episode_reward = 0
            self.episode_status = 'started'
            self.save_next_frame = False

        #self.export_image(
        #    screen_matrix,
        #    'playing_with_q/episode1/' + str(self.frame_count) + '.png')

        colorScreenMatrix = generalUtils.atari_to_color(self.color_map, screen_matrix)

        #self.export_color_image(
        #    colorScreenMatrix,
        #    'deepmind/' + str(self.frame_count) + '.png')
```

```
        grayscale = generalUtils.atari_to_grayscale(self.color_map, screen_matrix)
        resized_screen = imresize(grayscale, (83, 83))# - self.background
        resized_screen /= 255.

        if self.save_next_frame:
            self.database.add_episode(curr_state = self.prev_state, next_state = resized_screen, action =
    self.prev_action, reward = 0, terminal = 0)
            self.save_next_frame = False

        self.frame_count = self.frame_count + 1
        print 'Frame: ' + str(self.frame_count) + ', reward in this frame: ' + str(self.episode_reward)

        self.add_to_batch(resized_screen)

        if self.uniform_over_game_space:
            distribution = stats.rv_discrete(values = (np.arange(6), [0.1, 0.8, 0.1]))
            act_ind = distribution.rvs()
            if act_ind != 1 or np.random.uniform(high = 0.8) < 0.1:
                self.save_next_frame = True
        elif self.greedy_action_selection:
            act_ind = self.get_action(self.batch)
            #print self.get_influence(tmp_batch)
            #if self.frame_count == 64:
            #    for i in range(16):
            #        plt.imshow(self.get_outputs(tmp_batch, i), interpolation='nearest')
            #        plt.show()

            #plt.show()'''
            #if self.frame_count == 100:
                #plt.imshow(self.get_influence(tmp_batch)[0][act_ind], interpolation='nearest')
                #for i in range(256):
                    #print i
                    #plt.imshow(self.influence[i], interpolation='nearest')
                    #plt.show()
            #plt.imshow(self.get_influence(tmp_batch)[0][0], interpolation='nearest')
            #plt.show()
        else:
            values = (np.arange(6), self.get_probs(self.batch))
            distribution = stats.rv_discrete(values = values)
            print values
            #if np.random.uniform(high = 1.0) < 0.9:
            #    act_ind = 4
            #else:
            act_ind = distribution.rvs()

        if self.plot_histogram:
```

```
            fig = plt.figure()

            fig_0 = fig.add_subplot(1,2,1)
            fig_0.imshow(colorScreenMatrix, interpolation='nearest')

            fig_0.axes.get_xaxis().set_visible(False)
            fig_0.axes.get_yaxis().set_visible(False)
            #fig_1 = fig.add_subplot(2,2,2)
            #fig_1.imshow(resized_screen, interpolation='nearest', cmap = plt.get_cmap('gray'))

            #fig_2 = fig.add_subplot(2,2,3)
            #fig_2.imshow(self.get_influence(tmp_batch, act_ind)[0][0], interpolation='nearest')

            #print self.get_probs(tmp_batch)

            fig_3 = fig.add_subplot(1,2,2)
            fig_3.barh(np.arange(len(self.game_settings.possible_actions)), self.get_probs(self.batch))

            fig_3.set_yticks(np.arange(len(self.game_settings.possible_actions)) + 0.4)

            fig_3.set_yticklabels(self.game_settings.possible_actions )
            fig.tight_layout()

            #self.influence[0] += self.get_influence(tmp_batch, 0)[0][0]
            #plt.imshow(self.influence[0], interpolation='nearest')
            if not exists('artificial_gameplay'):
                mkdir('artificial_gameplay')
            plt.savefig('artificial_gameplay/with_hist_' + str(self.frame_count) + '.png')

        self.prev_state = resized_screen
        self.prev_action = act_ind

        #with open("playing_with_q/episode1_actions.txt", "a") as myfile:
        #    myfile.write(str(act_ind) + '\n')

        new_act = actions_map[self.game_settings.possible_actions[act_ind]]

        return new_act

def run_deepmind_agent(game_settings, which_network, plot_histogram, non_linearity, play_for_frames,
                       action_continuity_count, working_directory,
                       save_reward_history, plot_reward_history):
    "Runs A.L.E, and collects the specified number of ransom samples"
    if not exists(working_directory):
        mkdir(working_directory)
    player_agent = DeepMindAgent(game_settings, which_network, plot_histogram, non_linearity,
                                 play_for_frames, action_continuity_count,
                                 0.05, working_directory)
```

```
    run_ale(player_agent, game_settings, working_directory,
                save_reward_history, plot_reward_history)

if __name__ == "__main__":
    parser = argparse.ArgumentParser(description='Generates gameplay from a given network')
    parser.add_argument('-g','--game', help='Which game', required=True)
    parser.add_argument('-n','--network', help='Network location', required=True)
    parser.add_argument('-a', action='store_true', default=False, help = 'Plot histogram next to frame')
    parser.add_argument('-l','--non_linearity', help='tanh or ReLU', required=True)

    args = vars(parser.parse_args())

    game_settings = generalUtils.get_game_settings(args['game'])

    game_settings.uses_screen_matrix = True
    save_reward_history = True
    plot_reward_history = True
    working_directory = "./"
    play_for_frames = 1000
    action_continuity_count = 0
    run_deepmind_agent(
        game_settings,
        args['network'],
        args['a'],
        args['non_linearity'],
        play_for_frames,
        action_continuity_count,
        working_directory,
        save_reward_history,
        plot_reward_history)
```

code/playWithGivenNetwork.py

```
from deepmind_network import DeepMindNetwork
import numpy as np
import theano
import theano.tensor as T
import os
import os.path

from matplotlib import pyplot as plt

import generalUtils
from mlp import HiddenLayer
import deepmind_static_training as dst

import argparse
```

```
class RewardFunction(object):

    NEW_LAYER_SIZE = 30

    def __init__(
        self, network_file, game, history_size, non_linearity,
        batch_size = 32, discount_factor = 0.9, learning_rate = 0.1):


        game_settings = generalUtils.get_game_settings(game)
        num_actions = len(game_settings.possible_actions)
        self.batch_size = batch_size
        self.history_size = history_size

        ###############################
        # Construct network structure #
        ###############################

        state = T.tensor4('curr_state')
        state_prime = T.tensor4('next_state')
        action = T.ivector('action')

        rng = np.random.RandomState(23455)

        self.r_net = DeepMindNetwork.readFromFile(
            x = state,
            num_actions = num_actions,
            batch_size = batch_size,
            rng = rng,
            history_size = history_size,
            fileName = network_file,
            non_linearity = non_linearity)

        q_net = DeepMindNetwork.readFromFile(
            x = state,
            num_actions = num_actions,
            batch_size = batch_size,
            rng = rng,
            history_size = history_size,
            fileName = network_file,
            non_linearity = non_linearity)

        q_net_prime = DeepMindNetwork.readFromFile(
            x = state_prime,
            num_actions = num_actions,
            batch_size = batch_size,
            rng = rng,
```

```
            history_size = history_size,
            fileName = network_file,
            non_linearity = non_linearity)

        ##############################
        # Create training procedure #
        ##############################

        # TODO: Might need to take [0] on some axis
        # it should return array with size 1 for each
        # element in batch insted of just a number.
        # TODO: "batch_size" instead of "32".
        r_s_a = T.reshape(self.r_net.layer3.before_softmax, (32,))

        q_s_a = q_net.layer3.before_softmax[np.arange(32), action]
        max_q_s_prime_a_prime = T.max(q_net_prime.layer3.before_softmax, axis=1)

        cost = T.sum(T.sqr(r_s_a - (q_s_a - discount_factor * max_q_s_prime_a_prime)))

        #self.debug = theano.function(
        #    [state, state_prime, action],
        #    [distance],
        #    allow_input_downcast = True)

        # This is cost for the whole batch. "distance"
        # is an array with same size as the batch.
        # TODO: L1 instead od L2?
        #cost = distance.norm(L = 2)

        updates = dst.sgd_updates_adagrad(
            cost = cost,
            params = self.r_net.params,
            learning_rate = learning_rate,
            consider_constant=[q_s_a, max_q_s_prime_a_prime])

        self.train_model = theano.function(
            [state, state_prime, action],
            cost,
            updates = updates,
            allow_input_downcast = True)

    @staticmethod
    def load_transitions_from_file(fileName, history_size):
        state = np.load(fileName)['curr_state']
        state_prime = state[1:]
        state = state[:state.shape[0] - 1]
        action = np.load(fileName)['action']
        action = action[:action.shape[0] - 1]
```

```
        # TODO: Shuffle. I should share load code
        # with deepmind_static_training2.py

        return [dst.create_multiframe_states(state, history_size), dst.create_multiframe_states(state_prime
, history_size), action]


    def train(self, transition_data_folder):

        num_transition_files = generalUtils.get_num_files(transition_data_folder, '.npz')

        for k in range(num_transition_files):
            random_index = k #np.random.randint(low = 0, high = num_transition_files)
            [state, state_prime, action] = RewardFunction.load_transitions_from_file(
                transition_data_folder + str(random_index) + '.npz',
                self.history_size)
            print 'using episode ' + str(random_index)
            for i in range(200):
                print 'batch: ' + str(i)
                indices = np.random.randint(low = 0, high = state.shape[0], size = self.batch_size)
                # print self.debug(state[indices], state_prime[indices], action[indices])
                self.train_model(state[indices], state_prime[indices], action[indices])
            self.r_net.saveToFile(transition_data_folder + 'r_net_' + str(k))


if __name__ == "__main__":
    parser = argparse.ArgumentParser(description='Generates gameplay from a given network')
    parser.add_argument('-g','--game', help='Which game', required=True)
    parser.add_argument('-s','--history_size', help='Number of frames in input', required=True, type = int)
    parser.add_argument('-n','--network', help='Network location', required=True)
    parser.add_argument('-d','--database', help='Human gameplay database', required=True)
    parser.add_argument('-l','--non_linearity', help='tanh or ReLU', required=True)


    args = vars(parser.parse_args())


    reward_function = RewardFunction(args['network'], args['game'], args['history_size'], args['
non_linearity'])
    reward_function.train(args['database'])
```

code/generate_reward_function.py

APPENDIX B

# Video Demonstration

*"There are things known and there are things unknown, and in between are the doors of perception."*

– Aldous Huxley, *1954*

A video demonstration of an agent trained with our method for the game Freeway is given as a DVD accompanying the dissertation. A sample frame of the video is given in Figure B.1. Besides original and preprocessed screens from the game, histograms about the probability distribution of actions and influence maps as discussed in Chapter 4 are also given.

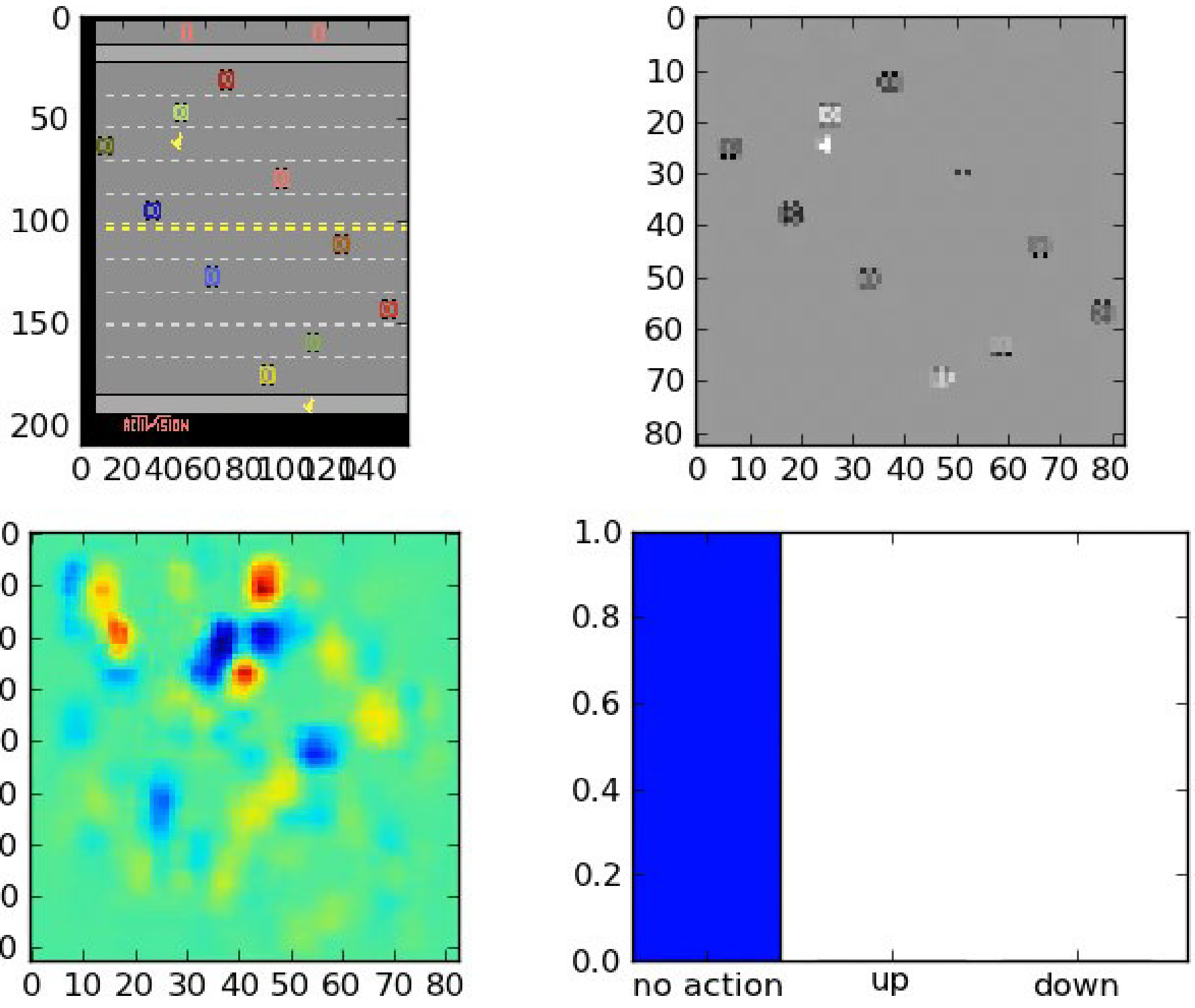

**Figure B.1:** Sample frame from the accompanying DVD